\documentclass[10pt,twocolumn,letterpaper]{article}

\usepackage[pagenumbers]{cvpr} %

\usepackage{booktabs}       %
\usepackage{amsfonts}       %
\usepackage{nicefrac}       %
\usepackage{microtype}      %
\usepackage{xcolor}         %

\usepackage{multirow}
\usepackage{colortbl}
\usepackage{graphicx}
\usepackage{wrapfig}
\usepackage{amssymb}
\newcommand{\cmark}{$\checkmark$}

\newcommand{\best}[1]{\textbf{#1}}

\providecommand{\RealAucC}{93.73 \pm 0.01}
\providecommand{\RealAucM}{79.31 \pm 0.39}

\providecommand{\EdmTrFidC}{12.6}
\providecommand{\EdmTrNrC}{0.400}
\providecommand{\EdmTrIrsC}{0.594}
\providecommand{\EdmTrCosC}{0.886}
\providecommand{\EdmTrAucC}{92.32 \pm 0.16}
\providecommand{\EdmTrFidM}{15.6}
\providecommand{\EdmTrNrM}{0.343}
\providecommand{\EdmTrIrsM}{0.499}
\providecommand{\EdmTrCosM}{0.907}
\providecommand{\EdmTrAucM}{74.50 \pm 0.86}

\providecommand{\EdmUnFidC}{8.9}
\providecommand{\EdmUnNrC}{0.427}
\providecommand{\EdmUnIrsC}{0.635}
\providecommand{\EdmUnCosC}{0.874}
\providecommand{\EdmUnAucC}{92.94 \pm 0.06}
\providecommand{\EdmUnFidM}{11.6}
\providecommand{\EdmUnNrM}{0.336}
\providecommand{\EdmUnIrsM}{0.712}
\providecommand{\EdmUnCosM}{0.905}
\providecommand{\EdmUnAucM}{75.74 \pm 0.79}

\providecommand{\EdmAgTrFidC}{13.3}
\providecommand{\EdmAgTrNrC}{0.467}
\providecommand{\EdmAgTrIrsC}{0.578}
\providecommand{\EdmAgTrCosC}{0.883}
\providecommand{\EdmAgTrAucC}{92.23 \pm 0.09}
\providecommand{\EdmAgTrFidM}{14.0}
\providecommand{\EdmAgTrNrM}{0.233}
\providecommand{\EdmAgTrIrsM}{0.506}
\providecommand{\EdmAgTrCosM}{0.900}
\providecommand{\EdmAgTrAucM}{71.84 \pm 1.08}

\providecommand{\EdmAgUnFidC}{5.4}
\providecommand{\EdmAgUnNrC}{0.480}
\providecommand{\EdmAgUnIrsC}{0.733}
\providecommand{\EdmAgUnCosC}{0.868}
\providecommand{\EdmAgUnAucC}{92.94 \pm 0.04}
\providecommand{\EdmAgUnFidM}{10.2}
\providecommand{\EdmAgUnNrM}{0.216}
\providecommand{\EdmAgUnIrsM}{0.723}
\providecommand{\EdmAgUnCosM}{0.893}
\providecommand{\EdmAgUnAucM}{73.31 \pm 0.87}

\providecommand{\NoIdTrFidC}{14.2}
\providecommand{\NoIdTrNrC}{0.212}
\providecommand{\NoIdTrIrsC}{0.586}
\providecommand{\NoIdTrCosC}{0.895}
\providecommand{\NoIdTrAucC}{91.27 \pm 0.16}
\providecommand{\NoIdTrFidM}{16.0}
\providecommand{\NoIdTrNrM}{0.251}
\providecommand{\NoIdTrIrsM}{0.475}
\providecommand{\NoIdTrCosM}{0.897}
\providecommand{\NoIdTrAucM}{72.10 \pm 1.39}

\providecommand{\NoIdUnFidC}{7.2}
\providecommand{\NoIdUnNrC}{0.486}
\providecommand{\NoIdUnIrsC}{0.654}
\providecommand{\NoIdUnCosC}{0.874}
\providecommand{\NoIdUnAucC}{92.53 \pm 0.10}
\providecommand{\NoIdUnFidM}{8.4}
\providecommand{\NoIdUnNrM}{0.140}
\providecommand{\NoIdUnIrsM}{0.608}
\providecommand{\NoIdUnCosM}{0.894}
\providecommand{\NoIdUnAucM}{72.59 \pm 0.84}

\providecommand{\NoAttrTrNrC}{0.048}
\providecommand{\NoAttrTrNrM}{0.071}

\providecommand{\NoAttrUnFidC}{7.9}
\providecommand{\NoAttrUnNrC}{0.490}
\providecommand{\NoAttrUnIrsC}{0.654}
\providecommand{\NoAttrUnCosC}{0.779}
\providecommand{\NoAttrUnAucC}{86.69 \pm 0.22}
\providecommand{\NoAttrUnFidM}{9.4}
\providecommand{\NoAttrUnNrM}{0.178}
\providecommand{\NoAttrUnIrsM}{0.584}
\providecommand{\NoAttrUnCosM}{0.851}
\providecommand{\NoAttrUnAucM}{68.03 \pm 0.56}

\providecommand{\OursTrFidC}{11.9}
\providecommand{\OursTrNrC}{0.476}
\providecommand{\OursTrIrsC}{0.622}
\providecommand{\OursTrCosC}{0.873}
\providecommand{\OursTrAucC}{91.86 \pm 0.07}
\providecommand{\OursTrFidM}{18.0}
\providecommand{\OursTrNrM}{0.227}
\providecommand{\OursTrIrsM}{0.543}
\providecommand{\OursTrCosM}{0.890}
\providecommand{\OursTrAucM}{70.50 \pm 1.63}

\providecommand{\OursUnFidC}{7.8}
\providecommand{\OursUnNrC}{0.402}
\providecommand{\OursUnIrsC}{0.726}
\providecommand{\OursUnCosC}{0.862}
\providecommand{\OursUnAucC}{91.99 \pm 0.09}
\providecommand{\OursUnFidM}{14.1}
\providecommand{\OursUnNrM}{0.431}
\providecommand{\OursUnIrsM}{0.706}
\providecommand{\OursUnCosM}{0.890}
\providecommand{\OursUnAucM}{74.28 \pm 0.45}

\providecommand{\PairedEdmCelebATr}{92.32}
\providecommand{\PairedEdmCelebAUn}{92.94}
\providecommand{\PairedEdmCelebADelta}{+0.62}
\providecommand{\PairedEdmCelebAPt}{0.0036}
\providecommand{\PairedEdmCelebAPW}{0.0625}
\providecommand{\PairedEdmCelebACIlow}{0.46}
\providecommand{\PairedEdmCelebACIhigh}{0.82}

\providecommand{\PairedEdmAgCelebATr}{92.23}
\providecommand{\PairedEdmAgCelebAUn}{92.94}
\providecommand{\PairedEdmAgCelebADelta}{+0.72}
\providecommand{\PairedEdmAgCelebAPt}{{<}0.001}
\providecommand{\PairedEdmAgCelebAPW}{0.0625}
\providecommand{\PairedEdmAgCelebACIlow}{0.65}
\providecommand{\PairedEdmAgCelebACIhigh}{0.77}

\providecommand{\PairedNoIdCelebATr}{91.27}
\providecommand{\PairedNoIdCelebAUn}{92.53}
\providecommand{\PairedNoIdCelebADelta}{+1.26}
\providecommand{\PairedNoIdCelebAPt}{0.0004}
\providecommand{\PairedNoIdCelebAPW}{0.0625}
\providecommand{\PairedNoIdCelebACIlow}{1.07}
\providecommand{\PairedNoIdCelebACIhigh}{1.46}

\providecommand{\PairedNoAttrCelebATr}{87.58}
\providecommand{\PairedNoAttrCelebAUn}{86.69}
\providecommand{\PairedNoAttrCelebADelta}{-0.89}
\providecommand{\PairedNoAttrCelebAPt}{0.0085}
\providecommand{\PairedNoAttrCelebAPW}{0.0625}
\providecommand{\PairedNoAttrCelebACIlow}{-1.26}
\providecommand{\PairedNoAttrCelebACIhigh}{-0.57}

\providecommand{\PairedOursCelebATr}{91.86}
\providecommand{\PairedOursCelebAUn}{91.99}
\providecommand{\PairedOursCelebADelta}{+0.12}
\providecommand{\PairedOursCelebAPt}{0.0987}
\providecommand{\PairedOursCelebAPW}{0.1250}
\providecommand{\PairedOursCelebACIlow}{0.04}
\providecommand{\PairedOursCelebACIhigh}{0.24}

\providecommand{\PairedEdmMimicTr}{74.50}
\providecommand{\PairedEdmMimicUn}{75.74}
\providecommand{\PairedEdmMimicDelta}{+1.24}
\providecommand{\PairedEdmMimicPt}{0.1729}
\providecommand{\PairedEdmMimicPW}{0.1875}
\providecommand{\PairedEdmMimicCIlow}{0.21}
\providecommand{\PairedEdmMimicCIhigh}{2.85}

\providecommand{\PairedEdmAgMimicTr}{71.84}
\providecommand{\PairedEdmAgMimicUn}{73.31}
\providecommand{\PairedEdmAgMimicDelta}{+1.47}
\providecommand{\PairedEdmAgMimicPt}{0.0619}
\providecommand{\PairedEdmAgMimicPW}{0.1250}
\providecommand{\PairedEdmAgMimicCIlow}{0.50}
\providecommand{\PairedEdmAgMimicCIhigh}{2.44}

\providecommand{\PairedNoIdMimicTr}{72.10}
\providecommand{\PairedNoIdMimicUn}{72.59}
\providecommand{\PairedNoIdMimicDelta}{+0.49}
\providecommand{\PairedNoIdMimicPt}{0.6384}
\providecommand{\PairedNoIdMimicPW}{0.6250}
\providecommand{\PairedNoIdMimicCIlow}{-1.41}
\providecommand{\PairedNoIdMimicCIhigh}{1.92}

\providecommand{\PairedNoAttrMimicTr}{67.54}
\providecommand{\PairedNoAttrMimicUn}{68.03}
\providecommand{\PairedNoAttrMimicDelta}{+0.49}
\providecommand{\PairedNoAttrMimicPt}{0.7147}
\providecommand{\PairedNoAttrMimicPW}{1.0000}
\providecommand{\PairedNoAttrMimicCIlow}{-1.09}
\providecommand{\PairedNoAttrMimicCIhigh}{3.84}

\providecommand{\PairedOursMimicTr}{70.50}
\providecommand{\PairedOursMimicUn}{74.28}
\providecommand{\PairedOursMimicDelta}{+3.78}
\providecommand{\PairedOursMimicPt}{0.0052}
\providecommand{\PairedOursMimicPW}{0.0625}
\providecommand{\PairedOursMimicCIlow}{2.69}
\providecommand{\PairedOursMimicCIhigh}{5.09}

\providecommand{\MimicWithinRowOneAttr}{Consolidation}
\providecommand{\MimicWithinRowOneTr}{65.69}
\providecommand{\MimicWithinRowOneUn}{72.93}
\providecommand{\MimicWithinRowOneDelta}{-7.23}

\providecommand{\MimicWithinRowTwoAttr}{Cardiomegaly}
\providecommand{\MimicWithinRowTwoTr}{66.42}
\providecommand{\MimicWithinRowTwoUn}{72.39}
\providecommand{\MimicWithinRowTwoDelta}{-5.97}

\providecommand{\MimicWithinRowThreeAttr}{Atelectasis}
\providecommand{\MimicWithinRowThreeTr}{61.40}
\providecommand{\MimicWithinRowThreeUn}{66.74}
\providecommand{\MimicWithinRowThreeDelta}{-5.34}

\providecommand{\MimicWithinRowFourAttr}{No Finding}
\providecommand{\MimicWithinRowFourTr}{79.45}
\providecommand{\MimicWithinRowFourUn}{84.75}
\providecommand{\MimicWithinRowFourDelta}{-5.30}

\providecommand{\MimicWithinRowFiveAttr}{Pleural Effusion}
\providecommand{\MimicWithinRowFiveTr}{80.80}
\providecommand{\MimicWithinRowFiveUn}{85.53}
\providecommand{\MimicWithinRowFiveDelta}{-4.73}

\providecommand{\MimicVsEdmRowOneAttr}{Consolidation}
\providecommand{\MimicVsEdmRowOneOurs}{65.69}
\providecommand{\MimicVsEdmRowOneEdm}{72.44}
\providecommand{\MimicVsEdmRowOneDelta}{-6.75}

\providecommand{\MimicVsEdmRowTwoAttr}{Cardiomegaly}
\providecommand{\MimicVsEdmRowTwoOurs}{66.42}
\providecommand{\MimicVsEdmRowTwoEdm}{72.80}
\providecommand{\MimicVsEdmRowTwoDelta}{-6.38}

\providecommand{\MimicVsEdmRowThreeAttr}{No Finding}
\providecommand{\MimicVsEdmRowThreeOurs}{79.45}
\providecommand{\MimicVsEdmRowThreeEdm}{85.47}
\providecommand{\MimicVsEdmRowThreeDelta}{-6.02}

\providecommand{\MimicVsEdmRowFourAttr}{Pneumothorax}
\providecommand{\MimicVsEdmRowFourOurs}{72.06}
\providecommand{\MimicVsEdmRowFourEdm}{77.99}
\providecommand{\MimicVsEdmRowFourDelta}{-5.93}

\providecommand{\MimicVsEdmRowFiveAttr}{Edema}
\providecommand{\MimicVsEdmRowFiveOurs}{81.41}
\providecommand{\MimicVsEdmRowFiveEdm}{85.83}
\providecommand{\MimicVsEdmRowFiveDelta}{-4.42}

\providecommand{\OursTrRC}{0.861}
\providecommand{\OursTrRM}{0.906}

\providecommand{\AudTauC}{0.658}
\providecommand{\AudTauM}{0.553}
\providecommand{\AudRefC}{5.4}
\providecommand{\AudRefM}{5.5}
\providecommand{\AudSrcEdmC}{8.4}
\providecommand{\AudOthEdmC}{3.1}
\providecommand{\AudSrcEdmM}{0.1}
\providecommand{\AudOthEdmM}{6.8}
\providecommand{\AudSrcEdmAgC}{20.1}
\providecommand{\AudOthEdmAgC}{6.7}
\providecommand{\AudSrcEdmAgM}{0.9}
\providecommand{\AudOthEdmAgM}{17.2}
\providecommand{\AudSrcNoAttrC}{96.9}
\providecommand{\AudOthNoAttrC}{6.0}
\providecommand{\AudSrcNoAttrM}{95.0}
\providecommand{\AudOthNoAttrM}{5.8}
\providecommand{\AudSrcNoIdC}{40.3}
\providecommand{\AudOthNoIdC}{4.1}
\providecommand{\AudSrcNoIdM}{3.2}
\providecommand{\AudOthNoIdM}{15.3}
\providecommand{\AudSrcOursC}{1.3}
\providecommand{\AudOthOursC}{5.6}
\providecommand{\AudSrcOursM}{0.1}
\providecommand{\AudOthOursM}{24.3}

\usepackage{tikz}
\usepackage{amsmath}
\usepackage{amssymb}
\usepackage{amsthm}
\usetikzlibrary{arrows.meta, positioning, decorations.pathreplacing, patterns, calc}
\usepackage{pgfplots}
\usepgfplotslibrary{groupplots}
\pgfplotsset{compat=newest}
\newtheorem{proposition}{Proposition}

\definecolor{darkblue}{RGB}{0, 51, 102}
\definecolor{darkred}{RGB}{153, 0, 0}
\definecolor{darkgreen}{RGB}{0, 102, 51}

\definecolor{primaryblue}{RGB}{24, 86, 178}
\definecolor{secondaryblue}{RGB}{35, 120, 195}

\definecolor{plum}{RGB}{102, 45, 145}
\definecolor{forest}{RGB}{27, 94, 32}
\definecolor{ocean}{RGB}{0, 96, 160}
\definecolor{burgundy}{RGB}{128, 0, 32}
\definecolor{sage}{RGB}{106, 153, 78}
\definecolor{steel}{RGB}{70, 130, 180}

\graphicspath{{./}}

\providecommand{\secondbest}[1]{\underline{#1}}
\definecolor{ecblue}{HTML}{4C72B0}
\definecolor{emorange}{HTML}{DD8452}

\PassOptionsToPackage{table}{xcolor}
\usepackage{acro}
\acsetup{make-links=false}

\usepackage{amsmath,amsfonts,bm}

\def\eqref#1{equation~\ref{#1}}

\def\1{\bm{1}}

\newcommand{\imgdim}{H \times W \times C}     %

\newcommand{\idl}{\mathrm{id}}                %
\newcommand{\reid}{R}                         %

\DeclareRobustCommand{\FID}{\ensuremath{\text{FID}}\xspace}
\DeclareRobustCommand{\IRS}{\ensuremath{\text{IRS}}\xspace}
\DeclareRobustCommand{\AUCROC}{\ensuremath{\text{AUC-ROC}}\xspace}

\DeclareRobustCommand{\RatOne}{\ensuremath{\text{R@1}}\xspace}

\input{acronyms/acronym-symbol.ac}
\input{acronyms/acronym-abbreviation.ac}
\usepackage{colortbl}
\definecolor{fau}{HTML}{04316A}
\definecolor{fau_tech}{HTML}{8C9FB1}
\definecolor{fau_phil}{HTML}{FDB735}
\definecolor{fau_rw}{HTML}{C50F3C}
\definecolor{fau_med}{HTML}{18B4F1}
\definecolor{fau_nat}{HTML}{7BB725}

\definecolor{fau_alpha1}{HTML}{617DA1}
\definecolor{fau_alpha2}{HTML}{A0B1C6}
\definecolor{fau_alpha3}{HTML}{C0CBDA}

\definecolor{fau_tech_alpha1}{HTML}{B6C2CE}
\definecolor{fau_tech_alpha2}{HTML}{D3DAE1}
\definecolor{fau_tech_alpha3}{HTML}{E2E7EB}

\definecolor{fau_nat_alpha1}{HTML}{ACD275}
\definecolor{fau_nat_alpha2}{HTML}{CDE4AC}
\definecolor{fau_nat_alpha3}{HTML}{DEEDC8}

\definecolor{fau_med_alpha1}{HTML}{6DD0F6}
\definecolor{fau_med_alpha2}{HTML}{A7E2FA}
\definecolor{fau_med_alpha3}{HTML}{C5ECFB}

\definecolor{fau_rw_alpha1}{HTML}{DA6784}
\definecolor{fau_rw_alpha2}{HTML}{E9A4B5}
\definecolor{fau_rw_alpha3}{HTML}{F0C3CE}

\definecolor{fau_phil_alpha1}{HTML}{FECE76}
\definecolor{fau_phil_alpha2}{HTML}{FEE4B2}
\definecolor{fau_phil_alpha3}{HTML}{FEEDCC}

\definecolor{fau_light}{HTML}{CED9E7}
\definecolor{fau_light_tech}{HTML}{D3DDE6}
\definecolor{fau_light_phil}{HTML}{F4E8CB}
\definecolor{fau_light_rw}{HTML}{F2DED1}
\definecolor{fau_light_med}{HTML}{B9D9EC}
\definecolor{fau_light_nat}{HTML}{CFE0D8}

\definecolor{fau_dark}{HTML}{041E42}
\definecolor{fau_dark_tech}{HTML}{2F586E}
\definecolor{fau_dark_phil}{HTML}{FDB735}
\definecolor{fau_dark_rw}{HTML}{971B2F}
\definecolor{fau_dark_med}{HTML}{005287}
\definecolor{fau_dark_nat}{HTML}{266141}

\definecolor{ecblue}{HTML}{4C72B0}
\definecolor{emorange}{HTML}{DD8452}

\iftoggle{cvprfinal}{}{%
  \definecolor{fau}{gray}{0.25}%
  \definecolor{fau_tech_alpha3}{gray}{0.90}%
}

\definecolor{cvprblue}{rgb}{0.21,0.49,0.74}
\usepackage[pagebackref,breaklinks,colorlinks,allcolors=cvprblue]{hyperref}

\title{TraceDiffusion: Traceable Image Generation for Auditable Private Data Sharing}

\author{Mischa Dombrowski$^{1}$\quad
Felix N\"utzel$^{1}$\quad
Jiayi Wang$^{1}$\quad
Bernhard Kainz$^{1,2}$\\
$^{1}$Friedrich-Alexander-Universit\"at Erlangen-N\"urnberg\quad
$^{2}$Imperial College London
}

\begin{document}
\maketitle
\begin{abstract}
Vast repositories of images are locked behind closed doors or are freely accessible online, but unclear sharing policies and copyright restrictions increasingly limit their use for model training. 
Synthetic generation promises a compliant alternative, yet diffusion models risk silently reproducing the identities of individuals present in their training data.
Existing approaches either assume that privacy is not compromised or use post hoc sample filtering.
We define \emph{traceability}, a per-sample certificate anchoring every generated image to a named training source, and introduce \emph{TraceDiffusion} to instantiate it via two adversarially disentangled encoders and a guidance scheme that steers generation away from a source identity while preserving its class attributes.
An independent memorization detector that compares every sample with the entire training set confirms the effect of this guidance.
On CelebA, $\AudSrcNoAttrC\%$ of the identity-conditioned reconstructions match their source identity, but only $\AudSrcOursC\%$ of the guided samples do, while matches to any \emph{other} training identity ($\AudOthOursC\%$) stay at the false-alarm level of real unseen images ($\AudRefC\%$).
On MIMIC-CXR the source patient is left as reliably ($\AudSrcNoAttrM\%$ to $\AudSrcOursM\%$).
We further show that disentangling input encoders leads to disentangled output generations: the class-conditioned model achieves near-chance identity retrieval ($1.1\%$) while the identity-conditioned model reaches above $62\%$, confirming clean separation of the two information streams.
On CelebA, traceability costs only $0.12$\,pp of \AUCROC{} in downstream classification, five to ten times less than for the baselines.
Code will be made available upon publication.
\end{abstract}

\begin{figure*}[t]
\centering\small
\begin{minipage}[c]{8pt}
  \centering\rotatebox{90}{\textbf{CelebA}}
\end{minipage}%
\hspace{2pt}%
\begin{minipage}[c]{\dimexpr\linewidth-10pt\relax}
  \includegraphics[width=\linewidth]{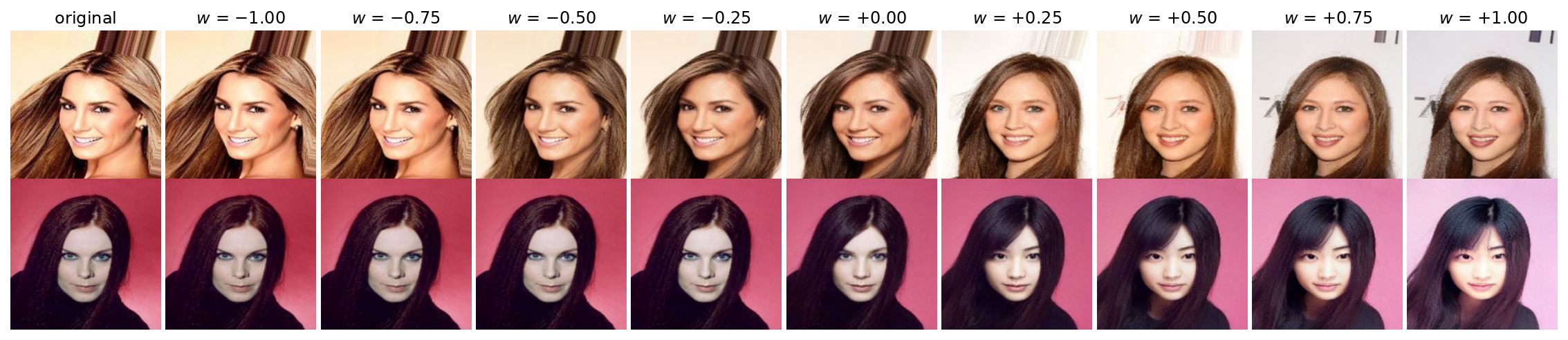}
\end{minipage}

\vspace{3pt}

\begin{minipage}[c]{8pt}
  \centering\rotatebox{90}{\textbf{MIMIC-CXR}}
\end{minipage}%
\hspace{2pt}%
\begin{minipage}[c]{\dimexpr\linewidth-10pt\relax}
  \includegraphics[trim={0 0 0 10pt}, clip, width=\linewidth]{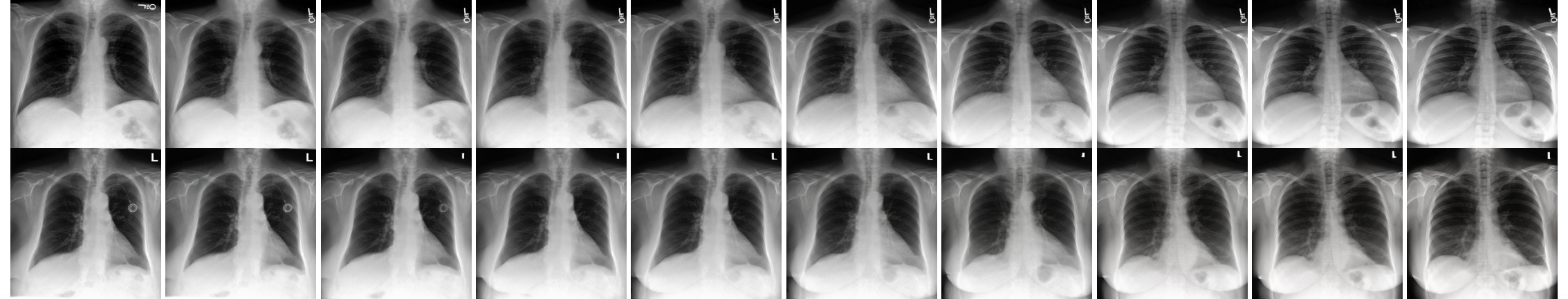}
\end{minipage}
\caption[Memory-guided generation at varying guidance scales]{%
    \textbf{Memory-guided generation at varying guidance scales.}
    Each panel shows two source identities (rows) alongside outputs at increasing guidance
    scales $w \in \{-1.00, -0.75, \ldots, +1.00\}$.
    At $w = -1$ (leftmost generated column) the denoiser reduces to $\epsilon_\theta^i$ and
    reproduces the source identity; at $w = 0$ it reduces to the class-conditional
    baseline $\epsilon_\theta^c$; increasing $w > 0$ steers generation progressively away
    from the source identity while preserving class attributes.
    \emph{Top:} CelebA face generation (40 binary attributes).
    \emph{Bottom:} MIMIC-CXR chest X-ray generation (14 pathology labels).
}
\label{traceability:fig:qualitative}
\end{figure*}

\section{Introduction}
The landscape of generative modeling has undergone a paradigm shift with the emergence of diffusion models, which now achieve unprecedented photorealistic synthesis. 
By consistently improving in Fréchet Inception Distance (\FID), sample diversity, and fine-grained controllability \cite{heusel2017gans,dhariwal2021diffusion,karras2024guiding,sajjadiassessinggenerative}, 
these models have become practical tools. 
Modern frameworks now excel at tasks such as class-conditional generation, text-to-image synthesis, inpainting, and super-resolution \cite{dombrowski_iccv,nichol2022glidetowards,zhalanguageguided,tumanyan2022plugplay}, 
establishing them as highly effective tools for automated data augmentation and content creation. 
This evolution is best exemplified by landmark architectures such as Stable Diffusion, and EDM2 \cite{karras2024guiding,karras_analyzing_2024,peebles_scalable_2023,rombach2022high}, which have set new benchmarks for fidelity and scalability in the field.

Despite advancements in generative modeling, architectures lacking robust semantic priors, such as explicit anatomical structure or identity constraints, fail to bridge the Train Synthetic,
Test Real (TSTR) gap, as synthetic distributions remain misaligned with the complexities of downstream data \cite{geng2025unmetpromise,alemohammad2023selfconsuming,dar2024unconditional}.
Without such priors, models frequently fall into mode collapse and skewed within-class diversity \cite{xing2023youdon,verine2023precisionrecall}. 
Stronger reliance on labels usually decreases this gap, but the influence on privacy preservation remains underexplored \cite{ktena2024generativemodels,dombrowski_bridge}.

Despite the unprecedented fidelity of modern diffusion models, their propensity to memorize and reconstruct training data presents significant privacy risks, often leading to the unintended leakage of individual identities, which undermines their use  as digital twins \cite{carlini2023extracting, dombrowski_midl, tang2023membershipinference}. 
Current mitigation strategies predominantly rely on post-hoc, density-based dataset filtering \cite{dombrowski_bridge, dar2024unconditional}; 
however, these approaches lack a principled definition of "similarity," resulting in arbitrary thresholds that fail to provide rigorous privacy guarantees \cite{koeken2025sensitivityspecificity}. 
Consequently, while digital twins potentially could provide a centralized alternative to federated learning, their regulatory adoption remains limited by the lack of auditable evidence required under regulations such as the GDPR, which connects anonymity to the possibility of singling out people \cite{noauthor_opinion_nodate,noauthor_gdpr_nodate,noauthor_gdpr_nodate-1,vollmer_recital_2023}.

We address this deficiency by introducing \emph{traceability} as a new operational primitive for synthetic data.
Every generated sample is anchored to a named training image whose identity it must demonstrably not reproduce.
Unlike post-hoc filtering, traceability provides per-sample evidence rather than global similarity bounds; unlike formal differential privacy, it requires no upfront utility tax.

To investigate this, we propose \emph{TraceDiffusion}: adversarially disentangled encoders supply class and identity conditioning to two latent diffusion models, combined through guidance that steers generation \emph{away} from a named source identity while preserving the target label cell. 
A re-identification oracle certifies novelty per sample, making every generation auditably traceable to its source.
An independent one-vs-all audit against the full training set then tests whether leaving the source identity leads into another training identity.

We make the following contributions:
\textbf{(i)}~We introduce \emph{traceability} as a per-sample auditable primitive for identity-novel generation and propose a paired traceability-utility test as a reusable evaluation axis for privacy critical diffusion models.
\textbf{(ii)}~We propose TraceDiffusion, which instantiates traceability via two adversarially disentangled encoders ($E_c$, $E_i$), two latent diffusion models, and guidance away from the identity towards the attributes.
\textbf{(iii)}~An independent detector audits every generated sample against the entire training set: on CelebA, guidance lowers the share of samples that match their source identity from $\AudSrcNoAttrC\%$ (reconstruction) to $\AudSrcOursC\%$, while matches to other training identities ($\AudOthOursC\%$) stay at the false-alarm level of real unseen images ($\AudRefC\%$).
\textbf{(iv)}~on CelebA, TraceDiffusion is the only method whose traceable start costs almost no downstream utility  ($\Delta{=}0.12$\,pp \AUCROC, $5$-$10\times$ smaller than all baselines at $p{<}0.01$);
\textbf{(v)}~on MIMIC-CXR the method leaves the source identity in only $\AudSrcOursM\%$ of samples, and the paired utility analysis localises the remaining cost ($\Delta{=}3.78$) to chronic structural findings, establishing identity-pathology entanglement as the primary cost factor.

\section{Related Work}
\noindent\textbf{Diffusion models.}
Diffusion models learn to progressively denoise a signal through a learned score function~\cite{ho2020denoising};
latent diffusion models \cite{rombach2022high,kouzelis2025eqvae} compress this process into the space of a pretrained autoencoder for efficient, high-fidelity generation, with EDM2~\cite{karras_elucidating_2022,karras_analyzing_2024,karras2024guiding} further refining training dynamics and scaling.
Classifier guidance~\cite{dhariwal2021diffusion} and classifier-free guidance~\cite{ho2022classifier,shi2023exploringcompositional} steer the denoising trajectory by contrasting conditional and unconditional score estimates; DDIM~\cite{song2020denoising} independently enables deterministic sampling and exact inversion~\cite{hertz2022prompttopromptimageeditingcross,Mokady_2023_CVPR}, underpinning image editing workflows.
Conditioning strategies range from class labels and text embeddings to image-derived representations \cite{zhalanguageguided,zhang2023textimage,nichol2022glidetowards,zhang_adding_2023,dombrowski_cvpr,dombrowski_bridge,wang2026difficultycontrolled}.

\noindent\textbf{Disentangled representation learning.}
$\beta$-VAE~\cite{Higgins2016betaVAELB}, FactorVAE~\cite{kim2019disentanglingfactorising}, and $\beta$-TCVAE~\cite{NEURIPS2018_1ee3dfcd} decouple factors of variation by penalising the KL divergence or total correlation between latent dimensions; domain-adversarial networks~\cite{ganin2016domainadversarial} instead enforce invariance via a gradient reversal layer.
We build on this paradigm by combining supervised contrastive training with CLUB mutual information minimisation~\cite{cheng2020clubcontrastive} to disentangle patient identity from pathology labels, enforced by a hard post-hoc \AUCROC gate prior to diffusion training.

\noindent\textbf{Memorization and privacy.}
Diffusion models memorize training data and can be queried to reproduce individual samples~\cite{dombrowski_midl,somepalli2023understandingmitigating,bonnaire_why_2025,carlini2023extracting,somepalli2023diffusion,kriplani2025solidmarkevaluating}. 
Membership inference attacks~\cite{carlini_membership_2022,tang2023membershipinference,pang2025whitebox,pang2024blackbox,li2025towardsblack} allow adversaries to confirm training set membership, demonstrating privacy issues.
Mitigation strategies include post-hoc filtering \cite{dar2024unconditional,reynaud2024echonet,packhäuser2022deeplearning} and differential privacy \cite{mcsherry2007mechanismdesign,ghalebikesabi2023differentiallyprivate,lin2023differentiallyprivate,dockhorn2023differentiallyprivate,tsai2025differentially}, 
but the former lacks a principled similarity threshold \cite{koeken2025sensitivityspecificity} and the latter imposes a blanket utility cost with no per-sample auditability.
Other methods use adversarial training to transform existing images into label-preserving images that are labeled as coming from a different identity~\cite{packhäuser2023deeplearningbased}. 
Concept erasure methods \cite{gandikota2023erasing,zhang2024forget} remove identities by fine-tuning with negated score objectives, but require knowing which samples to erase and offer no per-sample audit trail.
Our approach anchors each generated image to a specific training source at generation time. 
By replacing population-level bounds with per-sample traceable novelty claims, we achieve traceability at only twice the sampling time. %

\noindent\textbf{Re-identification and face verification.}
Identity preservation methods for multi-view synthesis aim to maintain a consistent identity across outputs~\cite{papantoniou2024arc2face,galanakis2025spinmeround,papantoniou2025id}, while face protection approaches pursue the opposite goal via adversarial perturbations optimised against face-specific metrics such as ArcFace~\cite{le2024styleadv,le2025diffprivate,deng2019arcface,wang2025diffusion}.
Unlike both, TraceDiffusion targets no predetermined identity and relies on no domain-specific verifier, transferring from faces to chest X-ray with the same pipeline.

\section{Method}\label{traceability:sec:method}

\noindent\textbf{Data Space and Label-Identity Structure.}
Let $\mathcal{D} \subset \mathbb{R}^{\imgdim} \times \{1,\dots,n\} \times \mathcal{I}$ denote the data space, where $\mathbb{R}^{\imgdim}$ is the image space, $\{1,\dots,n\}$ is a finite set of semantic class labels (e.g.\ hair color, age), and $\mathcal{I} = \{i_1, \ldots, i_m\}$ is the set of person identities. 
We assume $n \ll m$, reflecting the practical setting in which the number of distinct attribute classes is far smaller than the number of unique identities.

The label assignment $\ell : \mathbb{R}^{\imgdim} \rightarrow \{1,\dots,n\}$ induces a partition of $\mathcal{D}$ into label cells:
\begin{equation}
    \mathcal{D}_c = \{ x \in \mathcal{D} \mid \ell(x) = c \}, \quad c \in \{1,\dots,n\}.
\end{equation}
Since $n \ll m$, each cell $\mathcal{D}_c$ contains on average $\bar{k} = m/n \gg 1$ distinct identities, and we further assume that label cells are approximately balanced (i.e.\ the cell sizes $|\mathcal{D}_c|$ are of similar order across $c$), so that each individual cell is also densely populated. 
This means that one label does not fully determine the identity.

More precisely, we distinguish the \emph{observed} identity set $\mathcal{I}_{\text{obs}} = \{i_1, \ldots, i_m\}$ from the full underlying population $\mathcal{I}_{\text{full}}$ with $|\mathcal{I}_{\text{full}}| \gg |\mathcal{I}_{\text{obs}}|$. 
Since labels are strictly coarser than identities, membership in $\mathcal{D}_{c^*}$ does not exclude identities in $\mathcal{I}_{\text{full}} \setminus \mathcal{I}_{\text{obs}}$: novel identities are not ruled out by the label assignment.

\begin{figure*}[t]
\centering
\resizebox{\textwidth}{!}{%
\includegraphics{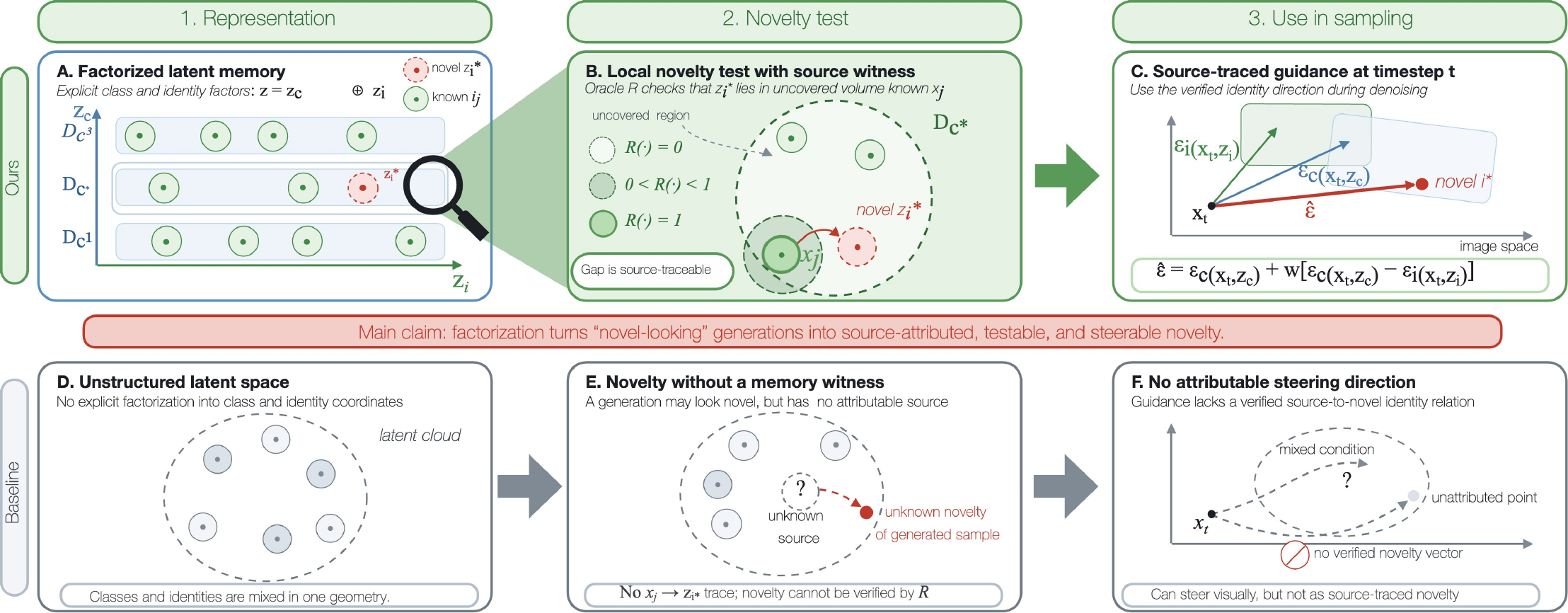}
}%
\caption[Latent space decomposition, packing argument, and guidance]{
    \textbf{Latent space decomposition, packing argument, and guidance.}
    \emph{Left:} The latent space $\mathcal{Z} = \mathcal{Z}_c \oplus \mathcal{Z}_i$
    is partitioned into label cells $\mathcal{D}_{c_k}$ along $\mathcal{Z}_c$.
    Within each cell, variation along $\mathcal{Z}_i$ is purely identity variation
    (adversarial disentanglement, $I(z_i;c)\approx 0$).
    \emph{Middle:} A cross-section of $\mathcal{Z}_i$ at target label $c^*$.
    Known identities occupy $\delta$-balls; since
    $m(\delta/\rho)^{d_i}\!\ll\!1$, the uncovered region is vast, guaranteeing
    novel $z_i^*$ exists (\cref{traceability:prop:novelty}).
    \emph{Right:} At diffusion timestep $t$, the class score
    $\epsilon_\theta^c$ (blue) and identity score $\epsilon_\theta^i$ (orange)
    point toward their respective manifolds. 
    Subtracting the memory direction
    and scaling by $w$ steers the trajectory $\hat{\epsilon}$ (red) away from
    known identities and onto the label-consistent manifold $\mathcal{D}_{c^*}$,
    yielding a novel identity $i^*$.
}
\label{traceability:fig:latent_decomp}
\end{figure*}

\noindent\textbf{Novel Identity Generation.}
We now show that $\mathcal{Z}_i$ is not exhausted by the $m$ observed identities; Fig.~\ref{traceability:fig:latent_decomp} illustrates the latent decomposition (left), the $\delta$-ball packing geometry (middle), and the resulting guidance trajectory (right).
Let $\delta > 0$ be the \emph{identity separation radius}. 
We assume that the contrastive training objective enforces mutual separation among observed embeddings: $\|z_i^{(j)} - z_i^{(k)}\| > 2\delta$ for all $j \neq k$. 
In practice $\delta$ is estimated as half the 5th percentile of pairwise distances in $\mathcal{Z}_i$. 
Under this assumption the $\delta$-balls $\{B(z_i^{(j)}, \delta)\}_{j=1}^m$ are disjoint, and the fraction of $\mathcal{Z}_i$ (modelled as a ball of radius $\rho$) covered by known identities satisfies:
\begin{equation}
    \frac{\mathrm{Vol}\!\left(\bigcup_{j=1}^m B_j\right)}{\mathrm{Vol}(\mathcal{Z}_i)}
    \leq m \left(\frac{\delta}{\rho}\right)^{d_i} < 1,
\end{equation}
provided $d_i > \dfrac{\log m}{\log(\rho/\delta)}$, which holds whenever the identity subspace is sufficiently high-dimensional relative to the packing ratio $\rho/\delta$. 
We verify this condition empirically for both datasets in Appx.~\ref{traceability:sec:app_delta}.

\begin{proposition}[Novel Identity Existence]
\label{traceability:prop:novelty}
Let $\mathcal{Z}_i$ be the identity subspace obtained under adversarial disentanglement, with $\dim(\mathcal{Z}_i) = d_i > \frac{\log m}{\log(\rho/\delta)}$. 
Let $\{z_i^{(1)}, \ldots, z_i^{(m)}\}$ be the embeddings of observed identities satisfying $\|z_i^{(j)} - z_i^{(k)}\| > 2\delta$ for all $j \neq k$. 
Then for any target label $c^* \in \{1,\dots,n\}$, there exists a point $z_i^* \in \mathcal{Z}_i$ such that:
\begin{equation}
    \|z_i^* - z_i^{(j)}\| > \delta \quad \forall\, j \in \{1, \ldots, m\},
\end{equation}
and, if $\reid$ is consistent with $\mathcal{Z}_i$ at radius $\delta$ (i.e., $\reid(x, x') = 0$ whenever the identity embeddings of $x$ and $x'$ are more than $\delta$ apart), the generated image $G(z_c^* \oplus z_i^*)$, where $z_c^* \in \mathcal{Z}_c$ is the class latent encoding $c^*$, satisfies $\reid(G(z_c^* \oplus z_i^*),\, x_j) = 0$ for all $j \in \{1, \ldots, m\}$.
\end{proposition}

\noindent\textbf{Disentangled Encoder.}
We use two independent encoders, $E_c : \mathbb{R}^{\imgdim} \to \mathcal{Z}_c \subset \mathbb{R}^{d_c}$ and $E_i : \mathbb{R}^{\imgdim} \to \mathcal{Z}_i \subset \mathbb{R}^{d_i}$, 
yielding $z_c = E_c(x)$ and $z_i = E_i(x)$ with $z = (z_c, z_i) \in \mathcal{Z}$. 
$E_c$ is supervised by a multi-label BCE loss on class labels; $E_i$ is trained with a Siamese contrastive loss to be discriminative over identities. 
To enforce $I(z_i;\, c) \approx 0$, $E_i$ is trained by minimizing the CLUB upper bound on mutual information~\cite{cheng2020clubcontrastive},
validated post-hoc by an attribute classifier on $z_i$ (\cref{traceability:par:encoder_ablation}).
Full architecture and training details are in Appx.~\ref{traceability:sec:app_encoder} (Fig.~\ref{traceability:fig:encoder}).

\noindent\textbf{Two Diffusion Models.}
$z_c$ and $z_i$ condition two separate latent diffusion models~\cite{rombach2022high} trained with the standard denoising objective: $\epsilon_\theta^c(x_t, z_c)$ captures label-level variation and $\epsilon_\theta^i(x_t, z_i)$ captures identity-specific variation.

\noindent\textbf{Re-identification Oracle.}
Let $\reid : \mathbb{R}^{\imgdim} \times \mathbb{R}^{\imgdim} \to \{0, 1\}$ be a binary re-identification oracle, where $\reid(x, x') = 1$ iff $x$ and $x'$ share the same identity. 
We instantiate $\reid$ with a learned Siamese re-identification network, which we refer to as the \emph{Oracle} throughout. 
A shared ConvNeXt-Tiny~\cite{liu2022convnext} backbone embeds the pre-computed FLUX.2 latents of both images, and a comparison head maps the absolute difference of the two embeddings to a same-identity probability, thresholded at $0.5$; it is trained with a combined binary cross-entropy and NT-Xent contrastive loss. 
Training details and per-dataset performance are in Appx.~\ref{traceability:sec:app_lcmem}. 
The Oracle is used for all method components and analyses; the independently trained LCMem~\cite{dombrowski_lcmem} is used only for the final evaluation.

\noindent\textbf{Guidance Formulation.}
At sampling time, we pick a \emph{known} source image $x_j$ and extract both its class and identity embeddings, $z_c^{(j)} = E_c(x_j)$ and $z_i^{(j)} = E_i(x_j)$. 
Generation is then steered away from this specific identity while remaining inside the target label cell $\mathcal{D}_{c^*}$ encoded by $z_c^{(j)}$:
\begin{equation}
    \hat{\epsilon} = \epsilon_\theta^c(x_t, z_c^{(j)}) + w \left( \epsilon_\theta^c(x_t, z_c^{(j)}) - \epsilon_\theta^i(x_t, z_i^{(j)}) \right),
    \label{traceability:eq:guidance}
\end{equation}
where $w \geq 0$ is the guidance scale. 
Both conditioning vectors originate from the same source $x_j$, so the target label cell is supplied implicitly through $z_c^{(j)}$ rather than via a separate label encoding. 
We use a fixed $w = 1$ in all experiments. 
Grounding generation in a known $(z_c^{(j)}, z_i^{(j)})$ makes novelty fully traceable: $\reid$ can be evaluated directly against the source $x_j$.

Disentanglement is a structural requirement: if $z_i$ retains label information, the guidance difference $\epsilon_\theta^c - \epsilon_\theta^i$ conflates identity and class variation, corrupting both novelty and class consistency simultaneously (Appx. Fig.~\ref{traceability:fig:app_all_mimic}).
Symmetrically, if $z_c$ absorbs identity variation, the guidance scale $w$ loses its role as a controllable novelty knob and the ReID certificate $\reid$ can no longer distinguish true identity novelty from attribute change.
We validate both conditions empirically and discuss all failure modes in Appx.~\ref{traceability:sec:app_disentanglement_necessity}.

\noindent\textbf{Traceability.}
Because generation is conditioned on a named source $x_j$, every output $\hat{x}$ carries a complete audit trail: the source identity, the guidance scale $w$, and the resulting $\reid$.
As shown in Appx. Fig.~\ref{traceability:fig:nearest_retrieval_i}, naive generation from random noise does not reproduce training samples. 
This is our base model without traceability.
However, knowing $x_j$ enables a stronger accountability mode we call \emph{traceability}.
We implement it via DDIM inversion: $x_j$ is mapped to a latent noise vector under $\epsilon_\theta^i$, and generation starts from this noise rather than a random sample, which roughly doubles per-image wall-clock relative to unconditional sampling (Appx.~\ref{traceability:sec:app_compute}).
Traceability is confirmed when sampling from this noise recovers an image attributed to $x_j$ by $\reid$ (i.e.\ the model has memorised the source).
Then we apply guidance with $w = 1$ and declare novelty when $\reid$ no longer attributes the output to $x_j$.
Once $\reid$ declares novelty, guidance has moved 
$z_i(\hat{x})$ away from $z_i(x_j)$, so $x_j$ no longer appears among the nearest training identities of $\hat{x}$ in $\mathcal{Z}_i$.
Knowing the departure point therefore gives an adversary no advantage in attributing the generated image: the source identity and the nearest output-space identity are decoupled by construction.

\noindent\textbf{Traceability Rate (TrR).}
Before verifying novelty, we confirm that the identity-conditioned denoiser $\epsilon_\theta^i$
is self-consistent on each source image.
Given source image $x_j$ with embedding $z_i^{(j)}$, we deterministically invert it to a
noisy latent via DDIM inversion,
$x_T^{(j)} = \mathrm{DDIM\text{-}Inv}(x_j,\, z_i^{(j)})$,
and reconstruct with the same denoiser,
$\hat{x}_j = \epsilon_\theta^i(x_T^{(j)},\, z_i^{(j)})$.
The source image is \emph{traceable} if $\reid$ confirms the reconstruction shares its identity:
\begin{equation}
    \mathrm{TrR} = \frac{1}{N} \sum_{j=1}^{N}
    \mathbf{1}\!\left[\reid(\hat{x}_j,\, x_j) = 1\right].
\end{equation}
Because the DDIM inversion is entirely deterministic (\emph{i.e.}, no stochasticity is introduced at any
step) $x_T^{(j)}$ is reproducible bit-for-bit, making every identity claim fully traceable
to a specific training sample without exhaustive nearest-neighbour search.
$\mathrm{TrR} < 1$ reflects inversion error, which trades off against the identity disentanglement of the inverting model (\cref{traceability:sec:generative_and_downstream}).

\noindent\textbf{ReID Novelty Rate (NR).}
The novelty rate measures the fraction of all $N$ source images for which memory guidance
produces a verified novel identity.
Let $\mathcal{T} = \{j : \reid(\hat{x}_j,\, x_j) = 1\}$ be the traceable subset; novelty is
only evaluated for $j \in \mathcal{T}$, since non-traceable images lack a well-defined
reconstruction baseline.
For each $j \in \mathcal{T}$ we reuse the precomputed $x_T^{(j)}$ and apply the
memory-guided denoiser (Eq.~\ref{traceability:eq:guidance}) to obtain $\hat{x}_j$.

Checking only against the source $x_j$ is insufficient: guidance away from $z_i^{(j)}$
may inadvertently steer the generation towards a \emph{different} training identity that
is nearby in $\mathcal{Z}_i$.
To guard against this, let $\mathcal{N}_k(j)$ be the indices of the $k$ nearest
neighbours of $x_j$ in identity embedding space ($k{=}5$ throughout) and
$\bar{\mathcal{N}}_k(j) = \{j\} \cup \mathcal{N}_k(j)$.
We declare $\hat{x}_j$ \emph{novel} only if it passes the ReID check against the source
\emph{and} all $k$ neighbours:
\begin{equation}
    \mathrm{NR}(w) = \frac{1}{N} \sum_{j \in \mathcal{T}}
    \mathbf{1}\!\left[\, \reid(\hat{x}_j,\, x_l) = 0 \;\; \forall\, l \in \bar{\mathcal{N}}_k(j) \right].
    \label{traceability:eq:nr}
\end{equation}
$\mathrm{NR}\,{=}\,1$ indicates every source image produced a verified novel identity.
At $w = 0$ the guidance reduces to $\epsilon_\theta^c$, the class-conditional baseline with no
identity steering; the degenerate case $w = -1$ recovers $\hat{\epsilon} = \epsilon_\theta^i$
exactly, reproducing $\hat{x}_j$ with $\mathrm{NR}(-1) = 0$.
Increasing $w$ from zero steers the trajectory away from $z_i^{(j)}$ until the identity
boundary is crossed.

\noindent\textbf{One-vs-All Audit.}
\label{traceability:sec:audit_def}
The Oracle compares a sample only with its source and $k$ neighbours.
To test a sample against the \emph{entire} training set we use LCMem~\cite{dombrowski_lcmem} as an independent auditor: it takes no part in training, in sampling, or in the computation of TrR and NR.
Let $\phi$ denote its $\ell_2$-normalised embedding and $i_l$ the identity of training image $x_l$.
For a sample $\hat{x}_j$ generated from source $x_j$ we record the best match within the source identity and the best match among all other identities,
\begin{equation}
    \begin{split}
    s_{\mathrm{src}} &= \max_{l:\, i_l = i_j} \phi(\hat{x}_j)^{\top} \phi(x_l), \\
    s_{\mathrm{oth}} &= \max_{l:\, i_l \neq i_j} \phi(\hat{x}_j)^{\top} \phi(x_l),
    \end{split}
    \label{traceability:eq:audit}
\end{equation}
where $l$ runs over all training images.
The \emph{source match rate} $\mathrm{MR}_{\mathrm{src}}$ and the \emph{other-identity match rate} $\mathrm{MR}_{\mathrm{oth}}$ are the fractions of samples with $s_{\mathrm{src}} > \tau$ and $s_{\mathrm{oth}} > \tau$.
The threshold $\tau$ is the 95th percentile of the best score of real validation images against the training set.
Since the splits are identity-disjoint, real unseen images are flagged in about $5\%$ of cases by construction; we measure this false-alarm rate on the test split and use it as the reference for $\mathrm{MR}_{\mathrm{oth}}$.
Traceable guidance should move $\mathrm{MR}_{\mathrm{src}}$ from near one at $w{=}{-}1$ to near zero at $w{=}1$ while $\mathrm{MR}_{\mathrm{oth}}$ stays at the reference.

\section{Experiments}

\noindent\textbf{Models.}
$E_c$ and $E_i$ are independent ResNet-50~\cite{he2016deep} encoders trained with CLUB~\cite{cheng2020clubcontrastive} mutual information penalties to enforce bidirectional disentanglement; full architecture and training details are in Appx.~\ref{traceability:sec:app_encoder}.
Both $\epsilon_\theta^c$ and $\epsilon_\theta^i$ are EDM2~\cite{karras_analyzing_2024} U-Nets operating in the FLUX.2-dev latent space~\cite{blackforestlabs_flux2} (32 channels), which we find improves learnability over pixel space, conditioned on $z_c$ and $z_i$ respectively via FiLM~\cite{perez2017filmvisualreasoninggeneral}, fine-tuned from a pretrained ImageNet EDM2-M checkpoint.
Training uses the EDM2 preconditioning loss with linear noise schedule $\sigma(t){=}t$ and log-normal timestep sampling, optimized with Adam ($\beta_1{=}0.9$, $\beta_2{=}0.99$) and EMA weight averaging.
Guided denoising at inference follows Eq.~\ref{traceability:eq:guidance} with fixed $w{=}1$, using the Heun second-order sampler with 32 discretization steps.
Full hyperparameters are in Appx.~\ref{traceability:sec:app_diffusion}.

\noindent\textbf{Evaluation Metrics.}
\label{traceability:sec:metrics}
We evaluate all experiments along three axes: generation quality, identity novelty, and within-class diversity.
Quality is measured by \FID~\cite{heusel2017gans} (50k samples vs.\ full training set).
Identity novelty is reported via Traceability Rate (TrR) and Novelty Rate NR$(w)$, both computed with the Oracle, and via the source and other-identity match rates $\mathrm{MR}_{\mathrm{src}}$ and $\mathrm{MR}_{\mathrm{oth}}$ of the one-vs-all audit with LCMem, all defined in Sec. 
\ref{traceability:sec:method}.
Diversity is measured by \IRS~\cite{dombrowski_cvpr} (fraction of real training images retrievable as the nearest neighbour of a generated image using SwAV~\cite{caron_unsupervised_2020} features).
Downstream utility follows the TSTR~\cite{esteban2017realvaluedmedicaltimeseries} protocol: a ResNet-50 is trained on synthetic data and evaluated on the real held-out set, with \AUCROC macro-averaged over all $n$ attributes and 5-fold cross-validation stratified by identity.
Full metric definitions, class-conditioning fidelity ($\cos(z_c)$), and paired significance tests are in Appx.~\ref{traceability:sec:app_metrics}.

\noindent\textbf{Baselines.}
\label{traceability:sec:baselines}
Both baselines share the generative architecture and training protocol of our models ($\epsilon_\theta^c$ and $\epsilon_\theta^i$), conditioned via a learnable two-layer MLP that maps the raw binary attribute vector $c \in \{0,1\}^n$ to a $d$-dimensional $\ell_2$-normalised embedding; without $\ell_2$-normalisation the input-dependent magnitude of $c$ interferes with EDM2's activation-magnitude preconditioning, degrading \FID.
EDM2 (class-conditional)~\cite{karras_analyzing_2024} conditions solely on this normalised attribute embedding with no access to identity information.
EDM2 AG (autoguidance)~\cite{karras2024guiding} additionally contrasts the score of the main model against an underfitted guide of the same architecture at inference.
For both baselines we report results at the \FID-optimal guidance scale and training length within the same total compute budget, optimised independently on each dataset.

\noindent\textbf{Datasets and Setup.}
We validate on two datasets.
CelebA~\cite{liu2015deep} has 202,599 images of $m=10{,}177$ identities with $n=40$ binary attributes (upscaled to $512{\times}512$), so $n \ll m$ holds with $\bar{k} \approx 250$ identities per label combination (Proposition~\ref{traceability:prop:novelty}).
MIMIC-CXR~\cite{johnson_mimic-cxr_2019} has chest X-ray studies from $m=48{,}496$ patients (each a unique identity) with $n=14$ binary pathology labels.
We apply the pipeline of Sec.~\ref{traceability:sec:method} to each dataset independently (CelebA choice and MIMIC-CXR encoder ablation: Appx.~\ref{traceability:sec:app_mimic_encoder}).

\subsection{Disentanglement Verification}
\label{traceability:sec:celeba}
Because $E_c$ and $E_i$ are trained independently, we verify each encoder separately
with post-hoc probes before proceeding to diffusion model training.

\noindent\textbf{Metrics.}
We use post-hoc logistic-regression probes on frozen embeddings and identity-disjoint validation data (Appx.~\ref{traceability:sec:app_encoder}).
$E_i$ must show low attribute \AUCROC (target ${\approx}\,0.5$) and high Recall@1 (${\geq}\,0.20$), since a collapsed embedding passes the former but fails the latter; $E_c$ must show high attribute \AUCROC (${\geq}\,0.85$) and \RatOne{} below the attribute-only baseline of $0.205$.
We select $\lambda_{\mathrm{MI},c}{=}50$ for $E_c$, which halves identity leakage (\RatOne{}$_{z_c}$: $0.076{\to}0.041$) for ${\sim}3$\,pp attribute \AUCROC (Appx.~\ref{traceability:par:class_encoder_ablation}).

\noindent\textbf{Identity Encoder Ablation.}
\label{traceability:par:encoder_ablation}
Training a low-dimensional identity encoder that simultaneously achieves strong identity
discriminability and suppresses attribute leakage is non-trivial: adversarial techniques
tend to collapse the identity signal, while weak penalties leave leakage unaddressed.
We compare GRL, CLUB, and bottleneck-only suppression across $d_i \in \{8,16,32,64,128,512\}$;
full setup and results are in Appx.~\ref{traceability:sec:app_id_encoder_ablation} (Table~\ref{traceability:tab:encoder_ablation}).
GRL collapses identity signal at every dimensionality tested (train \RatOne{}\,${<}\,0.01$).
Combining CLUB with $d_i{=}32$ achieves the best suppression among non-collapsed methods
(Val \AUCROC{}\,${=}\,0.640$) while retaining strong discriminability
(Val \RatOne{}\,${=}\,0.718$, Val R@5\,${=}\,0.851$) and is adopted as~$E_i$.

\subsection{Disentangled Generation Validation}
\label{traceability:sec:generation_validation}

\begin{figure*}[t]
\centering
\pgfplotsset{lc/.style={%
  width=0.33\textwidth, height=3.2cm,
  tick label style={font=\tiny},
  label style={font=\scriptsize},
  title style={font=\scriptsize\bfseries, yshift=-2pt},
  grid=major, grid style={dotted, gray!30},
  every axis plot/.append style={line width=0.9pt, mark size=1.6pt},
  xtick align=inside, ytick align=inside, enlarge x limits=0.05,%
}}
\begin{tikzpicture}
\begin{groupplot}[
    group style={group size=3 by 1, horizontal sep=1.65cm},
    lc,
    xlabel={Steps (k)},
]
\nextgroupplot[title={Attribute($\cos(z_c)$)}, ylabel={$\cos(z_c)$},
    legend style={
        at={($(current bounding box.south east)+(+0.95,-0.1cm)$)},
        anchor=north,
        legend columns=3,
        /tikz/every even column/.append style={column sep=0.5cm}
    }]
\addplot[ecblue, solid, mark=*] coordinates {(5.242,0.841582) (10.485,0.849990) (26.214,0.860056) (52.428,0.867337) (78.643,0.868974) (104.857,0.870005) (157.286,0.872809) (188.743,0.873822)};
\addlegendentry{$\epsilon_\theta^c$}
\addplot[emorange, dashed, mark=square*] coordinates {(5.242,0.734279) (10.485,0.747777) (26.214,0.766242) (52.428,0.770764) (78.643,0.774996) (104.857,0.776056) (157.286,0.777741) (188.743,0.777053)};
\addlegendentry{$\epsilon_\theta^i$}
\addlegendimage{gray!60, densely dotted, line width=1.0pt}
\addlegendentry{inter-identity reference}
\nextgroupplot[title={Identity ($\text{Rank-1 retr.}$)}, ylabel={$\text{Rank-1 retr.}$}]
\addplot[ecblue, solid, mark=*, forget plot] coordinates {(5.242,0.000667) (10.485,0.001333) (26.214,0.001333) (52.428,0.006667) (78.643,0.008000) (104.857,0.009667) (157.286,0.011333) (188.743,0.011000)};
\addplot[emorange, dashed, mark=square*, forget plot] coordinates {(5.242,0.064333) (10.485,0.203667) (26.214,0.471667) (52.428,0.562667) (78.643,0.582667) (104.857,0.609333) (157.286,0.632667) (188.743,0.622667)};
\nextgroupplot[title={Identity ($d_{z_i}$)}, ylabel={$d_{z_i}$}]
\addplot[ecblue, solid, mark=*, forget plot] coordinates {(5.242,0.961015) (10.485,0.947002) (26.214,0.920721) (52.428,0.902828) (78.643,0.887061) (104.857,0.876095) (157.286,0.866053) (188.743,0.856612)};
\addplot[emorange, dashed, mark=square*, forget plot] coordinates {(5.242,0.492996) (10.485,0.346333) (26.214,0.237849) (52.428,0.208053) (78.643,0.197322) (104.857,0.191816) (157.286,0.189261) (188.743,0.189498)};
\addplot[gray!60, densely dotted, line width=1.0pt, no markers, forget plot] coordinates {(5.242,0.986571) (188.743,0.986571)};
\end{groupplot}
\end{tikzpicture}
\caption[Learning curves of class- and identity-conditioned models]{%
    Learning curves for $\epsilon_\theta^c$ (\textcolor{ecblue}{solid blue})
    and $\epsilon_\theta^i$ (\textcolor{emorange}{dashed orange}) across training steps.
    (a)~Cosine similarity in $z_c$ space ($\downarrow$ for $\epsilon_\theta^i$, $\uparrow$ for $\epsilon_\theta^c$): how well $\epsilon_\theta^c$
    preserves class conditioning.
    (b)~Rank-1 identity retrieval in $z_i$ space ($\uparrow$ for $\epsilon_\theta^i$, $\downarrow$ for $\epsilon_\theta^c$): how faithfully
    $\epsilon_\theta^i$ reproduces the source identity.
    (c)~Intra-identity cosine distance ($\downarrow$ for $\epsilon_\theta^i$, $\uparrow$ for $\epsilon_\theta^c$) vs.\
    dataset-level inter-identity reference (dotted line); lower intra means
    tighter identity reproduction.
}
\label{traceability:fig:learning_curves}
\end{figure*}
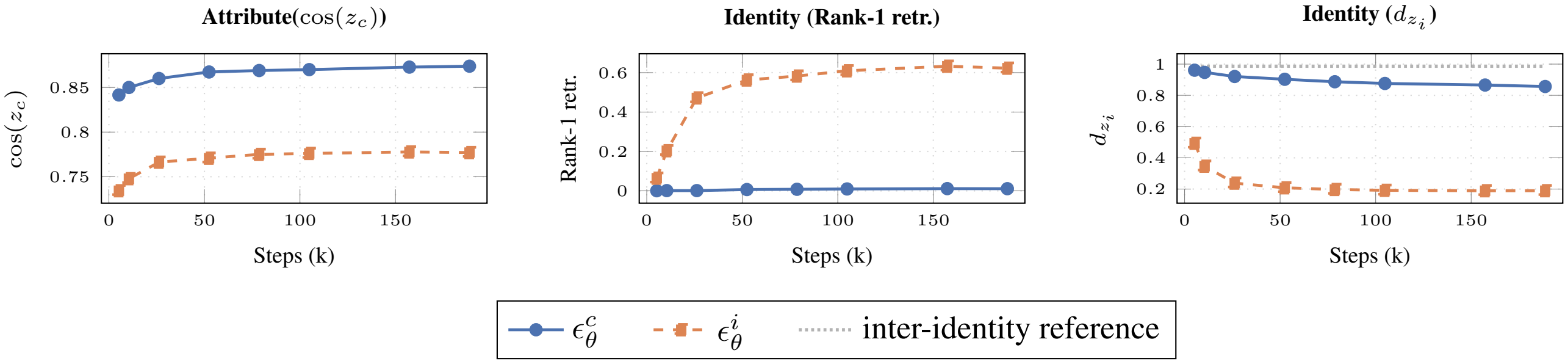

In Fig.~\ref{traceability:fig:learning_curves} and Appx.~\ref{traceability:sec:app_training_dynamics} we investigate the training dynamics of
$\epsilon_\theta^c$ and $\epsilon_\theta^i$.
We measure the attribute and identity consistency of $\epsilon_\theta^c$ and $\epsilon_\theta^i$
by computing the embeddings of the generated images and comparing them to the embeddings of the corresponding training images.
For the attribute encoding, we see that both models improve over time without saturating.
$\epsilon_\theta^c$ is clearly better than $\epsilon_\theta^i$, demonstrating its stronger attribute conditioning.
We also measure how well identity is preserved by checking the rank-1 identity retrieval of the output generations on the training datasets.
The identity model clearly learned this connection, reaching a retrieval rate of above $62\%$.
Notably, this improves over time, indicating that longer training is beneficial for this model.
$\epsilon_\theta^c$, by contrast, fails to retrieve identity entirely, barely reaching $1.1\%$.
We confirm this observation by measuring the intra-sample cosine distance of $z_i$ from both models.
The distance of $\epsilon_\theta^c$ is barely lower than the dataset-level reference,
whereas the distance of $\epsilon_\theta^i$ is low, saturating after roughly $52\,$k steps, earlier than the retrieval metric.
We show visual samples of the seed diversity in Appx.~\ref{traceability:sec:seeddiversity}.

Overall, these experiments confirm that the conditioning and disentanglement were successful.
Removing identity from attributes was easier than vice versa.
We mainly attribute this to the fact that the model can learn confounding attributes from identities
(e.g.\ similar identity implies similar attributes), which we also discussed in the encoder training section.

We also examine generative metrics, specifically \FID and \IRS on $20\,$k generated images.
While \FID saturates after roughly $52\,$k steps and increases slightly with further training,
we choose the last checkpoint as our model, trading a few \FID points in favor of stronger $\epsilon_\theta^c$ and $\epsilon_\theta^i$ attribute alignment.
The full results are given in Appx.~\ref{traceability:sec:app_training_dynamics}.

\subsection{Leaving the Source Identity}
\label{traceability:sec:audit}
We audit the traceable samples of every method one-vs-all with LCMem (Sec.~\ref{traceability:sec:method}, Eq.~\ref{traceability:eq:audit}): each sample is compared with all $122{,}825$ CelebA or $91{,}843$ MIMIC-CXR training images.
All sets were generated with the fixed guidance scale of each method; LCMem was not involved in sampling.
Table~\ref{traceability:tab:audit} reports generation from the inverted latents of the traceable sources; all methods start from the same latents, but only our model is conditioned on the source.
The three rows of our model follow the guidance scale $w$ of Eq.~\ref{traceability:eq:guidance}.

\begin{table}[t]
\centering
\caption[One-vs-all audit against the full training set]{%
    One-vs-all audit of generation from inverted latents with LCMem against the full training set (\%).
    $\mathrm{MR}_{\mathrm{src}}$: samples matching a training image of their source identity.
    $\mathrm{MR}_{\mathrm{oth}}$: samples matching a training image of any other identity.
    Real test images give the false-alarm level of $\mathrm{MR}_{\mathrm{oth}}$.
    $43{,}028$ (CelebA) and $45{,}284$ (MIMIC-CXR) samples per method; 95\% intervals are below $\pm 0.4$.
}
\label{traceability:tab:audit}
\setlength{\tabcolsep}{4pt}
\renewcommand{\arraystretch}{1.2}
\small
\resizebox{\columnwidth}{!}{%
\begin{tabular}{lcccc}
    \toprule
    \rowcolor{fau}
    & \multicolumn{2}{c}{\color{white}\textbf{CelebA}} & \multicolumn{2}{c}{\color{white}\textbf{MIMIC-CXR}} \\
    \rowcolor{fau}
    \color{white}\textbf{Method} & \color{white}$\mathrm{MR}_{\mathrm{src}}$ & \color{white}$\mathrm{MR}_{\mathrm{oth}}$ & \color{white}$\mathrm{MR}_{\mathrm{src}}$ & \color{white}$\mathrm{MR}_{\mathrm{oth}}$ \\
    \midrule
    Real test images (reference) & -- & \AudRefC & -- & \AudRefM \\
    \midrule
    EDM-2 & \AudSrcEdmC & \AudOthEdmC & \AudSrcEdmM & \AudOthEdmM \\
    EDM-2 AG & \AudSrcEdmAgC & \AudOthEdmAgC & \AudSrcEdmAgM & \AudOthEdmAgM \\
    \midrule
    Reconstruction ($\epsilon_\theta^i$, $w{=}{-}1$) & \AudSrcNoAttrC & \AudOthNoAttrC & \AudSrcNoAttrM & \AudOthNoAttrM \\
    No identity guidance ($\epsilon_\theta^c$, $w{=}0$) & \AudSrcNoIdC & \AudOthNoIdC & \AudSrcNoIdM & \AudOthNoIdM \\
    \rowcolor{fau_tech_alpha3}
    \textbf{TraceDiffusion} ($\hat{\epsilon}$, $w{=}1$) & \textbf{\AudSrcOursC} & \AudOthOursC & \textbf{\AudSrcOursM} & \AudOthOursM \\
    \bottomrule
\end{tabular}}
\end{table}

On CelebA the reconstruction ($w{=}{-}1$) matches its source identity in $\AudSrcNoAttrC\%$ of the samples, which confirms that traceable generation starts at the source.
The class-conditioned model alone ($w{=}0$) still matches it in $\AudSrcNoIdC\%$, and memory guidance ($w{=}1$) in only $\AudSrcOursC\%$.
The samples do not move into another training identity instead: $\mathrm{MR}_{\mathrm{oth}}$ is $\AudOthOursC\%$, against $\AudRefC\%$ for real test images, so these matches are at the false-alarm level of the auditor.
Both baselines retain more of the source ($\AudSrcEdmC\%$ and $\AudSrcEdmAgC\%$).

On MIMIC-CXR the source patient is left as reliably ($\AudSrcNoAttrM\%$ to $\AudSrcOursM\%$).
Matches to other training patients ($\mathrm{MR}_{\mathrm{oth}}{=}\AudOthOursM\%$) already arise in the class-conditional model ($\AudOthNoIdM\%$ at $w{=}0$) and equally for untraceable generation, so they originate in $\epsilon_\theta^c$ rather than in the traceable start (Appx.~\ref{traceability:sec:app_audit}, Limitations).
On CelebA, $\mathrm{MR}_{\mathrm{oth}}$ stays at the reference for every guidance scale from $-1$ to $1.5$ (Fig.~\ref{traceability:fig:app_audit_sweep}).

\subsection{Generative Performance}
\label{traceability:sec:generative_and_downstream}
\begin{figure*}[t]
\centering
\includegraphics[width=\linewidth]{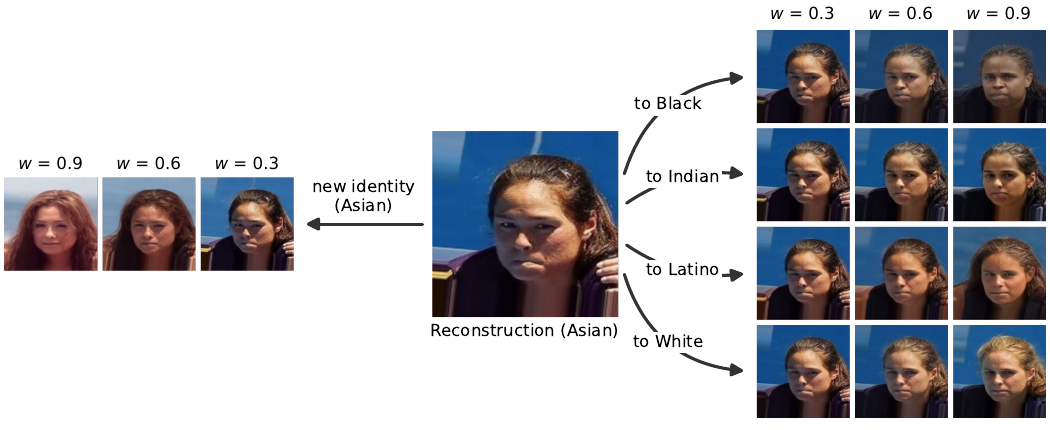}
\caption{%
    \textbf{Counterfactual and identity-reversing generation on CelebA.}
    Because the disentangled pair $(E_c, E_i)$ separates class from identity, the two guidance
    directions can be applied independently.
    \emph{Left:} TraceDiffusion.
    \emph{Right:} Inverse TraceDiffusion. Keep identity but change ethnicity. The identity and context are preserved while the ethnicity moves to the target class.
}
\label{traceability:fig:future_work}
\end{figure*}

\begin{table*}[t]
\centering
\caption[Generative metrics and downstream TSTR performance]{
    Generative metrics and downstream classification performance (TSTR).
    \FID~$\downarrow$: generation quality (50k samples).
    Starting latent: the $\epsilon_\theta^i$-inverted latent of a source image $x_j$, or Gaussian noise.
    Tr.: whether the output is traceable. See \cref{traceability:tab:audit}.
    NR~$\uparrow$: novelty rate, fraction of outputs verified as a novel
    identity by $\reid$ against the source image $x_j$; defined only for
    generation from an inverted latent (source $x_j$ known).
    \IRS~$\uparrow$: image retrieval score~\cite{dombrowski_cvpr}. 
    \AUCROC~$\uparrow$: mean \AUCROC (\%) under TSTR (ResNet-50, mean~$\pm$~std, 5 folds).
    Real data serves as upper bound for \AUCROC only.
    $\dagger$~Methods not designed for identity novelty.
}
\label{traceability:tab:results}
\renewcommand{\arraystretch}{1.2}
\resizebox{\textwidth}{!}{%
\begin{tabular}{l l c cccc cccc cc}
    \toprule
    \rowcolor{fau}
    & & & \multicolumn{8}{c}{\color{white}\textbf{Generative}}
    & \multicolumn{2}{c}{\color{white}\textbf{Downstream (TSTR)}} \\
    \rowcolor{fau}
    & & & \multicolumn{4}{c}{\color{white}\textbf{CelebA}}
    & \multicolumn{4}{c}{\color{white}\textbf{MIMIC-CXR}}
    & \color{white}\textbf{CelebA} & \color{white}\textbf{MIMIC-CXR} \\
    \rowcolor{fau}
    \color{white}\textbf{Method} & \color{white}\textbf{Starting latent} & \color{white}\textbf{Tr.}
        & \color{white}\textbf{\FID}~$\downarrow$ & \color{white}\textbf{NR}~$\uparrow$ & \color{white}\textbf{\IRS}~$\uparrow$ & \color{white}$\cos(z_c)$~$\uparrow$
        & \color{white}\textbf{\FID}~$\downarrow$ & \color{white}\textbf{NR}~$\uparrow$ & \color{white}\textbf{\IRS}~$\uparrow$ & \color{white}$\cos(z_c)$~$\uparrow$
        & \color{white}\textbf{\AUCROC}~$\uparrow$ & \color{white}\textbf{\AUCROC}~$\uparrow$ \\
    \midrule
    Real data (upper bound) & & &
        \multicolumn{8}{c}{---}
        & $\RealAucC$ & $\RealAucM$ \\
    \midrule
    \multirow{2}{*}{EDM-2~\cite{karras_analyzing_2024}$^\dagger$}
        & $\epsilon_\theta^i$-inverted & -- & \EdmTrFidC & \EdmTrNrC & \EdmTrIrsC & \EdmTrCosC & \EdmTrFidM & \EdmTrNrM & \EdmTrIrsM & \EdmTrCosM & $\EdmTrAucC$ & $\EdmTrAucM$ \\
        & Gaussian noise & -- & \EdmUnFidC & \EdmUnNrC & \EdmUnIrsC & \EdmUnCosC & \EdmUnFidM & \EdmUnNrM & \EdmUnIrsM & \EdmUnCosM & $\EdmUnAucC$ & $\EdmUnAucM$ \\[3pt]
    \multirow{2}{*}{EDM-2 AG~\cite{karras_analyzing_2024,karras2024guiding}$^\dagger$}
        & $\epsilon_\theta^i$-inverted & -- & \EdmAgTrFidC & \EdmAgTrNrC & \EdmAgTrIrsC & \EdmAgTrCosC & \EdmAgTrFidM & \EdmAgTrNrM & \EdmAgTrIrsM & \EdmAgTrCosM & $\EdmAgTrAucC$ & $\EdmAgTrAucM$ \\
        & Gaussian noise & -- & \EdmAgUnFidC & \EdmAgUnNrC & \EdmAgUnIrsC & \EdmAgUnCosC & \EdmAgUnFidM & \EdmAgUnNrM & \EdmAgUnIrsM & \EdmAgUnCosM & $\EdmAgUnAucC$ & $\EdmAgUnAucM$ \\
    \midrule
    \textit{Ablations} \\[2pt]
    \multirow{2}{*}{\quad No attribute guidance ($\epsilon_\theta^i, w{=}{-1}$)}
        & $\epsilon_\theta^i$-inverted & -- & $\times$ & $\times$ & $\times$ & $\times$ & $\times$& $\times$ & $\times$ & $\times$ & $\times$ & $\times$ \\
        & Gaussian noise & -- & \NoAttrUnFidC & \NoAttrUnNrC & \NoAttrUnIrsC & \NoAttrUnCosC & \NoAttrUnFidM & \NoAttrUnNrM & \NoAttrUnIrsM & \NoAttrUnCosM & $\NoAttrUnAucC$ & $\NoAttrUnAucM$ \\[4pt]
    \multirow{2}{*}{\quad No identity guidance ($\epsilon_\theta^c, w{=}0$)}
        & $\epsilon_\theta^i$-inverted & \cmark & \NoIdTrFidC & \NoIdTrNrC & \NoIdTrIrsC & \NoIdTrCosC & \NoIdTrFidM & \NoIdTrNrM & \NoIdTrIrsM & \NoIdTrCosM & $\NoIdTrAucC$ & $\NoIdTrAucM$ \\
        & Gaussian noise & -- & \NoIdUnFidC & \NoIdUnNrC & \NoIdUnIrsC & \NoIdUnCosC & \NoIdUnFidM & \NoIdUnNrM & \NoIdUnIrsM & \NoIdUnCosM & $\NoIdUnAucC$ & $\NoIdUnAucM$ \\[3pt]
    \rowcolor{fau_tech_alpha3}
        & $\epsilon_\theta^i$-inverted & \cmark & \OursTrFidC & \OursTrNrC & \OursTrIrsC & \OursTrCosC & \OursTrFidM & \OursTrNrM & \OursTrIrsM & \OursTrCosM & $\OursTrAucC$ & $\OursTrAucM$ \\
    \rowcolor{fau_tech_alpha3}
    \multirow{-2}{*}{\quad \textbf{TraceDiffusion} ($\hat{\epsilon}$, $w{=}1$)}
        & Gaussian noise & -- & \OursUnFidC & \OursUnNrC & \OursUnIrsC & \OursUnCosC & \OursUnFidM & \OursUnNrM & \OursUnIrsM & \OursUnCosM & $\OursUnAucC$ & $\OursUnAucM$ \\
    \bottomrule
\end{tabular}%
}
\end{table*}
Table~\ref{traceability:tab:results} reports generative metrics (\FID, NR, \IRS, $\cos(z_c)$) and downstream \AUCROC for all methods from both starting latents. 
Sampling starts either from the DDIM-inverted latent of a source training image $x_j$ under $\epsilon_\theta^i$ or from random Gaussian noise. 
The starting latent alone does not make a sample traceable: the output must also be generated by a model conditioned on the source, so that it carries an audit trail back to a named $x_j$. 
The EDM-2 baselines have no identity conditioning; we run them from the same inverted latents only as a matched reference and do not count them as traceable. 
Inverting with the converged $\epsilon_\theta^i$ recovers the source identity reliably, with $\mathrm{TrR}{=}\OursTrRC$ on CelebA and $\mathrm{TrR}{=}\OursTrRM$ on MIMIC-CXR; earlier or class-conditioned inverters reach higher TrR (up to $0.962$ for EDM-2~AG and $0.946$ for $\epsilon_\theta^i$ at $5{,}242$~kimg, Tab.~\ref{traceability:tab:app_inversion_trr}), but we adopt the converged $\epsilon_\theta^i$ throughout because it is the only inverter whose latent is disentangled along the identity axis, the methodological precondition that makes memory guidance interpretable as steering along $z_i$ (cf.\ Sec.~\ref{traceability:sec:method} and Appx.~\ref{traceability:sec:app_inversion_model}).
Contrary to our initial expectation that anchoring each sample to a distinct training identity would \emph{increase} inter-class diversity, \IRS decreases from inverted latents for \emph{every} method, so identity-aware generation still needs explicit diversity evaluation. %
Class-attribute fidelity is preserved throughout ($\cos(z_c)\in[0.86, 0.91]$ for all non-degenerate methods). 
At its canonical guidance scale TraceDiffusion produces verified novel identities on $\OursTrNrC$ of CelebA sources and $\OursTrNrM$ of MIMIC-CXR sources. 
On CelebA this exceeds every baseline (EDM-2 $\EdmTrNrC$, EDM-2~AG $\EdmAgTrNrC$).

\noindent\textbf{Downstream utility.}
All generators remain below real data under TSTR (Table~\ref{traceability:tab:results}); the quantity that matters for traceability is the cost of the traceable start relative to the untraceable one.
On CelebA this cost is $\Delta{=}{\PairedOursCelebADelta}$ \AUCROC ($p_t{=}\PairedOursCelebAPt$), five to ten times smaller than for EDM-2 ($\PairedEdmCelebADelta$), EDM-2~AG ($\PairedEdmAgCelebADelta$), and the class-conditional model ($\PairedNoIdCelebADelta$), whose costs are all significant at $p{<}0.01$.
On MIMIC-CXR it is $\Delta{=}{\PairedOursMimicDelta}$ \AUCROC ($p_t{=}\PairedOursMimicPt$), concentrated on chronic structural findings such as Cardiomegaly that are coupled to patient anatomy. 
The paired tests and the per-attribute decomposition are in Appx.~\ref{traceability:sec:app_downstream}.

\noindent\textbf{Counterfactual generation.}
The disentanglement also allows editing the class while keeping the identity, the reverse direction of the previous experiments.
In a preliminary experiment we labelled each CelebA identity with an ethnicity using Qwen3-VL, retrained the encoders and diffusion models, and reversed the guidance (Fig.~\ref{traceability:fig:future_work}).
For 40 edits (8 per target ethnicity), the Qwen3-VL-8B judge perceives the target ethnicity at some guidance strength in 77.5\% of cases; failures concentrate on underrepresented classes (latino 3/8, indian 5/8, versus 8/8 for white and asian).
This suggests the disentanglement carries over to generated images and could support fairness audits of how much ethnicity affects a model's predictions.

\noindent\textbf{Limitations.}
The MIMIC-CXR utility gap is explained by identity-anatomy entanglement (Appx.~\ref{traceability:sec:per_attr}) but is not zero; retraining $E_i$ with patient-stratified hard negatives from anatomy-similar patients is a natural mitigation.
On MIMIC-CXR $\AudOthOursM\%$ of guided samples match another training patient, against $\AudRefM\%$ for real test images.
The class-conditioned model alone already shows this overlap ($\AudOthNoIdM\%$), so it originates in $\epsilon_\theta^c$ and is increased by guidance; applying the audit during sampling and re-steering flagged samples is the direct remedy.
Traceability provides \emph{accountability} but no formal guarantee, and depends on the trustworthiness of the Oracle and LCMem.

\section{Conclusion}
\label{traceability:sec:conclusion}
We introduced TraceDiffusion, a latent diffusion framework that
steers generation away from a named training identity while preserving
class consistency, and certifies novelty per sample via a
re-identification oracle. 
Every generated image carries a verifiable claim about which training identity it does not reproduce.
An independent audit of every sample against the full training set supports this claim on CelebA: the share of samples matching their source identity falls from $\AudSrcNoAttrC\%$ for the reconstruction to $\AudSrcOursC\%$ under guidance, and matches to other training identities stay at the false-alarm level of real images.
Traceability costs only $\Delta{=}0.12$\,pp \AUCROC downstream on CelebA, five to ten times less than the baselines, and $\Delta{=}3.78$ on MIMIC-CXR, localised to chronic structural findings: identity-pathology entanglement, not the method, is the limiting factor.

\section*{LLM Usage Statement}
We used Anthropic's Claude models, through the Claude Code coding assistant, throughout this work. Affected parts: (i) the code of the experiments and evaluation pipeline (writing, refactoring and debugging); (ii) the plotting code of figures and tables; and (iii) the text of the paper, where the LLM drafted and edited passages and improved grammar and wording in all sections. The research ideas, the method, the experimental design and the interpretation of results are the authors'. The authors reviewed all generated code, results and text, verified all citations, and take full responsibility for the paper.

\section*{Acknowledgments}
HPC resources were provided by the Erlangen National High Performance Computing Center (NHR@FAU), under the NHR projects b180dc and y100dc. NHR is funded by federal and Bavarian state authorities, and NHR@FAU hardware is partially funded by the DFG, 440719683.
The authors received funding from the ERC-project MIA-NORMAL 101083647, 101326871, DFG 513220538, 512819079, and through the Hightech Agenda Bayern (HTA) of the Free State of Bavaria, Germany. This work was supported through the Bavarian AI Foundation Model Initiative, funded by the Bavarian State Ministry of Science and the Arts (StMWK).

{
    \small
    \bibliographystyle{ieeenat_fullname}
    \bibliography{references}
}

\clearpage
\appendix
\section{Appendix}

\subsection{Evaluation Metrics}
\label{traceability:sec:app_metrics}

We evaluate all experiments along three axes: generation quality, identity
novelty, and within-class diversity.

\paragraph{Fr\'echet Inception Distance (FID).}
\FID~\cite{heusel2017gans} measures the distributional distance between
real and generated images in the feature space of a pretrained Inception-v3
network. 
Lower \FID indicates higher overall generation quality and better
coverage of the real data distribution. 
We report \FID computed over 50k
generated samples against the full training set.

\paragraph{Traceability Rate (TrR) and Novelty Rate (NR).}
Both metrics are defined in Section~\ref{traceability:sec:method} as they are intrinsic to the pipeline.
In brief, $\mathrm{TrR}$ measures the fraction of source images that the identity denoiser
$\epsilon_\theta^i$ faithfully reconstructs under DDIM inversion; $\mathrm{NR}(w)$
(Eq.~\ref{traceability:eq:nr}) measures the fraction of those traceable images for which memory guidance
at scale $w$ produces a generation verified as novel against the source and its $k{=}5$
nearest identity neighbours.

\paragraph{Image Retrieval Score (IRS).}
\IRS~\cite{dombrowski_cvpr} measures diversity by asking what fraction of
real training images can be retrieved as the nearest neighbour of at least
one generated image in feature space. 
Intuitively, a model that covers the
training distribution well retrieves many real images; a model that
collapses to a single mode retrieves very few. 
We report
\IRS as a function of $w$ to detect guidance collapse
within $\mathcal{D}_{c^*}$. 
Following~\cite{dombrowski_cvpr}, we use
self-supervised SwAV~\cite{caron_unsupervised_2020} features as the embedding space.

\paragraph{Downstream Utility (AUC).}
To evaluate whether synthetic images preserve label-discriminative structure,
we adopt the \emph{train-on-synthetic, test-on-real} (TSTR)~\cite{esteban2017realvaluedmedicaltimeseries} protocol.
A ResNet-50 classifier with a multi-label sigmoid head is trained exclusively on
the synthetic training set and evaluated on the real held-out test set.
We use 5-fold cross-validation with folds defined by patient or subject identity
(all images of one identity remain in the same fold), so that the reported variance
reflects genuine distributional uncertainty rather than seed sensitivity.
The \AUCROC metric is macro-averaged over all $n$ binary attributes; we report
mean~$\pm$~std across folds.
Higher \AUCROC indicates that the synthetic data preserves label-discriminative signal;
\textbf{Real data} (trained on real images) serves as the upper bound.
We further evaluate two operational \emph{modes} of generation: \textbf{traceable}
(sampling initialised from the DDIM-inverted noise of source $x_j$, anchoring each
output to a named training identity) and \textbf{untraceable} (sampling initialised
from random Gaussian noise, breaking the audit trail).
Both modes produce synthetic datasets of the same size; their per-fold \AUCROC{}s are
compared in pairs.

\paragraph{Class-conditioning fidelity ($\cos(z_c)$).}
To verify that guidance does not corrupt the target attribute profile, we measure
$\cos(z_c)\,{=}\,\cos\bigl(E_c(\hat{x}),\, z_c^{(j)}\bigr)$, the cosine similarity
between the class embedding of the generated image $\hat{x}$ and the class embedding
of the source $x_j$. 
Higher is better; values close to $1$ indicate that the generated
sample retains the attribute composition of the source. 
Class-encoding collapse, e.g.\
when the no-attribute-guidance ablation drifts off the source attribute profile, manifests
as a sharp drop on this metric while leaving \FID and \IRS only mildly affected.

\paragraph{Paired significance tests.}
The two operational modes share fold splits, so traceable and untraceable \AUCROC{}s are
naturally paired: $(\bar{x}_{\mathrm{Tr}}^{(f)}, \bar{x}_{\mathrm{Un}}^{(f)})$ for
each fold $f \in \{1,\dots,5\}$.
For every method we report the per-fold mean difference
$\Delta = \bar{x}_{\mathrm{Tr}} - \bar{x}_{\mathrm{Un}}$, a one-sample paired
$t$-test on the differences ($p_t$), a Wilcoxon signed-rank test ($p_W$), and a
$\mathrm{BCa}$-bootstrapped 95\% confidence interval on $\Delta$ over $10^4$
resamples. 
With $n{=}5$ folds, the Wilcoxon test has a hard floor at $p_W{=}0.0625$
(achieved whenever all fold-wise signs agree), so we treat the $t$-test plus the
bootstrap CI as the primary reporters and use Wilcoxon as a sign-agreement check
only.
The interpretive question is not ``does each method differ from zero in absolute
\AUCROC'', but ``\emph{which methods are statistically indistinguishable across the two
operational modes}'', because a method whose paired test cannot reject the null is one
whose traceability comes effectively for free.

\subsection{Nearest-Neighbor Retrieval: Memorization Check}
\label{traceability:app:nearest_retrieval}

\begin{figure}[h]
    \centering
    \includegraphics[width=\linewidth]{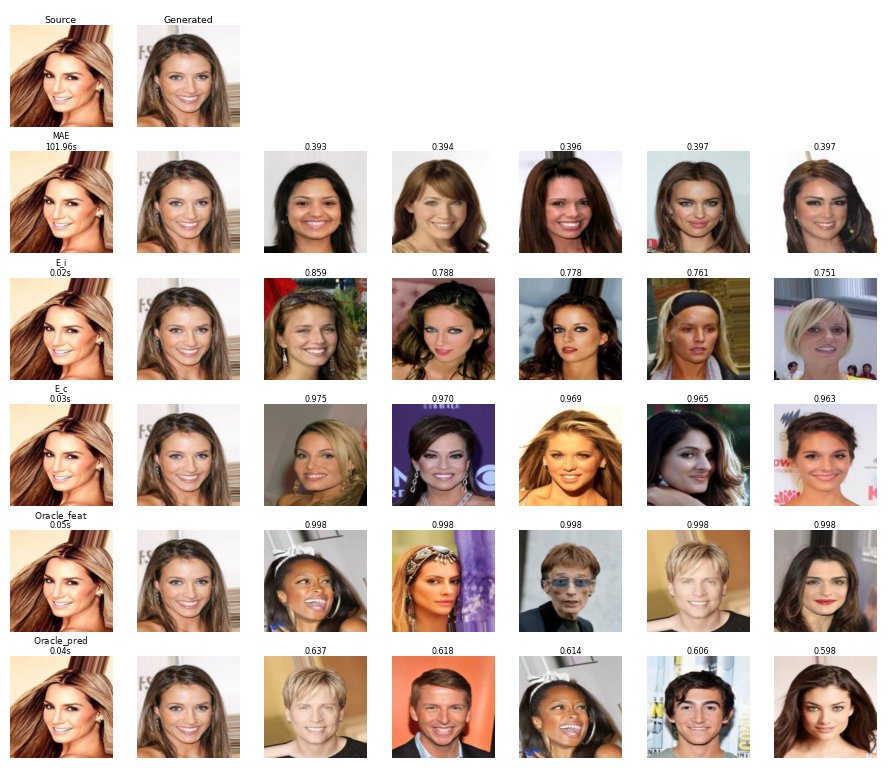}
    \caption[Nearest-neighbor retrieval for a class-conditioned sample]{%
        Nearest-neighbor retrieval for one representative \textbf{class-conditioned} ($\epsilon_\theta^c$) generated image.
        Each row shows the five closest training images found by one of the five retrieval
        metrics (MAE, $E_i$, $E_c$, Oracle features, Oracle prediction).
        The low visual similarity across all metrics indicates the model does not memorize training data.
        The number on top of the source images indicate how long retrieval takes (retrieve from entire train set).  
    }
    \label{traceability:fig:nearest_retrieval_c}
\end{figure}

\begin{figure}[h]
    \centering
    \includegraphics[width=\linewidth]{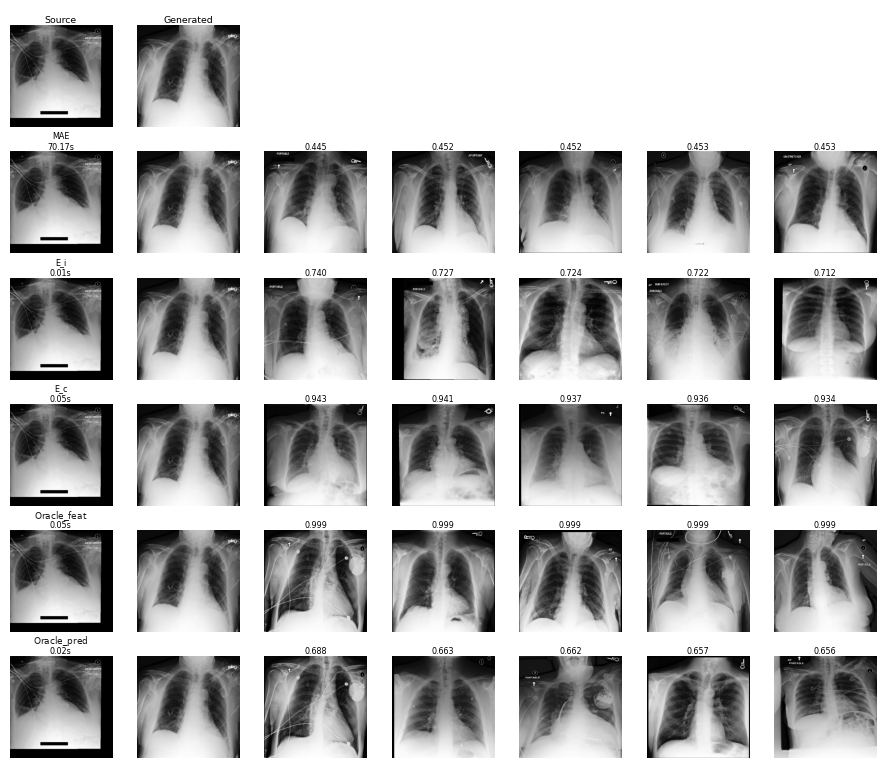}
    \caption[Class-conditioned nearest-neighbor retrieval on MIMIC-CXR]{%
        Nearest-neighbor retrieval for one representative \textbf{class-conditioned} ($\epsilon_\theta^c$) generated image on MIMIC-CXR.
    }
    \label{traceability:fig:nearest_retrieval_c_mimic}
\end{figure}

\begin{figure}[h]
    \centering
    \includegraphics[width=\linewidth]{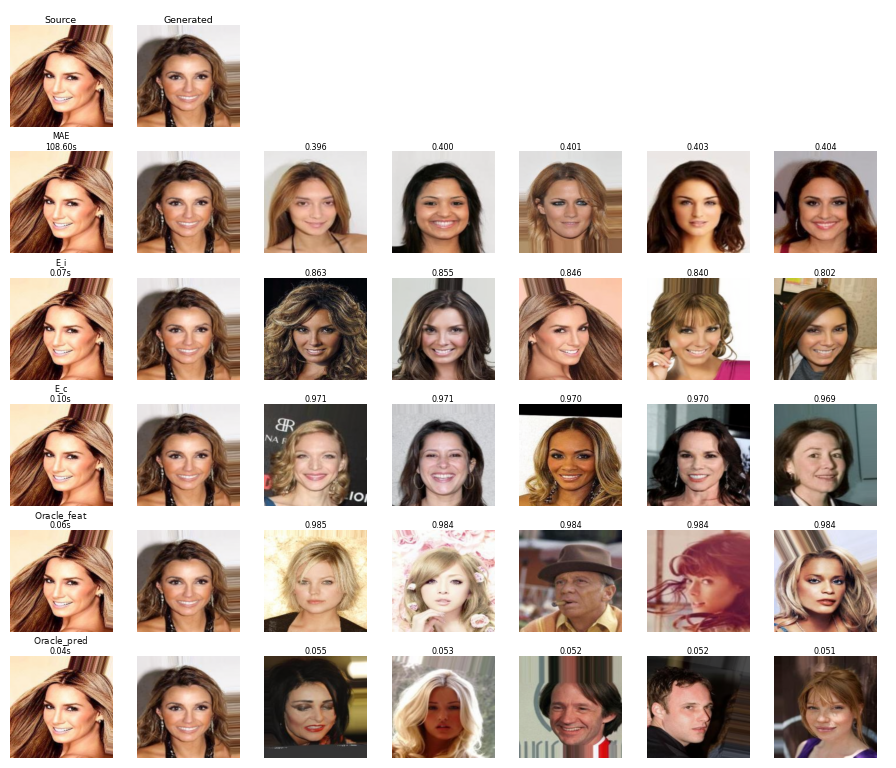}
    \caption[Nearest-neighbor retrieval for an identity-conditioned sample]{%
        Nearest-neighbor retrieval for one representative \textbf{identity-conditioned} ($\epsilon_\theta^i$) generated image.
        Same five retrieval metrics as \cref{traceability:fig:nearest_retrieval_c}.
        The low visual similarity confirms $\epsilon_\theta^i$ also generalizes rather than memorizes.
    }
    \label{traceability:fig:nearest_retrieval_i}
\end{figure}

To verify that neither $\epsilon_\theta^c$ nor $\epsilon_\theta^i$ exclusively memorize training images,
we selected five source images from the CelebA training set, generated one image per source, and
retrieved the five closest training images for each generated output using five independent metrics:
(1)~pixel-space MAE, (2)~cosine similarity on cached identity embeddings $z_i$,
(3)~cosine similarity on cached class embeddings $z_c$, (4)~cosine similarity on Oracle backbone
features, and (5)~the full Oracle similarity score $\sigma(\mathrm{fc}(|f_{\mathrm{gen}} - f_{\mathrm{train}}|))$.
We visually verified that across all five metrics and all five sources, no generated image was a near-duplicate of any
training image, confirming that both models are able to generalize rather than only reproduce seen samples.
Figs.~\ref{traceability:fig:nearest_retrieval_c} (CelebA, $\epsilon_\theta^c$), \ref{traceability:fig:nearest_retrieval_c_mimic} (MIMIC-CXR, $\epsilon_\theta^c$),
and~\ref{traceability:fig:nearest_retrieval_i} (CelebA, $\epsilon_\theta^i$) show one representative source per case.

\subsection{Why Disentanglement Is Necessary}
\label{traceability:sec:app_disentanglement_necessity}

Disentanglement is not merely a quality improvement; it is a structural requirement for the entire pipeline to function as claimed.
We identify three failure modes that arise if $I(z_i;\,c) > 0$, i.e.\ if the identity embedding retains residual label information.

\paragraph{(i) Guidance corruption.}
The guidance formula (Eq.~\ref{traceability:eq:guidance}) subtracts the memory score from the class score at every denoising step.
For this subtraction to steer generation \emph{only} in the identity direction, the two scores must capture orthogonal modes of variation.
If $z_i$ leaks label information, $\epsilon_\theta^i$ also models class-level variation, so the difference
$\epsilon_\theta^c(x_t, z_c^{(j)}) - \epsilon_\theta^i(x_t, z_i^{(j)})$
contains a class component: guidance then simultaneously steers away from the source identity \emph{and} away from the target label cell.
At scale $w > 0$ this degrades both novelty and class consistency, with the degradation compounding as $w$ increases.

\paragraph{(ii) Invalidation of the novelty existence proof.}
Proposition~\ref{traceability:prop:novelty} reasons about a pure identity subspace $\mathcal{Z}_i$ of dimension $d_i$.
If $z_i$ encodes label information, then $\mathcal{Z}_i$ is not a pure identity manifold: the $\delta$-ball packing argument holds over a mixed latent space where the ``uncovered'' region may correspond to novel label combinations rather than novel identities.
The novelty guarantee then loses its semantic content.

\paragraph{(iii) Opacity of the ReID verification.}
$\reid$ is calibrated to measure identity novelty.
If $z_i$ encodes class attributes, a generated image might pass the ReID test by differing in a visible attribute (e.g.\ hair color) rather than in identity, yielding a false positive novelty claim.
Disentanglement is therefore a pre-condition for $\reid$ to serve as a valid novelty certificate.

The symmetric condition, namely that $z_c$ does not leak identity information, is equally important: if $\epsilon_\theta^c$ absorbs identity variation, the class-conditioned model already varies identity at $w=0$, and the guidance scale $w$ loses its role as a controllable novelty knob.
Both directions of leakage are validated empirically: we require post-hoc \AUCROC on $z_i \approx 0.5$ across all attributes \emph{and} near-chance identity retrieval from $z_c$ before proceeding to diffusion model training (Section~\ref{traceability:sec:celeba}).

\subsection{Disentangled Encoder: Architecture and Training}
\label{traceability:sec:app_encoder}

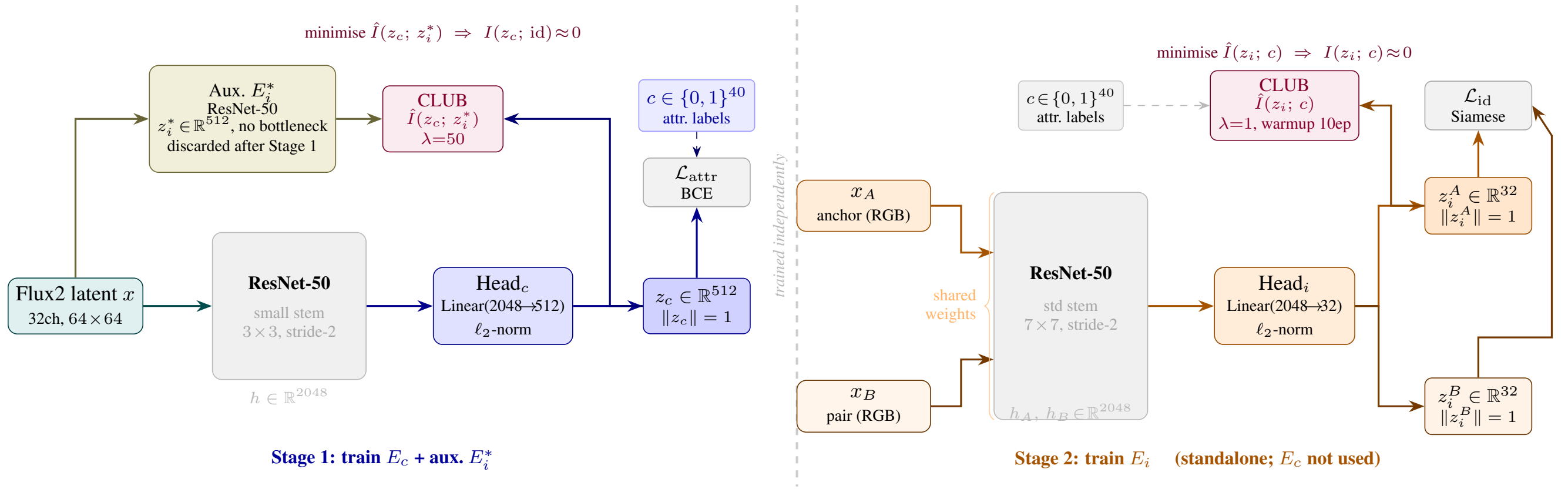
\begin{figure*}[t]
\centering
\resizebox{\textwidth}{!}{%
\begin{tikzpicture}[
    >=Stealth,
    font=\small,
    block/.style={
        rectangle, draw, rounded corners=4pt,
        fill=#1, minimum width=2.0cm, minimum height=0.75cm,
        align=center
    },
    block/.default=white,
    loss/.style={
        rectangle, draw=gray!70, rounded corners=3pt,
        fill=gray!10, minimum width=1.6cm, minimum height=0.65cm,
        align=center, font=\footnotesize
    },
    label_node/.style={
        rectangle, draw=blue!40, rounded corners=2pt,
        fill=blue!8, minimum width=1.2cm, minimum height=0.5cm,
        align=center, font=\footnotesize, text=blue!60!black
    },
    emb/.style={
        rectangle, draw, rounded corners=3pt,
        fill=#1, minimum width=1.6cm, minimum height=0.65cm,
        align=center, font=\footnotesize
    },
    emb/.default=white,
    club/.style={
        rectangle, draw=purple!60!black, rounded corners=3pt,
        fill=purple!8, minimum width=1.8cm, minimum height=0.85cm,
        align=center, font=\footnotesize, text=purple!70!black
    },
    backbone/.style={
        rectangle, draw=gray!55, rounded corners=4pt,
        fill=gray!12, minimum width=2.3cm, minimum height=2.2cm,
        align=center
    },
]

\node[block=teal!12, draw=teal!50!black] (lat) at (0, 0)
    {Flux2 latent $x$\\[-1pt]\scriptsize 32ch, $64\!\times\!64$};

\node[backbone] (backboneC) at (3.2, 0) {};
\node[font=\footnotesize\bfseries, align=center] at (3.2, 0.35)
    {ResNet-50};
\node[font=\scriptsize, align=center, gray!70] at (3.2, -0.25)
    {small stem\\$3\!\times\!3$, stride-2};
\node[font=\scriptsize, gray!55] at (3.2, -1.35)
    {$h \in \mathbb{R}^{2048}$};

\draw[->, thick, teal!60!black] (lat.east) -- (backboneC.west);

\node[block=blue!12, draw=blue!50!black] (headC) at (6.4, 0)
    {Head$_c$\\[-2pt]\scriptsize Linear(2048$\!\to\!$512)\\[-1pt]\scriptsize $\ell_2$-norm};

\draw[->, thick, blue!55!black] (backboneC.east) -- (headC.west);

\node[emb=blue!18, draw=blue!55!black] (zc) at (9.3, 0)
    {$z_c \in \mathbb{R}^{512}$\\[-1pt]\scriptsize $\|z_c\|=1$};

\draw[->, thick, blue!55!black] (headC.east) -- (zc.west);

\node[label_node] (attrsC) at (9.3, 3.0)
    {$c\in\{0,1\}^{40}$\\[-1pt]\scriptsize attr.\ labels};
\node[loss] (lossAttr) at (9.3, 1.85)
    {$\mathcal{L}_{\mathrm{attr}}$\\[-1pt]\scriptsize BCE};

\draw[->, dashed, blue!40!black] (attrsC.south) -- (lossAttr.north);
\draw[->, thick, blue!55!black]  (zc.north) -- (lossAttr.south);

\node[club] (clubC) at (5.5, 2.8)
    {CLUB\\[-2pt]\scriptsize $\hat{I}(z_c;\,z_i^*)$\\[-1pt]\scriptsize $\lambda{=}50$};
\node[block=olive!12, draw=olive!60!black, minimum width=2.3cm, minimum height=1.6cm,
      align=center, font=\footnotesize] (auxEi) at (2.5, 2.8)
    {Aux.\ $E_i^*$\\[-2pt]\scriptsize ResNet-50\\[-2pt]\scriptsize $z_i^*\!\in\!\mathbb{R}^{512}$, no bottleneck\\[-1pt]\scriptsize discarded after Stage~1};

\draw[->, thick, blue!45!black]  (zc.west) -- (8.0, 0) -- (8.0, 2.8) -- (clubC.east);
\draw[->, thick, olive!55!black] (lat.north) |- (auxEi.west);
\draw[->, thick, olive!55!black] (auxEi.east) -- (clubC.west);
\node[font=\scriptsize, purple!55!black] at (5.5, 4.1)
    {minimise $\hat{I}(z_c;\,z_i^*)\;\Rightarrow\;I(z_c;\,\idl)\!\approx\!0$};

\node[font=\footnotesize\bfseries, blue!60!black] at (4.6, -2.3)
    {Stage 1: train $E_c$ + aux.\ $E_i^*$};

\draw[dashed, gray!40, line width=1pt] (10.8, 4.5) -- (10.8, -2.7);
\node[font=\scriptsize\itshape, gray!55, rotate=90] at (10.55, 1.2)
    {trained independently};

\node[block=orange!15, draw=orange!60!black] (imgA) at (11.8, 1.5)
    {$x_A$\\[-1pt]\scriptsize anchor (RGB)};
\node[block=orange!8, draw=orange!40!black]  (imgB) at (11.8, -1.5)
    {$x_B$\\[-1pt]\scriptsize pair (RGB)};

\node[backbone, minimum height=3.4cm] (backboneI) at (14.9, 0) {};
\node[font=\footnotesize\bfseries, align=center] at (14.9, 0.5)
    {ResNet-50};
\node[font=\scriptsize, align=center, gray!70] at (14.9, -0.15)
    {std stem\\$7\!\times\!7$, stride-2};
\node[font=\scriptsize, gray!55] at (14.9, -1.6)
    {$h_A,\,h_B\!\in\!\mathbb{R}^{2048}$};

\draw[->, thick, orange!65!black] (imgA.east) -- ++(0.45,0) |- (backboneI.west |- {(0,0.8)});
\draw[->, thick, orange!45!black] (imgB.east) -- ++(0.45,0) |- (backboneI.west |- {(0,-0.8)});

\draw[decorate, decoration={brace, mirror, amplitude=4pt}, orange!40]
    (14.9-1.15, 1.7) -- (14.9-1.15, -1.7)
    node[midway, left=4pt, font=\scriptsize, orange!60, align=right]
    {shared\\weights};

\node[block=orange!15, draw=orange!60!black, minimum width=2.1cm] (headI) at (18.1, 0)
    {Head$_i$\\[-2pt]\scriptsize Linear(2048$\!\to\!$32)\\[-1pt]\scriptsize $\ell_2$-norm};

\draw[->, thick, orange!65!black] (backboneI.east) -- (headI.west);

\node[emb=orange!20, draw=orange!60!black] (ziA) at (21.0, 1.5)
    {$z_i^A \in \mathbb{R}^{32}$\\[-1pt]\scriptsize $\|z_i^A\|=1$};
\node[emb=orange!10, draw=orange!40!black] (ziB) at (21.0, -1.5)
    {$z_i^B \in \mathbb{R}^{32}$\\[-1pt]\scriptsize $\|z_i^B\|=1$};

\draw[->, thick, orange!65!black] (headI.east) -- ++(0.35,0) |- (ziA.west);
\draw[->, thick, orange!45!black] (headI.east) -- ++(0.35,0) |- (ziB.west);

\node[loss] (lossSiam) at (21.0, 3.0)
    {$\mathcal{L}_{\mathrm{id}}$\\[-1pt]\scriptsize Siamese};

\draw[->, thick, orange!65!black] (ziA.north) -- (lossSiam.south);
\draw[->, thick, orange!45!black] (ziB.north) -- ++(0, 0.6) -- ++(1.1, 0)
    -- (21.0+1.1, 2.55) -- (lossSiam.east);

\node[club] (clubI) at (18.1, 3.0)
    {CLUB\\[-2pt]\scriptsize $\hat{I}(z_i;\,c)$\\[-1pt]\scriptsize $\lambda{=}1$, warmup 10ep};
\node[block=gray!10, draw=gray!45, rounded corners=2pt,
      minimum width=1.3cm, minimum height=0.5cm,
      align=center, font=\scriptsize] (attrsI) at (14.9, 3.0)
    {$c\!\in\!\{0,1\}^{40}$\\attr.\ labels};

\draw[->, thick, orange!55!black] (ziA.west) -- ++(-0.5,0) |- (clubI.east);
\draw[->, dashed, gray!55] (attrsI.east) -- (clubI.west);
\node[font=\scriptsize, purple!55!black] at (18.1, 3.8)
    {minimise $\hat{I}(z_i;\,c)\;\Rightarrow\;I(z_i;\,c)\!\approx\!0$};

\node[font=\footnotesize\bfseries, orange!65!black, align=center] at (16.8, -2.3)
    {Stage 2: train $E_i$ \quad(standalone; $E_c$ not used)};

\end{tikzpicture}
}%
\caption[Disentangled encoder architecture (split backbone)]{
    \textbf{Disentangled encoder architecture (split backbone).}
    $E_c$ and $E_i$ are \emph{separate} ResNet-50 networks with no shared parameters,
    trained in two independent stages.
    \textbf{Stage~1} trains the \emph{class encoder} $E_c$ (blue):
    a Flux2 latent $x \in \mathbb{R}^{32\times64\times64}$ is processed by a
    ResNet-50 with a reduced $3\times3$ stem (suited to $64\times64$ inputs)
    and projected to $z_c \in \mathbb{R}^{512}$ ($\ell_2$-normalised).
    $E_c$ is supervised by a multi-label BCE loss on 40 binary attribute labels
    ($\mathcal{L}_{\mathrm{attr}}$) and a CLUB mutual-information penalty that
    pushes identity information out of $z_c$
    ($\hat{I}(z_c;\,z_i^*)\!\to\!0$, $\lambda{=}50$),
    verified post-hoc by identity recall@1 $= 0.041$.
    The CLUB term requires an identity signal: an auxiliary encoder $E_i^*$ (olive)
    is trained jointly with $E_c$, sharing the same Siamese contrastive objective as
    the final $E_i$ but without the 16-dimensional bottleneck
    ($z_i^* \in \mathbb{R}^{512}$); it is discarded once Stage~1 completes.
    \textbf{Stage~2} trains the \emph{identity encoder} $E_i$ (orange) independently and without any involvement of $E_c$:
    a Siamese pair $(x_A, x_B)$ of same-identity RGB images is
    encoded by a shared-weight ResNet-50 (standard $7\times7$ stem) and
    projected through a 32-dimensional bottleneck head to $z_i \in \mathbb{R}^{32}$.
    A Siamese contrastive loss ($\mathcal{L}_{\mathrm{id}}$) pulls same-identity
    embeddings together; a CLUB penalty with staged warmup (10 epochs) suppresses
    attribute leakage ($\hat{I}(z_i;\,c)\!\to\!0$, $\lambda{=}1$).
    The bottleneck capacity ($d{=}32$) prevents the CLUB estimator collapse
    observed at $d{=}512$.
}
\label{traceability:fig:encoder}
\end{figure*}

\paragraph{Encoder structure.}
The full encoder pipeline is depicted in Fig.~\ref{traceability:fig:encoder}.
We use two encoders with independent parameters, each mapping directly from image space:
\begin{equation}
    E_c : \mathbb{R}^{\imgdim} \rightarrow \mathcal{Z}_c \subset \mathbb{R}^{d_c}, \qquad
    E_i : \mathbb{R}^{\imgdim} \rightarrow \mathcal{Z}_i \subset \mathbb{R}^{d_i},
\end{equation}
yielding $z_c = E_c(x)$ and $z_i = E_i(x)$. 
The full latent representation is $z = (z_c, z_i) \in \mathcal{Z}_c \oplus \mathcal{Z}_i = \mathcal{Z}$.

The vector $z_c$ is a continuous label embedding, trained via a multi-label binary cross-entropy loss against the class labels $c$:
\begin{equation}
    \mathcal{L}_c = \mathrm{BCE}(h_{\mathrm{label}}(z_c),\, c).
\end{equation}
The vector $z_i$ is an identity embedding, trained with a Siamese contrastive loss $\mathcal{L}_{\mathrm{id}}$ to be discriminative over identities.

\paragraph{Adversarial Disentanglement.}
To ensure $z_i$ carries no label information, $E_i$ is trained adversarially against a label discriminator $A_c$ via the minimax objective:
\begin{equation}
    \min_{E_i} \max_{A_c}\ \mathbb{E}_{x \sim \mathcal{D}}\left[\log A_c(E_i(x))\right].
\end{equation}
The full training objective for the encoder is:
\begin{equation}
    \mathcal{L} = \mathcal{L}_{\mathrm{attr}} + \mathcal{L}_{\mathrm{id}} + \lambda\, \mathcal{L}_{\mathrm{adv}},
\end{equation}
where $\lambda$ controls the strength of the disentanglement penalty. 
At the minimax equilibrium~\cite{goodfellow2014generative,ganin2016domainadversarial}, $A_c$ cannot perform better than random guessing on $z_i$, implying $I(z_i;\, c) = 0$. In practice adversaries only approximate this. 
In practice we have $I(z_i;\, c) \leq \epsilon$ for a small empirical $\epsilon$, validated by measuring the \AUCROC of a post-hoc attribute classifier trained on $z_i$, which should approach $0.5$ across all attributes.

\paragraph{Identity encoder $E_i$.}
A ResNet-50 backbone modified with a $3{\times}3$, stride-2 stem (no max-pool) for compatibility with Flux2 latent resolution maps each input to a 2048-d feature, projected to $z_i \in \mathbb{R}^{32}$, $\ell_2$-normalised.
The 32-dimensional bottleneck physically limits the capacity of $z_i$ to encode 40 independent binary attributes while still supporting identity discrimination (see ablation, Table~\ref{traceability:tab:encoder_ablation}).
$E_i$ is trained with the Siamese contrastive loss of the Oracle (Sec.~\ref{traceability:sec:app_lcmem}) on paired samples $(x_A, x_B)$ from the same identity:
\begin{equation}
    \mathcal{L}_{\mathrm{id}}
    = 0.8\,\mathcal{L}_{\mathrm{fc}}
    + 0.2\,\mathcal{L}_{\mathrm{NT\text{-}Xent}},
    \label{traceability:eq:enc_i_id_loss}
\end{equation}
where $\mathcal{L}_{\mathrm{fc}}$ is BCE on $|z_i^A - z_i^B|$ via a learned comparison head and
$\mathcal{L}_{\mathrm{NT\text{-}Xent}}$~\cite{chen2020simpleframework} uses temperature $\tau{=}0.07$.
To suppress attribute leakage, a CLUB penalty minimizes $I(z_i;\, c)$ with a staged warm-up ramping $\lambda_{\mathrm{MI}}$ from 0 to 1 over 10 epochs, allowing the Siamese objective to first establish a stable identity manifold:
\begin{equation}
    \mathcal{L}_i = \mathcal{L}_{\mathrm{id}}
                  + \lambda_{\mathrm{MI}}(t)\,\mathcal{L}_{\mathrm{CLUB}}(z_i,\, c).
    \label{traceability:eq:enc_i_loss}
\end{equation}

\paragraph{Class encoder $E_c$.}
A ResNet-50 backbone projects to $z_c \in \mathbb{R}^{512}$, $\ell_2$-normalised, trained with multi-label BCE over all $n$ class attributes and a CLUB penalty minimizing $I(z_c;\, \idl)$ with weight $\lambda_{\mathrm{MI},c}{=}0.1$:
\begin{equation}
    \mathcal{L}_c = \mathcal{L}_{\mathrm{BCE}}(z_c,\, c)
                  + \lambda_{\mathrm{MI},c}\,\mathcal{L}_{\mathrm{CLUB}}(z_c,\, \idl).
    \label{traceability:eq:enc_c_loss}
\end{equation}
The CLUB term requires an identity signal during $E_c$ training; we provide it by jointly training an auxiliary identity encoder with the same architecture as $E_i$ but projecting to $\mathbb{R}^{512}$ (no bottleneck), discarded after $E_c$ training completes.

\paragraph{Optimiser.}
Both encoders use AdamW~\cite{kingmaadammethod,loshchilov2018decoupled} ($\beta_1{=}0.9$, $\beta_2{=}0.999$, weight decay $10^{-4}$) with cosine learning-rate decay, a one-epoch linear warm-up, and per-group rates (backbone $3{\times}10^{-4}$, head $3{\times}10^{-3}$) for 30 epochs each.
After training, $z_c$ and $z_i$ are extracted and cached for all training images; these cached embeddings serve as conditioning signals for $\epsilon_\theta^c$ and $\epsilon_\theta^i$.

\subsection{Diffusion Model Training and Sampling}
\label{traceability:sec:app_diffusion}

\paragraph{Architecture.}
Both $\epsilon_\theta^c$ and $\epsilon_\theta^i$ use the EDM2~\cite{karras_analyzing_2024} U-Net, which preserves the magnitudes of activations, weights, and updates throughout training via careful preconditioning, yielding improved training dynamics.
The U-Net is conditioned on the respective encoder embedding ($z_c$ or $z_i$) via cross-attention at each resolution level.
We fine-tune from a pretrained ImageNet EDM2-M checkpoint (32-channel Flux2 latent space).

\paragraph{Training.}
Both models are trained with the EDM2 preconditioning loss:
\begin{equation}
    \mathcal{L}_{\mathrm{diff}} = \mathbb{E}_{x,\sigma,\epsilon}
    \left[ \lambda(\sigma)
    \left\| D_\theta(x + \sigma\epsilon;\, \sigma,\, z) - x \right\|^2
    \right],
\end{equation}
where $D_\theta$ is the preconditioned denoiser, $\lambda(\sigma)$ is the EDM2 loss weighting, and $z \in \{z_c, z_i\}$.
The linear noise schedule $\sigma(t){=}t$ with log-normal timestep sampling is used throughout.
Optimisation uses Adam ($\beta_1{=}0.9$, $\beta_2{=}0.99$, lr ${\approx}10^{-3}$) with EMA weight averaging.

\paragraph{Guided Sampling.}
At inference, a source image $x_j$ is selected from the training set; both $z_c^{(j)} {=} E_c(x_j)$ and $z_i^{(j)} {=} E_i(x_j)$ are extracted from that same image, so the output retains the attribute profile of $x_j$ while being steered away from its identity.
The guided denoising score at each timestep is:
\begin{equation}
    \hat{\epsilon} = \epsilon_\theta^c(x_t,\, z_c^{(j)})
    + s\!\left(\epsilon_\theta^c(x_t,\, z_c^{(j)})
    - \epsilon_\theta^i(x_t,\, z_i^{(j)})\right),
\end{equation}
where $w \geq 0$ is the guidance scale.
We use the Heun second-order sampler with 35 discretization steps and fix $w{=}1$ for all reported experiments.
By grounding $\epsilon_\theta^i$ in a known training identity $z_i^{(j)}$, novelty can be verified directly against $x_j$ via $\reid$ without exhaustive comparison against the full training set.

\subsection{Latent space identity encoder ablation}
\begin{table*}[t]
\centering
\caption[Small-stem identity encoder ablation on CelebA]{%
    Small stem latent space identity encoder ablation on CelebA ($m=10{,}177$ identities, $n=40$ attributes,
    identity-disjoint train\,/\,val split). 
    Following the same settings as \cref{traceability:tab:encoder_ablation}.
}
\label{traceability:tab:latent_encoder_ablation}
\setlength{\tabcolsep}{5pt}
\renewcommand{\arraystretch}{1.2}
\footnotesize
\resizebox{\textwidth}{!}{%
\begin{tabular}{llcccccc}
    \toprule
    \rowcolor{fau}
    & & \multicolumn{2}{c}{\color{white}\textbf{Class encoding} (\AUCROC $\downarrow$)}
      & \multicolumn{4}{c}{\color{white}\textbf{Identity retrieval} (Recall@$k$ $\uparrow$)} \\
    \rowcolor{fau}
    \color{white}\textbf{Method} & \color{white}$d_i$
        & \color{white}Train & \color{white}Val
        & \multicolumn{2}{c}{\color{white}Train} & \multicolumn{2}{c}{\color{white}Val} \\
    \rowcolor{fau}
    & & & & \color{white}\RatOne & \color{white}R@5 & \color{white}\RatOne & \color{white}R@5 \\
    \midrule
    Siamese, no suppression
        & 512 & 0.827 & 0.819
              & \best{0.824} & \best{0.825} & \best{0.567} & \best{0.759} \\
    \quad + CLUB
        & 512 & 0.883 & 0.876
              & \secondbest{0.822} & \secondbest{0.824} & \secondbest{0.426} & \secondbest{0.632} \\
    \quad + GRL $\lambda{=}5$~$\dagger$
        & 512 & 0.740 & 0.741
              & 0.007 & 0.017 & 0.043 & 0.110 \\
    \quad + GRL $\lambda{=}20$~$\dagger$
        & 512 & \best{0.502} & \best{0.504}
              & 0.000 & 0.001 & 0.002 & 0.008 \\
    \midrule
    Siamese, no suppression
        & 64  & 0.821 & 0.821
              & \best{0.824} & \best{0.825} & 0.320 & 0.549 \\
    Siamese, no suppression
        & 16  & 0.718 & 0.728
              & \best{0.823} & \best{0.825} & 0.237 & 0.470 \\
    \quad + GRL $\lambda{=}5$~$\dagger$
        & 16  & 0.777 & 0.775
              & 0.003 & 0.015 & 0.023 & 0.087 \\
    \quad + GRL $\lambda{=}10$~$\dagger$
        & 16  & 0.686 & 0.684
              & 0.001 & 0.003 & 0.007 & 0.024 \\
    \quad Siamese only (no CLUB)
        & 16  & 0.700 & 0.711
              & \best{0.824} & \best{0.825} & 0.337 & 0.589 \\
    \rowcolor{fau_tech_alpha3}
    \quad \textbf{+ CLUB (Ours)}
        & 16  & \secondbest{0.569} & \secondbest{0.572}
              & \best{0.824} & \best{0.825} & 0.315 & 0.549 \\
    \bottomrule
\end{tabular}}
\end{table*}
Due to the observed advantages of using a latent space feature encoder, we simlarily experiment with a latent space idenity encoder in Tab.~\ref{traceability:tab:latent_encoder_ablation}. 
The results are generatlly inferior, suggesting that the identity encoder profits a lot from image space features.

\subsection{Identity Encoder Ablation}
\label{traceability:sec:app_id_encoder_ablation}

Training a low-dimensional identity encoder that simultaneously achieves strong identity
discriminability and suppresses attribute leakage presents a fundamental challenge: standard
adversarial disentanglement techniques tend to collapse the identity signal entirely once the
suppression penalty is strong enough, while weak penalties leave attribute leakage unaddressed.

\textit{Architecture.}
All runs share the same ResNet-50~\cite{he2016deep} backbone pretrained on
ImageNet~\cite{deng2009imagenet} (V2 weights). 
The identity encoder operates in image space.
The backbone outputs a 2048-d global average-pool feature, projected through a single linear
head to $z_i \in \mathbb{R}^{d_i}$, $\ell_2$-normalized.
The projection head dimensionality $d_i \in \{8, 16, 32, 64, 128, 512\}$ is the primary ablation axis.
$E_i$ is trained entirely independently of $E_c$ with the
CLUB objective targeting the raw CelebA attribute annotations directly, not $z_c$ embeddings.

\textit{Identity loss.}
All runs use the Siamese contrastive loss of the Oracle (Sec.~\ref{traceability:sec:app_lcmem}) on paired samples
$(x_A, x_B)$ drawn from the same identity:
\begin{equation}
    \mathcal{L}_{\mathrm{id}}
    = 0.8\,\mathcal{L}_{\mathrm{fc}}
    + 0.2\,\mathcal{L}_{\mathrm{NT\text{-}Xent}},
    \label{traceability:eq:id_loss}
\end{equation}
where $\mathcal{L}_{\mathrm{fc}}$ is BCE on the absolute difference $|z_i^A - z_i^B|$ via a
learned comparison head, and $\mathcal{L}_{\mathrm{NT\text{-}Xent}}$~\cite{chen2020simpleframework} is the
normalized temperature-scaled cross-entropy loss with temperature $\tau = 0.07$.
Positive pairs are formed by sampling two distinct images of the same CelebA identity within each
minibatch; negatives are the rolled-within-batch complement.

\textit{Suppression mechanisms.}
We compare three suppression strategies applied on top of the Siamese loss, each targeting all
40 binary attributes simultaneously.
\emph{(i) Gradient Reversal Layer (GRL)}~\cite{ganin2016domainadversarial}: an adversarial MLP
$A_c : \mathbb{R}^{d_i} \to \mathbb{R}^{40}$ is inserted after the projection head; a
GRL negates its gradient before backpropagation into the encoder, forcing $z_i$ to be
uninformative about attributes.
The adversarial penalty weight $\lambda \in \{1, 5, 20\}$ is swept per dimensionality.
\emph{(ii) CLUB}~\cite{cheng2020clubcontrastive}: a variational mutual information upper bound estimator
$q_\phi(c \mid z_i)$ is jointly trained with the encoder; the CLUB objective minimizes an
upper bound on $I(z_i; c)$. 
A staged warm-up ramps the CLUB penalty $\lambda_{\mathrm{MI}}$
from 0 to 1 over 10 epochs, allowing the Siamese objective to first establish a stable identity
manifold before MI minimization begins.
\emph{(iii) Bottleneck only}: no adversarial or MI objective is applied; disentanglement relies
entirely on the limited capacity of the $d_i$-dimensional projection head.
Our final model combines CLUB with $d_i = 32$.

\textit{Training.}
All runs use 30 epochs, batch size 1024, Adam
($\beta_1 = 0.9$, $\beta_2 = 0.999$)~\cite{kingmaadammethod}, cosine learning-rate decay with a
one-epoch linear warm-up, lr\_heads $= 3 \times 10^{-3}$, lr\_backbone $= 3 \times 10^{-4}$,
and weight decay $10^{-4}$.
Training data are 3-channel RGB images at $224{\times}224$ spatial resolution; we compare
this to latent-space training in Appx.~\ref{traceability:par:input_space_ablation}.

\textit{Evaluation.}
We evaluate two properties of $z_i$ via post-hoc probes.
\emph{Class encoding} (attribute leakage): a logistic regression probe is fit per attribute on
training embeddings and evaluated on the val split (unseen identities); near-equal train and
val \AUCROC confirms suppression generalises and is not an artifact of probe overfitting.
\emph{Identity discriminability}: Recall@$k$ computed on the val split (unseen identities,
operationally relevant) and a 10k subsample of the train split (memorization check).

\begin{table*}[t]
\centering
\caption[Identity encoder ablation on CelebA]{%
    Identity encoder ablation on CelebA ($m=10{,}177$ identities, $n=40$ attributes,
    identity-disjoint train\,/\,val split).
    \textbf{Class encoding (\AUCROC $\downarrow$)}: linear-probe \AUCROC averaged over all 40 attribute
    classifiers fitted on $z_i$; lower means fewer attributes are decodable from the identity
    embedding. 
    Train and val \AUCROC near-equal indicates genuine suppression (probe not overfitting).
    \textbf{Identity retrieval (Recall@$k$ $\uparrow$)}: fraction of queries whose top-$k$
    nearest neighbours in $z_i$ space share the same identity label; higher means $z_i$ is more
    discriminative. 
    Train recall (10k subsample) measures identity memorization; val recall
    measures generalization to unseen identities, \emph{i.e.}, the operationally relevant quantity.
    $\dagger$~Identity collapsed: train recall $< 0.01$, embedding carries no discriminative signal.
    Best per column in \textbf{bold}, second best \underline{underlined}.
    Our method highlighted in gray.
}
\label{traceability:tab:encoder_ablation}
\setlength{\tabcolsep}{5pt}
\renewcommand{\arraystretch}{1.2}
\footnotesize
\resizebox{\textwidth}{!}{%
\begin{tabular}{llcccccc}
    \toprule
    \rowcolor{fau}
    & & \multicolumn{2}{c}{\color{white}\textbf{Class encoding} (\AUCROC $\downarrow$)}
      & \multicolumn{4}{c}{\color{white}\textbf{Identity retrieval} (Recall@$k$ $\uparrow$)} \\
    \rowcolor{fau}
    \color{white}\textbf{Method} & \color{white}$d_i$
        & \color{white}Train & \color{white}Val
        & \multicolumn{2}{c}{\color{white}Train} & \multicolumn{2}{c}{\color{white}Val} \\
    \rowcolor{fau}
    & & & & \color{white}\RatOne & \color{white}R@5 & \color{white}\RatOne & \color{white}R@5 \\
    \midrule
    Siamese, no suppression
        & 512 & 0.894 & 0.890
              & 0.771 & 0.815 & \best{0.808} & \best{0.903} \\
    \quad + CLUB
        & 512 & 0.846 & 0.840
              & 0.695 & 0.778 & 0.784 & 0.886 \\
    \quad + GRL~$\dagger$
        & 512 & 0.639 & \secondbest{0.640}
              & 0.001 & 0.007 & 0.017 & 0.060 \\
    \midrule
    Siamese, no suppression
        & 128 & 0.869 & 0.867
              & 0.776 & \secondbest{0.817} & \secondbest{0.796} & \secondbest{0.898} \\
    Siamese, no suppression
        & 16  & 0.795 & 0.798
              & 0.757 & 0.814 & 0.620 & 0.802 \\
    Siamese, no suppression
        & 8   & 0.773 & 0.780
              & 0.569 & 0.753 & 0.403 & 0.653 \\
    Siamese, no suppression
        & 64  & 0.851 & 0.849
              & \secondbest{0.777} & \secondbest{0.817} & 0.787 & 0.893 \\
    \quad + CLUB
        & 64  & 0.679 & 0.678
              & 0.737 & 0.800 & 0.755 & 0.868 \\
    \quad + CLUB (5\,steps)
        & 64  & 0.671 & 0.673
              & 0.564 & 0.713 & 0.696 & 0.834 \\
    \quad + CLUB ($\lambda{=}5$)
        & 64  & 0.715 & 0.714
              & 0.534 & 0.669 & 0.745 & 0.854 \\
    \quad + CLUB ($\lambda{=}5$, 5\,steps)
        & 64  & 0.639 & \best{0.637}
              & 0.013 & 0.045 & 0.089 & 0.223 \\
    \quad + GRL~$\dagger$
        & 64  & 0.652 & 0.648
              & 0.003 & 0.013 & 0.023 & 0.080 \\
    Siamese, no suppression
        & 32  & 0.812 & 0.815
              & \best{0.783} & \best{0.819} & 0.739 & 0.866 \\
    \rowcolor{fau_tech_alpha3}
    \quad \textbf{+ CLUB (Ours)}
        & 32  & \best{0.631} & \secondbest{0.640}
              & 0.733 & 0.801 & 0.718 & 0.851 \\
    \quad + CLUB ($\lambda{=}5$)
        & 32  & 0.641 & 0.644
              & 0.557 & 0.699 & 0.704 & 0.831 \\
    \quad + CLUB ($\lambda{=}5$, 5\,steps)~$\dagger$
        & 32  & 0.647 & 0.649
              & 0.007 & 0.030 & 0.076 & 0.194 \\
    \bottomrule
\end{tabular}}
\end{table*}

\textit{Analysis.}
Several findings emerge from Table~\ref{traceability:tab:encoder_ablation}.
First, the no-suppression baseline ($d_i{=}512$) achieves the highest identity retrieval
(Val \RatOne{}$\,{=}\,0.808$) but the worst class encoding (\AUCROC{}$\,{=}\,0.890$), confirming that
without any disentanglement objective the identity embedding is heavily confounded with attribute
information.
Second, CLUB at $d_i{=}512$ achieves modest suppression (\AUCROC{}$\,{=}\,0.840$ vs.\ nosup
$0.890$, $\Delta{=}{-}0.050$) while preserving most identity retrieval (Val \RatOne{}$\,{=}\,0.784$,
$\Delta{=}{-}0.024$).
Suppression is substantially weaker than at $d_i{=}16$ (Val \AUCROC $0.840$ vs.\ $0.642$),
consistent with the known degradation of variational MI estimators in high-dimensional spaces.
Third, GRL at $d_i{=}512$ achieves strong suppression (\AUCROC{}$\,{=}\,0.640$, near chance level),
but at the cost of a complete identity collapse (train \RatOne{}$\,{=}\,0.001$): the adversary forces
$z_i$ toward a constant embedding rather than a selective attribute-free representation.
This confirms that GRL is not \emph{selectively} suppressing attribute dimensions but is instead erasing
all structure in $z_i$.
Fourth, reducing $d_i$ progressively from 512 to 16 with no suppression objective yields a
monotone trade-off: val \AUCROC decreases from $0.890$ to $0.798$ while val \RatOne decreases from
$0.808$ to $0.620$.
The capacity constraint physically prevents $z_i$ from simultaneously encoding 40 binary
attributes and a 10k-identity discrimination signal.
At $d_i{=}8$ this trade-off becomes unfavourable (val \RatOne{}$\,{=}\,0.403$), establishing $d_i{=}16$
as the minimum useful bottleneck; $d_i{=}32$ improves both suppression and retrieval
over $d_i{=}16$, making it the preferred choice.
Train and val \AUCROC are near-equal throughout all bottleneck runs (within 0.01), confirming that
suppression generalises to unseen identities rather than being an artefact of probe overfitting.
Fifth, GRL combined with smaller bottlenecks does not recover: at every $d_i$ tested, GRL drives
train \RatOne below $0.01$, confirming that the adversarial penalty disrupts identity structure
without achieving the \emph{selective} attribute suppression that CLUB delivers.
Sixth, combining CLUB with $d_i{=}32$ (Ours) achieves the best val suppression among
non-collapsed methods (Val \AUCROC{}$\,{=}\,0.640$) while retaining near-complete identity
discriminability (Val \RatOne{}$\,{=}\,0.718$, Val R@5$\,{=}\,0.851$).
The gain over the pure-bottleneck baseline at $d_i{=}32$ (nosup Val \AUCROC $0.815 \to 0.640$,
$\Delta{=}{-}0.175$) confirms that CLUB provides genuine additional suppression beyond what the
bottleneck alone achieves, once the embedding capacity is constrained enough for the variational
estimator to function reliably.

\subsection{Class Encoder Ablation}
\label{traceability:par:class_encoder_ablation}

We train $E_c$ for 30 epochs on the CelebA training split, using Flux2 latents
with the small-stem ResNet-50 backbone, following Eq.~\eqref{traceability:eq:enc_c_loss}.
We ablate the CLUB penalty weight $\lambda_{\mathrm{MI},c} \in \{1, 10, 50\}$ while keeping
the backbone and head architecture fixed.
Results are shown in Table~\ref{traceability:tab:class_encoder_ablation}.

\begin{table}[t]
\centering
\caption[Class encoder ablation on CelebA]{%
    Class encoder ablation on CelebA ($n=40$ attributes).
    $\lambda_{\mathrm{MI},c}$: CLUB penalty weight on $I(z_c;\,\idl)$.
    \textbf{\AUCROC $\uparrow$}: mean attribute-probe \AUCROC on $z_c$; higher means class attributes
    are more decodable.
    \textbf{\RatOne $\downarrow$}: identity retrieval from $z_c$; must stay below the
    attribute-only baseline of $0.205^\dagger$.
    Best per column in \textbf{bold}.
}
\label{traceability:tab:class_encoder_ablation}
\setlength{\tabcolsep}{6pt}
\renewcommand{\arraystretch}{1.2}
\begin{tabular}{ccc}
    \toprule
    \rowcolor{fau}
    & \multicolumn{1}{c}{\color{white}\textbf{Class encoding}} & \multicolumn{1}{c}{\color{white}\textbf{Identity retrieval}} \\
    \rowcolor{fau}
    \color{white}$\lambda_{\mathrm{MI},c}$ & \color{white}\AUCROC $z_c$\,$\uparrow$ & \color{white}\RatOne $z_c$\,$\downarrow$ ($<0.205^\dagger$) \\
    \midrule
    0    & \best{0.894} & 0.076 \\
    1.0  & 0.864 & 0.046 \\
    10.0 & 0.863 & 0.043 \\
    \rowcolor{fau_tech_alpha3}
    50.0 & 0.865 & \best{0.041} \\
    \bottomrule
\end{tabular}
\end{table}

\textit{Analysis.}
All four runs pass the identity-leakage gate (\RatOne{}$_{z_c} < 0.205$), meaning none of
the configurations leak enough identity to invalidate the class embedding as a label-only
representation.
The unpenalised baseline ($\lambda_{\mathrm{MI},c}{=}0$) achieves the highest attribute \AUCROC
(0.894), as $z_c$ is free to encode any discriminative structure including identity-correlated
features, reflected in a higher leakage (\RatOne{}$_{z_c}{=}0.076$).
Once the CLUB penalty is applied ($\lambda_{\mathrm{MI},c}{\geq}1$), \AUCROC drops to 0.863--0.865
and identity leakage falls to 0.041--0.046, which is a ${\sim}2\times$ reduction in \RatOne at a cost of
only ${\sim}3$\,pp in \AUCROC.
The run with $\lambda_{\mathrm{MI},c}{=}50$ (highlighted) achieves the lowest leakage
(\RatOne{}$_{z_c} = 0.041$) and is selected as the class encoder checkpoint.
Preliminary downstream runs across $\lambda \in \{1, 50\}$ checkpoints did not yield measurable
differences in NR or \IRS, so we adopt the lowest-leakage configuration ($\lambda{=}50$) as the
default $E_c$ for all subsequent experiments and treat the residual leakage difference as
operationally irrelevant.

\subsection{Encoder input space ablation}

\label{traceability:par:input_space_ablation}

\begin{table}[t]
\centering
\caption[Input-space and stem ablation for both encoders]{Input-space and stem ablation for $E_i$ ($d{=}16$, CLUB) and $E_c$.
         Each row trains both encoders from scratch in the same input space.
         \textbf{Identity encoder ($z_i$)}: attribute-probe \AUCROC ($\downarrow$),
         Recall@1 and @5 ($\uparrow$); shown for both train and val splits.
         \textbf{Class encoder ($z_c$)}: attribute-probe \AUCROC ($\uparrow$) and identity retrieval
         \RatOne ($\downarrow$, must be ${<}\,0.205^\dagger$); val split only (probe fitted on train).
         Best per column in the val split in \textbf{bold};
         \colorbox{gray!10}{shading} marks the selected configuration.
         $\ddagger$~Fails $E_i$ disentanglement gate (val \AUCROC{}\,$>\,0.55$).}
\label{traceability:tab:input_space_ablation}
\setlength{\tabcolsep}{5pt}
\renewcommand{\arraystretch}{1.2}
\resizebox{\columnwidth}{!}{%
\begin{tabular}{cccccccc}
    \toprule
    \rowcolor{fau}
    & & & \multicolumn{3}{c}{\color{white}\textbf{Identity encoder} ($z_i$)}
        & \multicolumn{2}{c}{\color{white}\textbf{Class encoder} ($z_c$)} \\
    \rowcolor{fau}
    & \color{white}Latent & \color{white}Sm.\,stem & \color{white}\AUCROC{}\,$\downarrow$ & \color{white}\RatOne{}\,$\uparrow$ & \color{white}R@5\,$\uparrow$ & \color{white}\AUCROC{}\,$\uparrow$ & \color{white}\RatOne{}\,$\downarrow$ \\
    \midrule
    \multirow{3}{*}{\rotatebox{90}{Train\;}}
        &           &           & 0.624 & 0.670 & 0.783 & --    & --    \\
        & \cmark    &           & 0.571 & 0.824 & 0.824 & --    & --    \\
        & \cmark    & \cmark    & 0.569 & 0.824 & 0.825 & --    & --    \\
    \midrule
    \multirow{3}{*}{\rotatebox{90}{Val\;\;}}
        &           &           & $0.635^{\ddagger}$ & \best{0.617} & \best{0.788} & \best{0.896} & 0.156        \\
        & \cmark    &           & \best{0.568}       & 0.256        & 0.482        & 0.854        & \best{0.038} \\
    \rowcolor{fau_tech_alpha3}
        & \cmark    & \cmark    & 0.572              & 0.315        & 0.549        & 0.865        & 0.041        \\
    \bottomrule
\end{tabular}}
\end{table}

To isolate the effect of the input representation and backbone stem, we train both encoders on all
three configurations, namely image space with standard stem, latent space with standard
stem, and latent space with small stem (Ours), from scratch using identical
hyperparameters ($d_i{=}16$, CLUB staged warmup, $\lambda_{\mathrm{MI}}{=}1$,
30 epochs). 
Results are reported in Table~\ref{traceability:tab:input_space_ablation}.

\textit{Image space vs.\ latent space.}
Operating in raw image space yields the highest identity discriminability
(Val~\RatOne{}\,=\,0.617, R@5\,=\,0.788) and the best class attribute decodability
(\AUCROC{}\,$z_c$\,=\,0.896), but attribute suppression in $z_i$ fails the
disentanglement gate (mean \AUCROC{}\,=\,0.635 $\gg$ 0.5).
Raw pixel features encode many attribute-correlated signals such as illumination,
background, and hair texture that are not separable from identity at this bottleneck
dimension; CLUB at $\lambda_{\mathrm{MI}}{=}1$ cannot suppress them to chance level.
Flux2 latents, by contrast, are produced by a variational encoder that already
discards high-frequency pixel-level variation, compressing out much of the
attribute-correlated nuisance; this makes CLUB suppression tractable.

\textit{Standard stem vs.\ small stem (within latent space).}
The standard ResNet-50 stem (7$\times$7 conv, stride 2, max-pool) reduces
$64{\times}64$ latent inputs to a $3{\times}3$ spatial grid before global
average pooling, which is spatially too coarse for fine-grained identity
features.
The small stem (3$\times$3 conv, stride 2, no max-pool) preserves a
$16{\times}16$ intermediate representation.
Switching to the small stem improves Val~\RatOne from 0.256 to 0.315 ($+$5.9\,pp)
at a negligible suppression cost (\AUCROC 0.568 $\to$ 0.572, $+$0.4\,pp).
This gain is operationally significant: a higher-recall $z_i$
produces a more informative guidance signal when steering generation away
from a source identity.

\subsection{MIMIC-CXR Encoder Ablation}
\label{traceability:sec:app_mimic_encoder}

\begin{table*}[t]
\centering
\caption[Encoder ablation on MIMIC-CXR]{%
    Encoder ablation on MIMIC-CXR ($m{=}38{,}577$ identities, $n{=}14$ pathology labels,
    identity-disjoint train\,/\,val split, 91{,}843 train\,/\,11{,}566 val images).
    \textbf{Class encoder} ($E_c$): attribute-probe \AUCROC on $z_c$ ($\uparrow$) and identity
    retrieval \RatOne on $z_c$ ($\downarrow$, must be ${<}\,0.005$, the attribute-retrieval baseline).
    \textbf{Memory encoder} ($E_m$): attribute-probe \AUCROC on $z_m$ ($\downarrow$, gate ${\leq}\,0.55$),
    shown for both train and val splits, and identity Recall@1/5 on the val split ($\uparrow$).
    Best per column in \textbf{bold}, second best \underline{underlined}.
    $\dagger$~Fails $E_m$ disentanglement gate (val \AUCROC{}\,$>\,0.55$).
    \colorbox{gray!10}{Shading} marks the selected configuration.
}
\label{traceability:tab:mimic_encoder_ablation}
\setlength{\tabcolsep}{5pt}
\renewcommand{\arraystretch}{1.2}
\resizebox{\textwidth}{!}{%
\begin{tabular}{lcccccc}
    \toprule
    \rowcolor{fau}
    & \multicolumn{2}{c}{\color{white}\textbf{Class encoder} ($E_c$)}
    & \multicolumn{4}{c}{\color{white}\textbf{Memory encoder} ($E_m$)} \\
    \rowcolor{fau}
    \color{white}\textbf{Config}
        & \color{white}\AUCROC $z_c$ $\uparrow$ & \color{white}\RatOne $z_c$ $\downarrow$
        & \color{white}\AUCROC val $\downarrow$ & \color{white}\AUCROC train $\downarrow$ & \color{white}\RatOne val $\uparrow$ & \color{white}R@5 val $\uparrow$ \\
    \midrule
    30 ep, $\lambda_c{=}50$, $\lambda_m{=}1$
        & 0.690 & \best{0.003}
        & $0.563^{\dagger}$ & \secondbest{0.559} & \secondbest{0.755} & \secondbest{0.764} \\
    50 ep, no CLUB ($\lambda{=}0$)
        & \secondbest{0.720} & \secondbest{0.004}
        & $0.706^{\dagger}$ & $0.714^{\dagger}$ & \best{0.756} & \best{0.764} \\
    \rowcolor{fau_tech_alpha3}
    50 ep, $\lambda_c{=}100$, $\lambda_m{=}2$
        & \best{0.733} & 0.005
        & \best{0.529} & \best{0.526} & 0.752 & 0.763 \\
    \bottomrule
\end{tabular}}
\end{table*}

MIMIC-CXR presents a harder encoder training problem than CelebA for two reasons.
First, the pathology label distribution is heavily long-tailed: labels such as
\textit{Support Devices} and \textit{Cardiomegaly} dominate, while \textit{Fracture}
and \textit{Pleural Other} are rare.
This imbalance makes the CLUB estimator unreliable at low $\lambda$ because the variational
net focuses on high-prevalence attributes and underestimates mutual information for
rare ones, leaving their signal in $z_m$ unsuppressed.
Second, MIMIC-CXR has 38k+ identities across 91k images (${\approx}2.4$ images per
patient), providing far fewer positive pairs per identity than CelebA (${\approx}16$),
which slows convergence of the Siamese objective.

Table~\ref{traceability:tab:mimic_encoder_ablation} summarises the three configurations we compared.
Compared to the initial 30-epoch, $\lambda_c{=}50$ run (01b), doubling the training
budget to 50 epochs and strengthening regularisation to $\lambda_c{=}100$ / $\lambda_m{=}2$
reduces $E_m$ attribute leakage from $0.563$ to $0.529$ (below the $0.55$ gate),
while improving $E_c$ attribute decodability from $0.690$ to $0.733$.
The no-CLUB baseline confirms that suppression is the decisive factor: without it,
$E_m$ val \AUCROC rises to $0.706$, even though recall is marginally higher ($0.756$ vs\
$0.752$), and the encoder memorises identity precisely \emph{because} it retains
pathology-correlated structure.

\subsection{Re-identification Oracle: Training}
\label{traceability:sec:app_lcmem}

\paragraph{Role in the pipeline.}
The re-identification oracle $\reid$ (the \emph{Oracle}) serves two distinct functions in TraceDiffusion.
First, it defines identity novelty: $\reid(\hat{x}_j, x_j) = 0$ is the condition a generated
image must satisfy to be certified as a new identity.
Second, its contrastive objective is reused verbatim to train the identity encoder $E_i$,
ensuring that the identity metric used to steer generation is compatible with the metric
used to certify its output.
We train $\reid$ in two stages: Stage~1 operates on pre-computed Flux2 latents;
Stage~2 fine-tunes the checkpoint in image space with on-the-fly VAE encoding and strong augmentations.

\paragraph{Architecture.}
$\reid$ is a Siamese network with a ConvNeXt-Tiny~\cite{liu2022convnext} backbone shared between
both branches. 
The backbone output is projected to a $512$-dimensional embedding; a learned
comparison head takes the absolute difference $|z^A - z^B|$ and produces a scalar logit.
The input channels are set to $32$ to match the Flux2 latent dimensionality.
Image size is $512 \times 512$ (the spatial resolution of the latents prior to any resizing), 
but following the pretrained model, we perform resizing to $256 \times 256$ and center cropping to $224 \times 224$.

\paragraph{Training data.}
We train on a combined pool of CelebA~\cite{liu2015deep} and MIMIC-CXR~\cite{johnson_mimic-cxr_2019}
Flux2 latents, totalling 296,056 samples split into 193,762\,/\,28,711\,/\,73,583 (train\,/\,val\,/\,test).
Identities are kept distinct across the two datasets, so no identity can occur in both.
Critically, negatives during both training and evaluation are sampled
\emph{within dataset}, because a CelebA face is never compared against a MIMIC-CXR chest
radiograph, which would be trivially easy and inflate \AUCROC.

\paragraph{Loss and optimiser.}
Training minimises a weighted combination of binary cross-entropy on the comparison head
and NT-Xent contrastive loss:
\begin{equation}
    \mathcal{L}_R = 0.80\,\mathcal{L}_{\mathrm{fc}} + 0.20\,\mathcal{L}_{\mathrm{NT\text{-}Xent}},
\end{equation}
with temperature $\tau{=}0.07$.
During training, positive pairs $(z^A, z^B)$ are two latents of the same identity;
negatives are all other identities present in the same batch (in-batch random negatives,
no explicit hard mining).
Optimiser: Adam ($\mathrm{lr}{=}5{\times}10^{-4}$, $\alpha{=}0.80$), batch size 512,
cosine LR decay ($\eta_{\mathrm{min}}{=}1$), early stopping with patience 25.
No spatial augmentations are applied to latents beyond a small per-sample normalisation
perturbation ($p{=}0.01$).

\paragraph{Compute.}
Training runs on a single NVIDIA H100 (95\,GB).
Training converged at epoch 177 (of 500 max) in $1$\,h\,$14$\,min wall time.

\paragraph{Results.}
Table~\ref{traceability:tab:lcmem_results} reports test-set performance evaluated
\emph{separately per dataset} to avoid inflation from cross-modality trivial pairs.

\begin{table}[h]
\centering
\caption[Oracle Stage~1 test performance on Flux2 latents]{Oracle Stage~1 test performance on Flux2 latents.
         Pairs are sampled within dataset; overall metrics pool both datasets.
         95\,\% bootstrap CI for overall \AUCROC computed over 10,000 resamples.}
\label{traceability:tab:lcmem_results}
\setlength{\tabcolsep}{8pt}
\renewcommand{\arraystretch}{1.2}
\resizebox{\columnwidth}{!}{%
\begin{tabular}{lccc}
    \toprule
    \rowcolor{fau}
    \color{white}\textbf{Subset} & \color{white}\textbf{\AUCROC} & \color{white}\textbf{Accuracy} & \color{white}\textbf{Pairs} \\
    \midrule
    CelebA (faces vs.\ faces)            & 0.9396 & 0.870 & 119,994 \\
    MIMIC-CXR (CXR vs.\ CXR)            & 0.9858 & 0.864 & \phantom{0}27,138 \\
    \midrule
    Overall                              & 0.9433 & 0.869 & 147,132 \\
    \quad 95\,\% CI                      & [0.9421, 0.9445] & -- & -- \\
    \bottomrule
\end{tabular}}
\end{table}

\paragraph{Stage~2: image-space fine-tuning.}
Stage~2 fine-tunes the Stage~1 checkpoint in image space: images are encoded on-the-fly to
Flux2 latents via a frozen VAE and the model is trained with strong data augmentations
(random crops, horizontal flip, blur, Gaussian noise, brightness/contrast jitter,
JPEG compression) at $\mathrm{lr}{=}8{\times}10^{-5}$, batch size 64, early stopping
with patience 10, combined BCE ($\alpha{=}0.80$) $+$ NT-Xent loss.

\begin{table}[h]
\centering
\caption[Oracle Stage~2 test performance (image-space fine-tuning)]{Oracle Stage~2 test performance (image-space fine-tuning).
         Pairs are sampled within dataset; overall metrics pool both datasets.
         95\,\% bootstrap CI for overall \AUCROC computed over 10,000 resamples.}
\label{traceability:tab:lcmem_stage2_results}
\setlength{\tabcolsep}{8pt}
\renewcommand{\arraystretch}{1.2}
\resizebox{\columnwidth}{!}{%
\begin{tabular}{lccc}
    \toprule
    \rowcolor{fau}
    \color{white}\textbf{Subset} & \color{white}\textbf{\AUCROC} & \color{white}\textbf{Accuracy} & \color{white}\textbf{Pairs} \\
    \midrule
    CelebA (faces vs.\ faces)            & 0.9482 & 0.884 & 119,994 \\
    MIMIC-CXR (CXR vs.\ CXR)            & 0.9902 & 0.868 & 27,138 \\
    \midrule
    Overall                              & 0.9477 & 0.881 & 147,132 \\
    \quad 95\,\% CI                      & [0.9466, 0.9488] & -- & -- \\
    \bottomrule
\end{tabular}}
\end{table}

Table~\ref{traceability:tab:lcmem_stage2_results} reports the per-dataset test performance after Stage~2:
overall \AUCROC improves from $0.9433$ (Stage~1) to $0.9477$ and accuracy from
$0.869$ to $0.881$.

\subsection{One-vs-All Audit: Protocol and Full Results}
\label{traceability:sec:app_audit}

\paragraph{Auditor.}
The audit of Sec.~\ref{traceability:sec:audit} uses LCMem~\cite{dombrowski_lcmem}, applied to the Flux2 latents and scored by the cosine of its $\ell_2$-normalised embeddings.
It is independent of the pipeline: the generators, the encoders, TrR and NR were all trained or computed with the Oracle (Sec.~\ref{traceability:sec:app_lcmem}), and all audited samples were generated before the audit, at the fixed guidance scale of each method.

\paragraph{Protocol.}
For each dataset we embed every training latent once ($122{,}825$ CelebA images of $6{,}106$ identities; $91{,}843$ MIMIC-CXR images of $38{,}577$ patients).
Every generated latent is compared with all of these embeddings; no nearest-neighbour pre-selection is used.
The threshold $\tau$ is the 95th percentile of the best score of the real validation images against the training embeddings ($\tau{=}\AudTauC$ on CelebA, $\tau{=}\AudTauM$ on MIMIC-CXR).
The reference rate is measured on the real test images, which share no identity with the training set: $3{,}232$ of $60{,}014$ CelebA images ($\AudRefC\%$) and $883$ of $16{,}125$ MIMIC-CXR images ($\AudRefM\%$) exceed $\tau$.
$\mathrm{MR}_{\mathrm{src}}$ uses all training images of the source identity, not only the source image.

\paragraph{Untraceable mode.}
Table~\ref{traceability:tab:app_audit_untraceable} repeats the audit for sampling from random noise (all 50k samples per method).
The other-identity match rates are close to those of traceable mode for every method, so they are a property of the denoisers and not of the inverted start.
The identity-conditioned model reproduces its conditioning identity in $27.0\%$ (CelebA) and $47.3\%$ (MIMIC-CXR) of the samples even from random noise.

\begin{table}[h]
\centering
\caption[One-vs-all audit in untraceable mode]{One-vs-all audit with LCMem in untraceable mode (\%, 50k samples per method). Columns as in Table~\ref{traceability:tab:audit}.}
\label{traceability:tab:app_audit_untraceable}
\setlength{\tabcolsep}{4pt}
\renewcommand{\arraystretch}{1.2}
\small
\resizebox{\columnwidth}{!}{%
\begin{tabular}{lcccc}
    \toprule
    \rowcolor{fau}
    & \multicolumn{2}{c}{\color{white}\textbf{CelebA}} & \multicolumn{2}{c}{\color{white}\textbf{MIMIC-CXR}} \\
    \rowcolor{fau}
    \color{white}\textbf{Method} & \color{white}$\mathrm{MR}_{\mathrm{src}}$ & \color{white}$\mathrm{MR}_{\mathrm{oth}}$ & \color{white}$\mathrm{MR}_{\mathrm{src}}$ & \color{white}$\mathrm{MR}_{\mathrm{oth}}$ \\
    \midrule
    Real test images (reference) & -- & \AudRefC & -- & \AudRefM \\
    EDM-2 & 0.9 & 2.7 & 0.0 & 6.0 \\
    EDM-2 AG & 1.5 & 7.0 & 0.0 & 16.5 \\
    $\epsilon_\theta^i$ only ($w{=}{-}1$) & 27.0 & 6.7 & 47.3 & 10.7 \\
    $\epsilon_\theta^c$ only ($w{=}0$) & 0.7 & 4.0 & 0.1 & 14.9 \\
    TraceDiffusion ($w{=}1$) & 0.0 & 5.4 & 0.0 & 23.2 \\
    \bottomrule
\end{tabular}}
\end{table}

\paragraph{Agreement between the Oracle and LCMem.}
The Oracle is a pairwise classifier thresholded at $0.5$; LCMem is a cosine score thresholded at a one-vs-all false-alarm rate of $5\%$.
Table~\ref{traceability:tab:app_audit_agreement} compares their verdicts on whether a sample still shows its source image (traceable mode, traceable sources).
They agree on about $90\%$ of the reconstructions, but the Oracle attributes a much larger share of the guided and baseline samples to the source than LCMem does.
The novelty rates reported with the Oracle are therefore conservative relative to the audit.

\begin{table}[h]
\centering
\caption[Agreement between the Oracle and LCMem on the source verdict]{Share of samples the Oracle attributes to their source ($\reid{=}1$) and agreement of the Oracle with LCMem (cosine to the source image above $\tau$) on this verdict (\%).}
\label{traceability:tab:app_audit_agreement}
\setlength{\tabcolsep}{4pt}
\renewcommand{\arraystretch}{1.2}
\small
\resizebox{\columnwidth}{!}{%
\begin{tabular}{lcccc}
    \toprule
    \rowcolor{fau}
    & \multicolumn{2}{c}{\color{white}\textbf{CelebA}} & \multicolumn{2}{c}{\color{white}\textbf{MIMIC-CXR}} \\
    \rowcolor{fau}
    \color{white}\textbf{Method} & \color{white}$\reid{=}1$ & \color{white}Agree & \color{white}$\reid{=}1$ & \color{white}Agree \\
    \midrule
    EDM-2 & 40.1 & 61.0 & 40.0 & 60.1 \\
    EDM-2 AG & 36.1 & 68.1 & 52.2 & 48.1 \\
    $\epsilon_\theta^i$ only ($w{=}{-}1$) & 92.0 & 92.1 & 86.6 & 88.0 \\
    $\epsilon_\theta^c$ only ($w{=}0$) & 64.5 & 62.2 & 51.8 & 49.5 \\
    TraceDiffusion ($w{=}1$) & 32.5 & 68.0 & 47.3 & 52.7 \\
    \bottomrule
\end{tabular}}
\end{table}

\begin{figure*}[t]
\centering
\pgfplotsset{audsweep/.style={%
  width=0.46\textwidth, height=4.6cm,
  xmin=-1.15, xmax=1.65, ymin=0, ymax=100,
  xtick={-1,0,0.25,0.5,0.75,1,1.25,1.5},
  xticklabels={$-1$,$0$,$0.25$,$0.5$,$0.75$,$1$,$1.25$,$1.5$},
  ytick={0,25,50,75,100},
  tick label style={font=\scriptsize},
  label style={font=\small},
  title style={font=\small\bfseries, yshift=-2pt},
  grid=major, grid style={dotted, gray!40},
  every axis plot/.append style={line width=0.9pt, mark size=1.6pt},
  xtick align=inside, ytick align=inside,%
}}
\begin{tikzpicture}
\begin{groupplot}[
    group style={group size=2 by 1, horizontal sep=1.6cm},
    audsweep,
    xlabel={Guidance scale $w$},
    ylabel={Match rate (\%)},
]
\nextgroupplot[title={CelebA},
    legend style={at={(0.97,0.95)}, anchor=north east, font=\scriptsize, legend cell align=left}]
\addplot[gray!70, densely dashed, line width=0.7pt, no markers, forget plot]
    coordinates {(1,0) (1,100)};
\addplot[ecblue, solid, mark=*] coordinates {
    (-1,97.70) (0,39.35) (0.25,17.75) (0.5,7.30) (0.75,2.85) (1,1.50) (1.25,0.70) (1.5,0.35)};
\addlegendentry{$\mathrm{MR}_{\mathrm{src}}$}
\addplot[emorange, solid, mark=square*] coordinates {
    (-1,6.00) (0,3.95) (0.25,4.50) (0.5,6.10) (0.75,6.35) (1,5.65) (1.25,5.35) (1.5,5.15)};
\addlegendentry{$\mathrm{MR}_{\mathrm{oth}}$}
\addplot[black, dashed, line width=0.7pt, no markers] coordinates {(-1.15,\AudRefC) (1.65,\AudRefC)};
\addlegendentry{Real test images}
\nextgroupplot[title={MIMIC-CXR}]
\addplot[gray!70, densely dashed, line width=0.7pt, no markers, forget plot]
    coordinates {(1,0) (1,100)};
\addplot[ecblue, solid, mark=*] coordinates {
    (-1,94.60) (0,3.35) (0.25,1.15) (0.5,0.70) (0.75,0.25) (1,0.15) (1.25,0.10) (1.5,0.10)};
\addplot[emorange, solid, mark=square*] coordinates {
    (-1,5.05) (0,15.25) (0.25,19.65) (0.5,23.65) (0.75,24.60) (1,25.10) (1.25,22.75) (1.5,22.20)};
\addplot[black, dashed, line width=0.7pt, no markers] coordinates {(-1.15,\AudRefM) (1.65,\AudRefM)};
\end{groupplot}
\end{tikzpicture}
\caption[One-vs-all audit over the guidance scale]{%
    One-vs-all audit with LCMem over the guidance scale $w$ ($2{,}000$ traceable sources per dataset, one sample per source and scale, all from the same inverted latent).
    $\mathrm{MR}_{\mathrm{src}}$ and $\mathrm{MR}_{\mathrm{oth}}$ as in Table~\ref{traceability:tab:audit}; the black dashed line is the false-alarm level of $\mathrm{MR}_{\mathrm{oth}}$ on real test images, and the grey vertical line marks the scale $w{=}1$ used in all other experiments.
    $w{=}{-}1$ is the reconstruction with $\epsilon_\theta^i$ and $w{=}0$ the class-conditioned model alone.
}
\label{traceability:fig:app_audit_sweep}
\end{figure*}
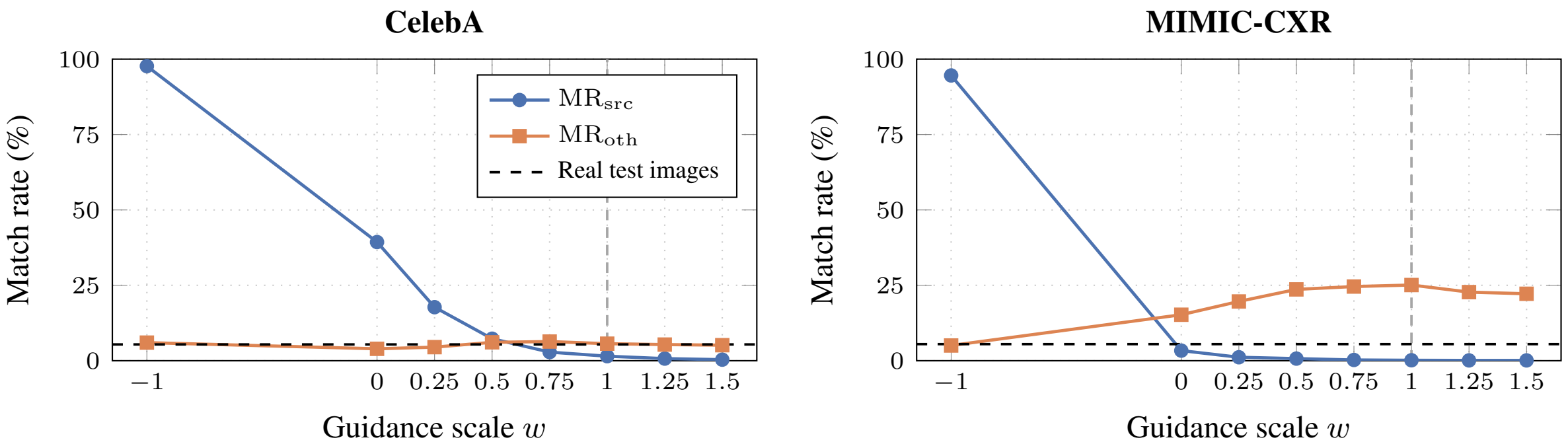

\paragraph{Guidance-scale sweep.}
Table~\ref{traceability:tab:audit} audits three guidance scales.
Figure~\ref{traceability:fig:app_audit_sweep} fills in the scales between and beyond them.
For each dataset we draw $2{,}000$ sources uniformly at random, without replacement, from the traceable sources of Table~\ref{traceability:tab:audit} ($43{,}028$ on CelebA, $45{,}284$ on MIMIC-CXR).
Each source is sampled once per scale $w \in \{-1, 0, 0.25, 0.5, 0.75, 1, 1.25, 1.5\}$ of Eq.~\ref{traceability:eq:guidance}, always from its stored inverted latent, which is the latent used for the 50k sets.
All scales of one source therefore share the same start, and the curves differ only in $w$.
The denoisers $\epsilon_\theta^c$ and $\epsilon_\theta^i$, the conditioning embeddings and the sampler (Heun, 32 steps) are those of the main experiments.
Every sample is then audited as described above, against the same training embeddings and with the same thresholds $\tau$.
With $2{,}000$ samples per point the 95\% intervals are at most $\pm 2.2$ percentage points.

At $w \in \{-1, 0, 1\}$ the sweep reproduces Table~\ref{traceability:tab:audit} to within about one percentage point on both datasets.
On CelebA the source match rate falls steadily with the scale, from $97.7\%$ for the reconstruction to $39.4\%$ at $w{=}0$, $7.3\%$ at $w{=}0.5$, $1.5\%$ at $w{=}1$ and $0.4\%$ at $w{=}1.5$.
The other-identity match rate stays between $4.0\%$ and $6.4\%$ at every scale, around the reference of $\AudRefC\%$: stronger guidance removes the source identity without moving the samples into another training identity.
On MIMIC-CXR the source is left almost immediately ($94.6\%$ for the reconstruction, $3.4\%$ at $w{=}0$, $0.2\%$ at $w{=}1$).
The other-identity match rate behaves differently: it is $5.1\%$ for the reconstruction, $15.3\%$ at $w{=}0$, rises to $25.1\%$ at $w{=}1$ and is still $22.2\%$ at $w{=}1.5$, against a reference of $\AudRefM\%$.
No scale in the tested range brings it back to the reference, so the overlap with other training patients cannot be removed by choosing a different guidance scale.

\subsection{Training Dynamics: Visual Progression and Generative Metrics}
\label{traceability:sec:app_training_dynamics}

To understand \emph{when} the diffusion models acquire their respective capabilities,
we track both qualitative generation quality and quantitative metrics across eight
checkpoint snapshots (5\,k--189\,k steps). 
We generate 20\,k images for each checkpoint. 

\begin{figure}[h]
    \centering
    \includegraphics[width=\linewidth]{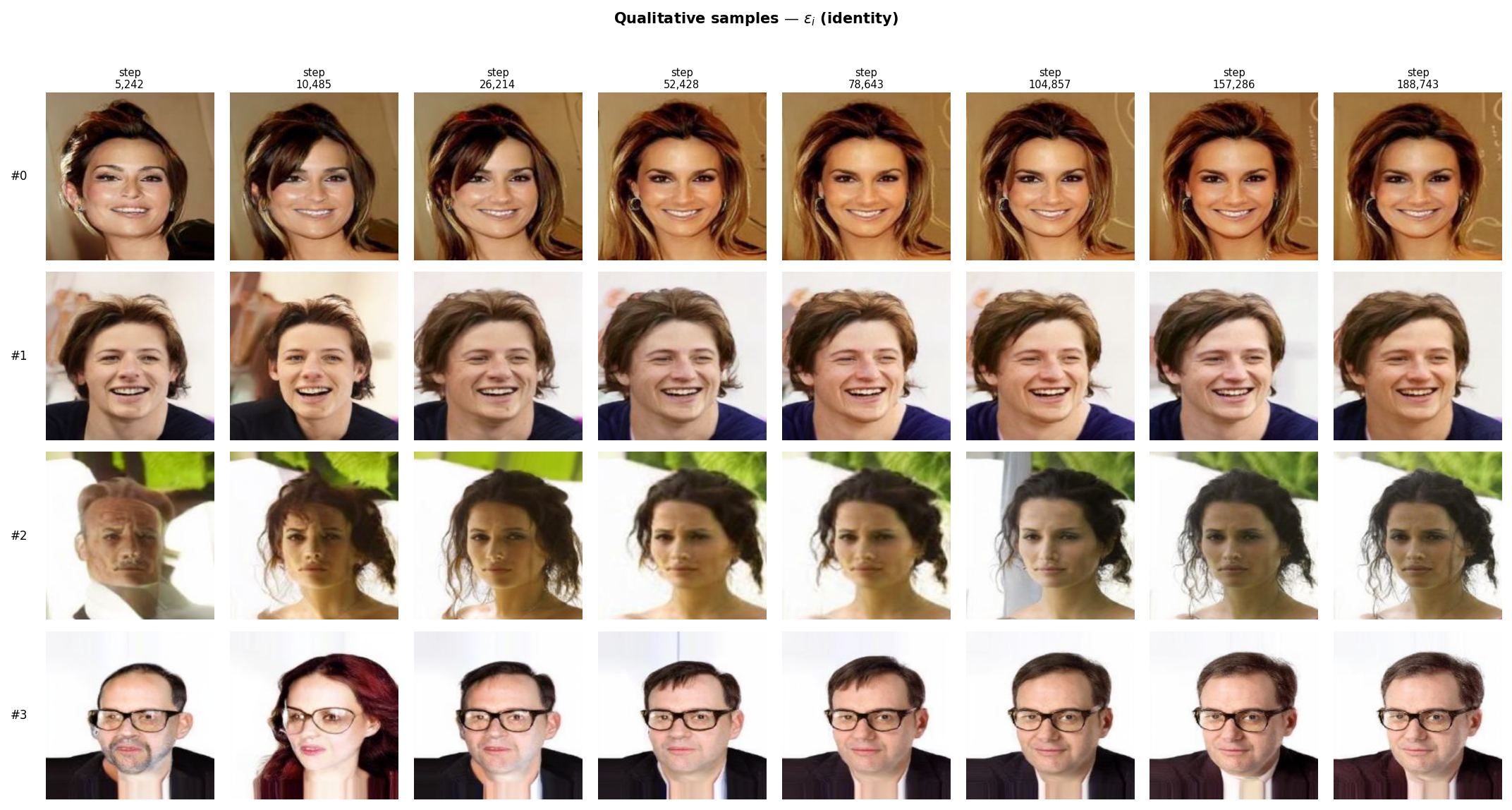}
    \caption[Generated samples across eight training checkpoints]{%
        Visual progression of $\epsilon_\theta^c$ (top row per pair) and $\epsilon_\theta^i$
        (bottom row per pair) across eight training checkpoints for four source
        identities.
        Conditioning image and random seed are fixed.
        Both models converge to a stable macro-identity early; later steps
        primarily sharpen texture without altering the global appearance.
    }
    \label{traceability:fig:training_progression}
\end{figure}

\paragraph{Qualitative progression.}
Figure~\ref{traceability:fig:training_progression} compares $\epsilon_\theta^c$ and $\epsilon_\theta^i$
side-by-side at each checkpoint for four held-out source identities, keeping the
conditioning image and random seed fixed across all checkpoints.
Two observations stand out.
First, despite operating on different conditioning signals ($z_c$ vs.\ $z_i$),
both models produce structurally similar outputs for the same seed: the overall
composition, pose, and rough facial structure agree across the pair at every step.
Second, the global appearance of each generated identity is already
established by ${\sim}10$\,k steps; later training introduces progressively finer
texture and sharpness but does not alter the macro-level identity mapping.
This suggests that the mapping from conditioning embedding to semantic structure
is learned in a short initial phase, while subsequent training is largely spent
refining high-frequency detail.

\begin{figure*}[t]
\centering
\definecolor{ecblue}{HTML}{4C72B0}
\definecolor{emorange}{HTML}{DD8452}
\pgfplotsset{lc/.style={%
  width=0.44\textwidth, height=3.4cm,
  tick label style={font=\tiny},
  label style={font=\scriptsize},
  title style={font=\scriptsize\bfseries, yshift=-2pt},
  grid=major, grid style={dotted, gray!30},
  every axis plot/.append style={line width=0.9pt, mark size=1.6pt},
  xtick align=inside, ytick align=inside, enlarge x limits=0.05,%
}}
\begin{tikzpicture}
\begin{groupplot}[
    group style={group size=2 by 1, horizontal sep=1.8cm},
    lc,
    xlabel={Steps (k)},
]
\nextgroupplot[title={\FID ($\downarrow$)}, ylabel={\FID},
    legend style={
        at={(0.98,0.98)}, anchor=north east,
        font=\scriptsize, legend columns=1,
        /tikz/every even column/.append style={column sep=0.3cm}
    }]
\addplot[ecblue, solid, mark=*] coordinates {
    (5.242,8.215) (10.485,7.395) (26.214,6.663) (52.428,6.657)
    (78.643,7.077) (104.857,7.241) (157.286,7.757) (188.743,7.925)};
\addlegendentry{$\epsilon_\theta^c$}
\addplot[emorange, dashed, mark=square*] coordinates {
    (5.242,8.549) (10.485,7.986) (26.214,6.826) (52.428,6.669)
    (78.643,7.212) (104.857,7.300) (157.286,8.078) (188.743,8.669)};
\addlegendentry{$\epsilon_\theta^i$}
\addplot[gray!70, densely dashed, line width=0.7pt, no markers, forget plot]
    coordinates {(52.428,6.0) (52.428,9.0)};
\nextgroupplot[title={\IRS ($\uparrow$)}, ylabel={\IRS}]
\addplot[ecblue, solid, mark=*, forget plot] coordinates {
    (5.242,0.673) (10.485,0.678) (26.214,0.737) (52.428,0.731)
    (78.643,0.738) (104.857,0.730) (157.286,0.744) (188.743,0.724)};
\addplot[emorange, dashed, mark=square*, forget plot] coordinates {
    (5.242,0.652) (10.485,0.670) (26.214,0.718) (52.428,0.747)
    (78.643,0.735) (104.857,0.730) (157.286,0.734) (188.743,0.718)};
\addplot[gray!70, densely dashed, line width=0.7pt, no markers, forget plot]
    coordinates {(52.428,0.62) (52.428,0.76)};
\end{groupplot}
\end{tikzpicture}
\caption[\FID and \IRS over training steps]{%
    \FID ($\downarrow$) and \IRS ($\uparrow$) over training steps for
    $\epsilon_\theta^c$ (\textcolor{ecblue}{solid blue}) and
    $\epsilon_\theta^i$ (\textcolor{emorange}{dashed orange}).
    The grey dashed line marks step~52\,k, where \FID is minimised for both models.
    \textbf{\FID} drops steeply within the first 50\,k steps and then degrades monotonically,
    suggesting that perceptual quality is acquired early while later optimisation
    primarily minimises training loss.
    \textbf{\IRS}, by contrast, remains elevated well beyond 50\,k: $\epsilon_\theta^c$
    reaches its highest diversity score at 157\,k (0.744), indicating that
    dataset-level diversity continues to improve even as sample fidelity declines.
    This decoupling is also visible in the encoder learning curves
    (\cref{traceability:fig:learning_curves}): identity retrieval and attribute cosine similarity
    keep improving long after \FID has peaked.
}
\label{traceability:fig:fid_irs_curves}
\end{figure*}
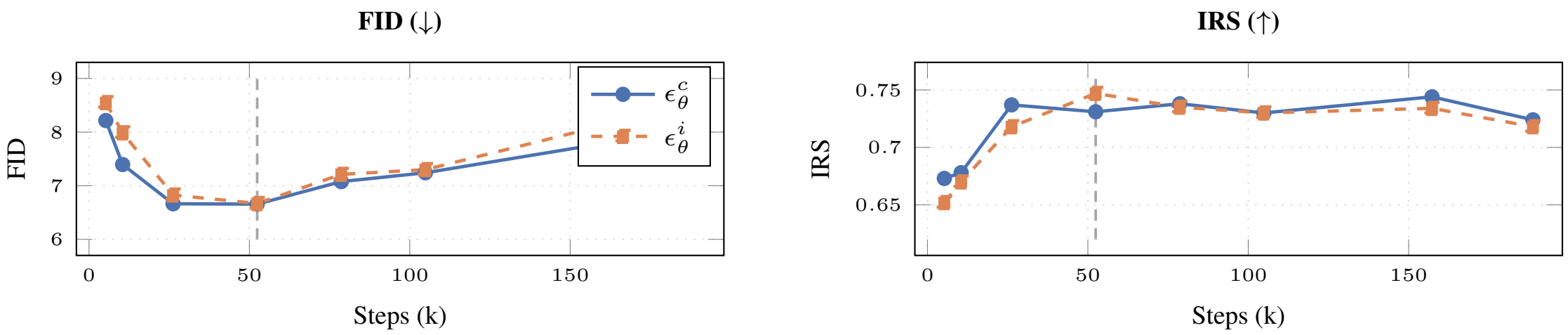

\paragraph{FID and IRS over training.}
Figure~\ref{traceability:fig:fid_irs_curves} shows \FID and \IRS at the same eight checkpoints.

\noindent
\FID improves sharply until ${\approx}52$\,k steps for both models ($\epsilon_\theta^c$: 8.22
$\to$ 6.66; $\epsilon_\theta^i$: 8.55 $\to$ 6.67) and degrades monotonically thereafter,
consistent with the qualitative observation that structural content is fixed early.
\IRS shows a different profile: diversity peaks near 52\,k for $\epsilon_\theta^i$ (0.747)
but continues rising for $\epsilon_\theta^c$, reaching its maximum at 157\,k (0.744) before
declining slightly.
This divergence suggests that optimising the class embedding objective encourages the model
to explore a broader region of attribute space even as raw sample fidelity decreases.
Taken together, \FID and \IRS argue for early-stopped checkpoints in the 26\,k--78\,k range,
consistent with the encoder learning curves in Fig.~\ref{traceability:fig:learning_curves}, which show
that identity retrieval and attribute cosine similarity also plateau in this window.

\paragraph{CelebA annotation noise.}
The CelebA \AUCROC of $0.940$ should be treated as a conservative lower bound.
CelebA is known to carry a non-trivial rate of annotation errors: both the 40 binary
attribute labels and, to a lesser extent, the identity cluster boundaries contain
mistakes~\cite{serianni2025attention,ramaswamy2021fairattributeclassificationlatent,qraitem2023biasmimickingsimplesampling}.
Mislabeled identity assignments create false negatives in the positive pair stream
(two images of the same true person assigned different IDs appear as a failed match),
depressing \AUCROC without reflecting a failure of $\reid$ to discriminate.
The MIMIC-CXR \AUCROC of $0.986$ is higher partly because radiograph identity is
determined by patient record linkage, which is substantially more reliable than
the crowd-sourced face clustering used for CelebA.

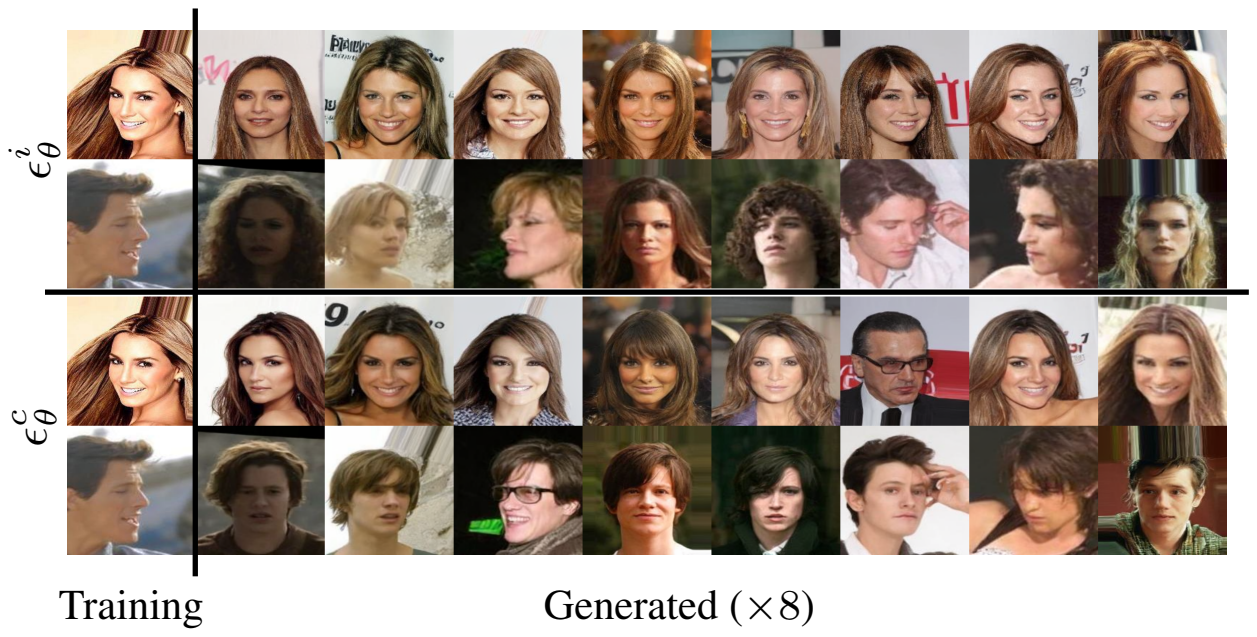
\begin{figure}[htbp]
    \centering
    \begin{tikzpicture}
        \node[inner sep=0pt] (img) {%
            \includegraphics[width=0.92\linewidth]{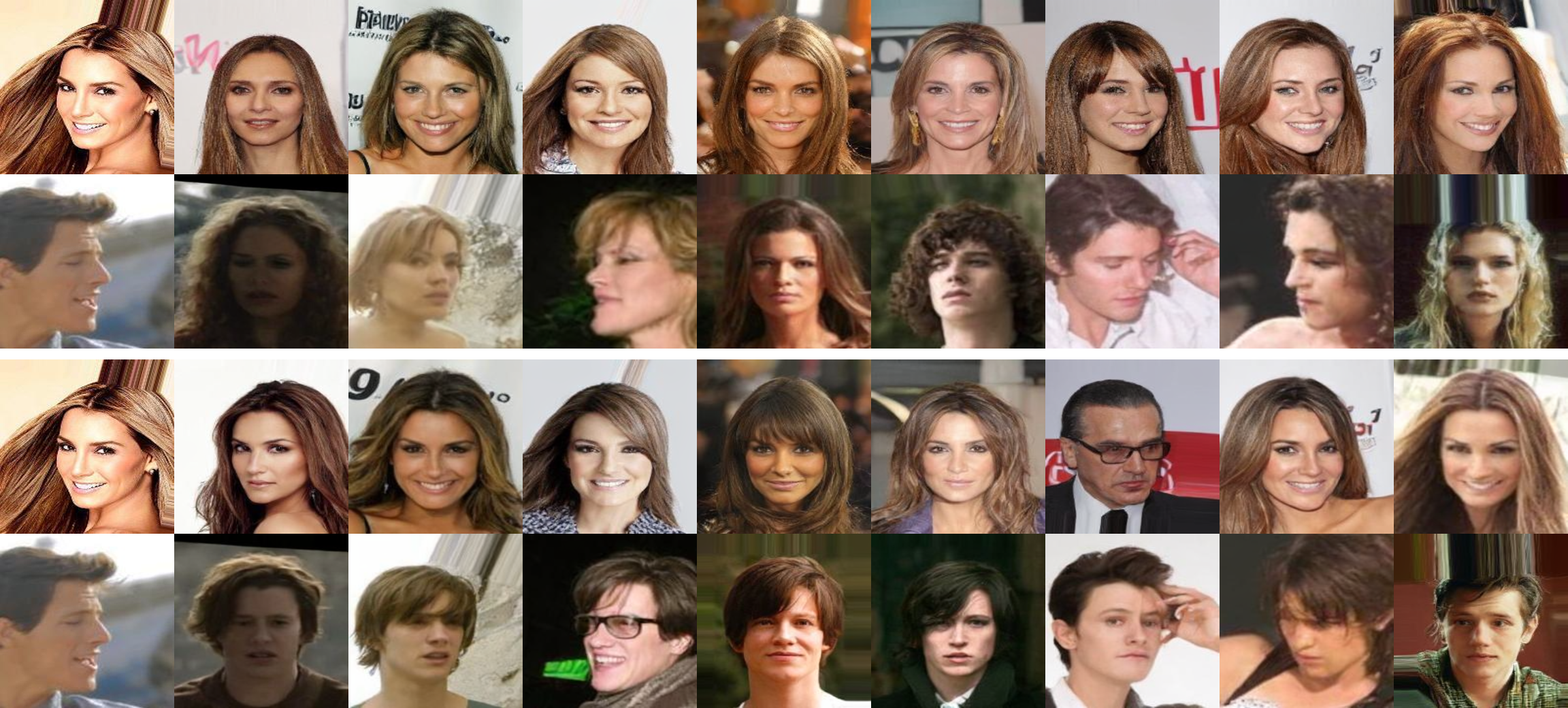}%
        };
        \node[rotate=90, font=\small, anchor=center]
            at ($(img.north west)!0.246!(img.south west) + (-6pt, 0)$)
            {$\epsilon_\theta^i$};
        \node[rotate=90, font=\small, anchor=center]
            at ($(img.north west)!0.753!(img.south west) + (-6pt, 0)$)
            {$\epsilon_\theta^c$};
        \draw[black, line width=1pt]
            ($(img.north west)!0.4995!(img.south west) + (-4pt, 0)$) --
            ($(img.north east)!0.4995!(img.south east) + (4pt, 0)$);
        \draw[black, line width=1pt]
            ($(img.north west)!0.1111!(img.north east) + (0, 4pt)$) --
            ($(img.south west)!0.1111!(img.south east) + (0, -4pt)$);
        \node[below=3pt, font=\footnotesize, anchor=north]
            at ($(img.south west)!0.056!(img.south east)$)
            {Training};
        \node[below=3pt, font=\footnotesize, anchor=north]
            at ($(img.south west)!0.528!(img.south east)$)
            {Generated ($\times 8$)};
    \end{tikzpicture}
    \caption[Seed diversity for two source identities per model]{%
        Seed diversity for two source identities per model.
        Each row shows the source (training) image followed by eight samples
        generated with the same conditioning embedding but different random seeds.
        Rows 1--2: $\epsilon_\theta^i$ (identity-conditioned diffusion model).
        Rows 3--4: $\epsilon_\theta^c$ (class-conditioned diffusion model).
        $\epsilon_\theta^i$ samples remain close to the source identity across seeds;
        $\epsilon_\theta^c$ samples vary more freely in identity while preserving class attributes.
        Row 1 illustrates that identity features are not entirely separated from attribute features:
        the model generates similarly looking women from the same $\epsilon_\theta^i$ conditioning.
        Row 2, by contrast, produces visually diverse individuals, as the identity conditioning
        was trained to not hold much information about visual appearance labels from CelebA.
        Overall, this is consistent with the quantitative results in \cref{traceability:fig:learning_curves},
        which show that attributes are not entirely removable from the identity embedding.
    }
    \label{traceability:fig:seed_diversity}
\end{figure}

\paragraph{Seed diversity.}
\label{traceability:sec:seeddiversity}
Figure~\ref{traceability:fig:seed_diversity} shows how each model responds to different random seeds when
the conditioning embedding is fixed to a single source identity.
Each row holds one source image (leftmost) alongside eight samples drawn with different seeds;
rows 1--2 show $\epsilon_\theta^i$ and rows 3--4 show $\epsilon_\theta^c$.

\subsection{Downstream Utility and the Cost of Traceability}
\label{traceability:sec:app_downstream}
\label{traceability:sec:app_generative_full}

Downstream \AUCROC for all methods is in Table~\ref{traceability:tab:results}. 
EDM-2 leads in absolute \AUCROC from inverted latents on both datasets, with TraceDiffusion trailing by roughly $0.5$ on CelebA and $4$ on MIMIC-CXR. 
Here we test how much of this is the cost of traceability, by pairing the two starting latents of each method across folds (Tab.~\ref{traceability:tab:paired_traceability}).

\begin{table*}[t]
\centering
\caption[Paired significance tests: traceable versus untraceable generation]{%
    Paired traceable-untraceable significance tests. 
    $\bar{x}_{\mathrm{Tr}}$ and
    $\bar{x}_{\mathrm{Un}}$: mean \AUCROC (\%) across 5 identity-disjoint folds.
    $\Delta = \bar{x}_{\mathrm{Tr}} - \bar{x}_{\mathrm{Un}}$.
    $p_t$: one-sample paired $t$-test on per-fold differences.
    $p_W$: Wilcoxon signed-rank test (floor at $0.0625$ for $n{=}5$ when all signs agree).
    BCa CI: bias-corrected and accelerated bootstrap 95\% CI on $\Delta$ ($10^4$ resamples).
    \textbf{No} indicates the paired test cannot reject $H_0:\Delta{=}0$ at $\alpha{=}0.05$.
    TraceDiffusion is the only method with $p_t > 0.05$ on CelebA, indicating traceability without
    measurable cost, and the only method with $p_t < 0.05$ on MIMIC-CXR, where the cost
    is significant and analysed below.
    Tr and Un denote the $\epsilon_\theta^i$-inverted and the Gaussian starting latent; for the EDM-2 baselines Tr is a matched reference, not traceable generation.
    The rightmost column reports the ReID Novelty Rate (NR) in traceable mode at the canonical
    guidance scale used by each method.
}
\label{traceability:tab:paired_traceability}
\setlength{\tabcolsep}{4pt}
\renewcommand{\arraystretch}{1.2}
\footnotesize
\resizebox{\textwidth}{!}{%
\begin{tabular}{llccrcccc}
    \toprule
    \rowcolor{fau}
    \color{white}\textbf{Dataset} & \color{white}\textbf{Method}
        & \color{white}$\bar{x}_{\mathrm{Tr}}$ & \color{white}$\bar{x}_{\mathrm{Un}}$
        & \color{white}$\Delta$ & \color{white}$p_t$ & \color{white}$p_W$ & \color{white}\textbf{BCa CI} & \color{white}\textbf{NR} \\
    \midrule
    CelebA & EDM-2                       & $\PairedEdmCelebATr$ & $\PairedEdmCelebAUn$ & $\PairedEdmCelebADelta$ & $\PairedEdmCelebAPt$ & $\PairedEdmCelebAPW$ & $[\PairedEdmCelebACIlow, \PairedEdmCelebACIhigh]$ & \EdmTrNrC \\
    CelebA & EDM-2 AG                    & $\PairedEdmAgCelebATr$ & $\PairedEdmAgCelebAUn$ & $\PairedEdmAgCelebADelta$ & $\PairedEdmAgCelebAPt$ & $\PairedEdmAgCelebAPW$ & $[\PairedEdmAgCelebACIlow, \PairedEdmAgCelebACIhigh]$ & \EdmAgTrNrC \\
    CelebA & No-id-guidance ($\epsilon_\theta^c, w{=}0$)              & $\PairedNoIdCelebATr$ & $\PairedNoIdCelebAUn$ & $\PairedNoIdCelebADelta$ & $\PairedNoIdCelebAPt$ & $\PairedNoIdCelebAPW$ & $[\PairedNoIdCelebACIlow, \PairedNoIdCelebACIhigh]$ & \NoIdTrNrC \\
    CelebA & No-attr-guidance ($\epsilon_\theta^i, w{=}{-}1$)            & $\PairedNoAttrCelebATr$ & $\PairedNoAttrCelebAUn$ & $\PairedNoAttrCelebADelta$ & $\PairedNoAttrCelebAPt$ & $\PairedNoAttrCelebAPW$ & $[\PairedNoAttrCelebACIlow, \PairedNoAttrCelebACIhigh]$ & \NoAttrTrNrC \\
    \rowcolor{fau_tech_alpha3}
    CelebA & \textbf{TraceDiffusion} ($\hat{\epsilon}, w{=}1$)   & $\PairedOursCelebATr$ & $\PairedOursCelebAUn$ & $\PairedOursCelebADelta$ & $\PairedOursCelebAPt$ & $\PairedOursCelebAPW$ & $[\PairedOursCelebACIlow, \PairedOursCelebACIhigh]$ & \OursTrNrC \\
    \midrule
    MIMIC  & EDM-2                       & $\PairedEdmMimicTr$ & $\PairedEdmMimicUn$ & $\PairedEdmMimicDelta$ & $\PairedEdmMimicPt$ & $\PairedEdmMimicPW$ & $[\PairedEdmMimicCIlow, \PairedEdmMimicCIhigh]$ & \EdmTrNrM \\
    MIMIC  & EDM-2 AG                    & $\PairedEdmAgMimicTr$ & $\PairedEdmAgMimicUn$ & $\PairedEdmAgMimicDelta$ & $\PairedEdmAgMimicPt$ & $\PairedEdmAgMimicPW$ & $[\PairedEdmAgMimicCIlow, \PairedEdmAgMimicCIhigh]$ & \EdmAgTrNrM \\
    MIMIC  & No-id-guidance ($\epsilon_\theta^c, w{=}0$)              & $\PairedNoIdMimicTr$ & $\PairedNoIdMimicUn$ & $\PairedNoIdMimicDelta$ & $\PairedNoIdMimicPt$ & $\PairedNoIdMimicPW$ & $[\PairedNoIdMimicCIlow, \PairedNoIdMimicCIhigh]$ & \NoIdTrNrM \\
    MIMIC  & No-attr-guidance ($\epsilon_\theta^i, w{=}{-}1$)            & $\PairedNoAttrMimicTr$ & $\PairedNoAttrMimicUn$ & $\PairedNoAttrMimicDelta$ & $\PairedNoAttrMimicPt$ & $\PairedNoAttrMimicPW$ & $[\PairedNoAttrMimicCIlow, \PairedNoAttrMimicCIhigh]$ & \NoAttrTrNrM \\
    \rowcolor{fau_tech_alpha3}
    MIMIC  & \textbf{TraceDiffusion} ($\hat{\epsilon}, w{=}1$)   & $\PairedOursMimicTr$ & $\PairedOursMimicUn$ & $\PairedOursMimicDelta$ & $\PairedOursMimicPt$ & $\PairedOursMimicPW$ & $[\PairedOursMimicCIlow, \PairedOursMimicCIhigh]$ & \OursTrNrM \\
    \bottomrule
\end{tabular}}
\end{table*}

In terms of traceability gap, the two datasets exhibit opposite asymmetries.
On CelebA every baseline incurs a cost at $p_t<0.01$ with $\Delta$ between $0.62$ and $1.26$ \AUCROC, whereas TraceDiffusion records $\Delta{=}0.12$ at $p_t{=}0.0987$ with a bootstrap CI of $[0.04, 0.24]$. 
The CI is bounded away from zero, but the effect is five to ten times smaller than for any baseline, which means the effect is smaller by an order of magnitude. 
The Wilcoxon statistic saturates at its small-$n$ floor of $0.0625$ across most rows and serves only as a sign-agreement check. 
On MIMIC-CXR the asymmetry inverts: none of the baselines reject the null, while TraceDiffusion records $\Delta{=}3.78$ \AUCROC at $p_t{=}0.0052$ with a bootstrap CI $[2.69, 5.09]$ comfortably above zero. 
The gap is the cost of the identity-steering mechanism that enables the audit, and the per-attribute analysis below shows it concentrates on a specific family of attributes coupled to patient anatomy. 
Because the per-fold traceable \AUCROC{}s span $\{72.57, 71.31, 70.71, 70.27, 67.63\}$ with the highest fold standard deviation in the column ($1.63$), we ran leave-one-fold-out re-tests: 
removing fold~4 yields $\Delta{=}3.26$, $p_t{=}0.0099$, and removing fold~0 yields $\Delta{=}4.26$, $p_t{=}0.0062$, so no single fold drives the asymmetry (full tables in Appx.~\ref{traceability:sec:app_outliers}). 
Together the two datasets bracket the operating range of an identity-aware generator: CelebA establishes that traceability can be obtained essentially for free, and MIMIC-CXR identifies a regime in which it is paid in measurable \AUCROC.

\noindent\textbf{Per-attribute analysis.}
\label{traceability:sec:per_attr}
To distinguish whether the $\Delta{=}3.78$ \AUCROC gap is spread evenly across the 14 pathology labels or concentrated on a subset, we compute the TSTR \AUCROC per attribute and per fold and inspect two decompositions: $\Delta_{\mathrm{within}}$, the per-attribute paired test for TraceDiffusion (where it pays its own traceability cost), and $\Delta_{\mathrm{vs.\,EDM\text{-}2}}$, the per-attribute gap between TraceDiffusion and EDM-2 in traceable mode (the deficit relative to the strongest baseline). 
Table~\ref{traceability:tab:mimic_per_attr} reports the top five attributes under each.
Both decompositions identify overlapping families dominated by chronic structural findings such as Cardiomegaly, which is read from the heart silhouette and mediastinal contour. 
Once $E_i$ captures these signatures to satisfy its contrastive identity objective, steering away from $z_i^{(j)}$ perturbs the same features on which the downstream classifier relies, so the effect reflects how patient identity manifests in chest radiographs rather than an artefact of our pipeline. 
The CelebA decomposition (Appx.~\ref{traceability:sec:app_celeba_per_attr}) shows no attribute with $|\Delta_{\mathrm{within}}|>1$ \AUCROC, consistent with the small aggregate $\Delta{=}0.12$, so the MIMIC-CXR cost is a property of the domain rather than of the generator and would be incurred by any identity-aware method that must read anatomy to discriminate patients. 
The No-id-guidance ablation on MIMIC-CXR records $\Delta{=}0.49$ \AUCROC at $p_t{=}0.638$, confirming that traceable initialisation is approximately neutral on its own and that the gap arises only from the \emph{interaction} between identity steering and traceable-mode initialisation. 

\begin{table}[t]
\centering
\caption[Per-attribute decomposition of the MIMIC-CXR gap]{%
    Per-attribute decomposition of the MIMIC-CXR gap. 
    Top five attributes by
    $|\Delta|$ shown for each decomposition; remaining nine attributes have
    $|\Delta_{\mathrm{within}}| < 4$ \AUCROC.
    \textbf{$\Delta_{\mathrm{within}}$ (left)}: TraceDiffusion, traceable minus untraceable \AUCROC, capturing the cost of traceability for our method.
    \textbf{$\Delta_{\mathrm{vs.EDM\text{-}2}}$ (right)}: TraceDiffusion minus EDM-2, both in traceable mode, capturing the deficit relative to the strongest baseline in the audited regime.
    Both decompositions identify the same family of chronic structural
    findings whose radiographic appearance is tightly coupled to patient
    anatomy.
}
\label{traceability:tab:mimic_per_attr}
\setlength{\tabcolsep}{6pt}
\renewcommand{\arraystretch}{1.2}
\small
\resizebox{\columnwidth}{!}{%
\begin{tabular}{lccc@{\hskip 1.5em}lccc}
    \toprule
    \rowcolor{fau}
    \multicolumn{4}{c}{\color{white}$\Delta_{\mathrm{within}}$ (Ours: Tr$-$Un)} & \multicolumn{4}{c}{\color{white}$\Delta_{\mathrm{vs.EDM\text{-}2}}$ (Ours$-$EDM-2, Tr)} \\
    \rowcolor{fau}
    \color{white}Attribute & \color{white}Tr & \color{white}Un & \color{white}$\Delta$ & \color{white}Attribute & \color{white}Ours & \color{white}EDM-2 & \color{white}$\Delta$ \\
    \midrule
    \MimicWithinRowOneAttr   & $\MimicWithinRowOneTr$   & $\MimicWithinRowOneUn$   & $\MimicWithinRowOneDelta$   & \MimicVsEdmRowOneAttr   & $\MimicVsEdmRowOneOurs$   & $\MimicVsEdmRowOneEdm$   & $\MimicVsEdmRowOneDelta$   \\
    \MimicWithinRowTwoAttr   & $\MimicWithinRowTwoTr$   & $\MimicWithinRowTwoUn$   & $\MimicWithinRowTwoDelta$   & \MimicVsEdmRowTwoAttr   & $\MimicVsEdmRowTwoOurs$   & $\MimicVsEdmRowTwoEdm$   & $\MimicVsEdmRowTwoDelta$   \\
    \MimicWithinRowThreeAttr & $\MimicWithinRowThreeTr$ & $\MimicWithinRowThreeUn$ & $\MimicWithinRowThreeDelta$ & \MimicVsEdmRowThreeAttr & $\MimicVsEdmRowThreeOurs$ & $\MimicVsEdmRowThreeEdm$ & $\MimicVsEdmRowThreeDelta$ \\
    \MimicWithinRowFourAttr  & $\MimicWithinRowFourTr$  & $\MimicWithinRowFourUn$  & $\MimicWithinRowFourDelta$  & \MimicVsEdmRowFourAttr  & $\MimicVsEdmRowFourOurs$  & $\MimicVsEdmRowFourEdm$  & $\MimicVsEdmRowFourDelta$  \\
    \MimicWithinRowFiveAttr  & $\MimicWithinRowFiveTr$  & $\MimicWithinRowFiveUn$  & $\MimicWithinRowFiveDelta$  & \MimicVsEdmRowFiveAttr  & $\MimicVsEdmRowFiveOurs$  & $\MimicVsEdmRowFiveEdm$  & $\MimicVsEdmRowFiveDelta$  \\
    \bottomrule
\end{tabular}}
\end{table}

\subsection{Static Downstream Classification Validation}

We evaluate downstream utility using a locked static classification protocol. 
We measure the retention of classification performance when real training data are replaced by synthetic data from source $s \in \{\text{Real, EDM2, EDM2-AG, Ours}\}$. 
All other experimental factors, including dataset splits, classifier architecture, and optimization hyperparameters, are fixed a priori.

\subsubsection{Datasets and Task Definition}
Validation is performed on MIMIC-CXR (diagnostic utility) and CelebA (semantic utility). 
Tasks are formulated as binary classification problems $y_k \in \{0,1\}$ for each label $k$. 
Labels are included based on predefined support thresholds: $N_{\text{train}} \geq 1000$ and $N_{\text{test}} \geq 100$ for both positive and negative classes.

\subsubsection{Training-Set Construction}
For each source $s$ and label $k$, we construct a class-balanced training set $\mathcal{D}^{s,k}$ such that $|\mathcal{D}^{s,k}_{\text{pos}}| = |\mathcal{D}^{s,k}_{\text{neg}}| = N_k$. 
To ensure a fair comparison, the per-label budget $N_k$ is the maximum balanced budget supported by all sources, capped at a global maximum $N_{\max}$:
\[
N_k = \min \left( N_{\text{real,pos}}^k, N_{\text{real,neg}}^k, \dots, N_{\text{Ours,neg}}^k, N_{\max} \right)
\]

\subsubsection{Non-Inferiority and Comparisons}
We assess non-inferiority against the matched real-data baseline using a margin of $\delta = 0.03$. 
Our method is non-inferior if the lower bound of the 95\% CI for $\Delta^{\text{Ours}-\text{Real}}$ exceeds $-\delta$. 
Primary comparisons ($\text{Ours}$ vs. 
$\text{EDM2/EDM2-AG}$) are corrected for multiple comparisons using the Holm-Bonferroni method.

\subsection{Empirical Verification of the Novel Identity Existence Condition}
\label{traceability:sec:app_delta}

Proposition~\ref{traceability:prop:novelty} requires
\begin{equation}
    d_i \;>\; \frac{\log m}{\log(\rho/\delta)},
\end{equation}
where $m$ is the number of training identities, $d_i = 32$ is the identity embedding dimension, $\rho = 1$ is the enclosing ball radius (exact, since $z_i$ embeddings are $\ell_2$-normalised), and $\delta$ is the identity separation radius estimated as half the 5th percentile of cross-identity pairwise distances in $\mathcal{Z}_i$.

\paragraph{Measurement protocol.}
After caching all training embeddings $\{z_i^{(j)}\}$, we compute a per-identity mean embedding and re-normalise it to the unit sphere. 
We then compute all $\binom{m}{2}$ pairwise Euclidean distances between identity means and take the 5th percentile; $\delta = \mathrm{p5} / 2$. 
Only cross-identity pairs are used; within-identity distances are excluded by construction.

\paragraph{Results.}
Table~\ref{traceability:tab:delta} summarises the measurements for both datasets.

\begin{table}[h]
\centering
\caption[Empirical verification of the novel identity existence condition]{Empirical verification of the novel identity existence condition ($d_i = 32$, $\rho = 1$).}
\label{traceability:tab:delta}
\renewcommand{\arraystretch}{1.2}
\resizebox{\columnwidth}{!}{%
\begin{tabular}{lrrrrrc}
\toprule
\rowcolor{fau}
\color{white}\textbf{Dataset} & \color{white}$m$ & \color{white}$\mathrm{p5}$ & \color{white}$\delta$ & \color{white}$\rho/\delta$ & \color{white}$\log(m)/\log(\rho/\delta)$ & \color{white}\textbf{Condition} \\
\midrule
CelebA     & 6{,}106  & 1.123 & 0.562 & 1.781 & 15.10 & $32 > 15.10$ \checkmark \\
MIMIC-CXR  & 38{,}577 & 1.158 & 0.579 & 1.727 & 19.34 & $32 > 19.34$ \checkmark \\
\bottomrule
\end{tabular}}
\end{table}

Both datasets satisfy the condition comfortably. 
The pairwise distance distributions are similar across datasets (median $\approx 1.41$, close to the expected value of $\sqrt{2}$ for random unit vectors in $\mathbb{R}^{32}$), confirming that the trained encoder spreads identity means broadly across the unit sphere while maintaining clear separation ($\mathrm{p5} > 1.1$ out of a maximum of $2$).

\subsubsection{Ablations with and without adversarial}
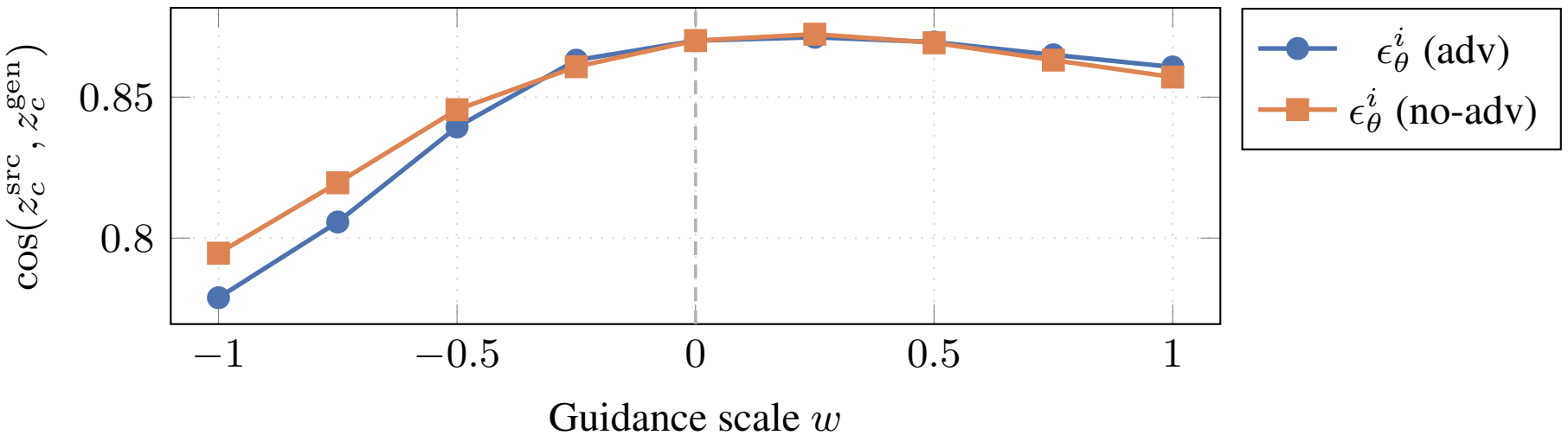
\begin{figure*}[t]
\centering
\definecolor{cadv}{HTML}{4C72B0}
\definecolor{cnoadv}{HTML}{DD8452}
\begin{tikzpicture}
\begin{axis}[
    width=0.55\textwidth, height=4cm,
    xlabel={Guidance scale $w$},
    ylabel={$\cos(z_c^{\mathrm{src}},z_c^{\mathrm{gen}})$},
    grid=major, grid style={dotted, gray!30},
    tick label style={font=\small},
    label style={font=\small},
    legend style={font=\small, at={(1.02,1)}, anchor=north west},
    every axis plot/.append style={line width=1pt, mark size=2pt},
    xmin=-1.1, xmax=1.1,
]
\draw[gray!60, densely dashed, line width=0.7pt] (axis cs:0,\pgfkeysvalueof{/pgfplots/ymin}) -- (axis cs:0,\pgfkeysvalueof{/pgfplots/ymax});
\addplot[cadv, solid, mark=*] coordinates {(-1.00,0.778863) (-0.75,0.805717) (-0.50,0.839404) (-0.25,0.863114) (0.00,0.870043) (0.25,0.871238) (0.50,0.869537) (0.75,0.865007) (1.00,0.860693)};
\addlegendentry{$\epsilon_\theta^i$ (adv)}
\addplot[cnoadv, solid, mark=square*] coordinates {(-1.00,0.794647) (-0.75,0.819625) (-0.50,0.845548) (-0.25,0.860792) (0.00,0.870043) (0.25,0.872328) (0.50,0.869281) (0.75,0.863133) (1.00,0.857123)};
\addlegendentry{$\epsilon_\theta^i$ (no-adv)}
\end{axis}
\end{tikzpicture}
\caption[Class embedding similarity versus guidance scale]{%
    Cosine similarity between source and generated class embedding $z_c$
    as a function of guidance scale $w$.
    At $w=0$ the sampler reduces to unconditional $\epsilon_\theta^c$.
    Higher $w$ pushes the output further from the source identity.
}
\label{traceability:fig:cosine_sim_vs_scale}
\end{figure*}

\begin{figure*}[t]
\centering
\definecolor{cadv}{HTML}{4C72B0}
\definecolor{cnoadv}{HTML}{DD8452}
\begin{tikzpicture}
\begin{axis}[
    width=0.55\textwidth, height=4cm,
    xlabel={Guidance scale $w$},
    ylabel={$\cos(z_c^{\mathrm{src}},z_c^{\mathrm{gen}})$},
    grid=major, grid style={dotted, gray!30},
    tick label style={font=\small},
    label style={font=\small},
    legend style={font=\small, at={(1.02,1)}, anchor=north west},
    every axis plot/.append style={line width=1pt, mark size=2pt},
    xmin=-1.1, xmax=1.1,
]
\draw[gray!60, densely dashed, line width=0.7pt] (axis cs:0,\pgfkeysvalueof{/pgfplots/ymin}) -- (axis cs:0,\pgfkeysvalueof{/pgfplots/ymax});
\addplot[cadv, solid, mark=*] coordinates {(-1.00,0.919589) (-0.75,0.930284) (-0.50,0.912567) (-0.25,0.900642) (0.00,0.893625) (0.25,0.887090) (0.50,0.884760) (0.75,0.878469) (1.00,0.871350)};
\addlegendentry{$\epsilon_\theta^i$ (adv)}
\addplot[cnoadv, solid, mark=square*] coordinates {(-1.00,0.925374) (-0.75,0.932839) (-0.50,0.913185) (-0.25,0.900419) (0.00,0.893253) (0.25,0.884331) (0.50,0.880504) (0.75,0.873445) (1.00,0.867963)};
\addlegendentry{$\epsilon_\theta^i$ (no-adv)}
\end{axis}
\end{tikzpicture}
\caption[Class embedding similarity versus guidance scale (DDIM inversion)]{%
    Cosine similarity between source and generated class embedding $z_c$
    vs.\ guidance scale $w$ (DDIM inversion start).
    At $w=0$ the sampler reconstructs the source via $\epsilon_\theta^c$.
    Higher $w$ pushes the output further from the source identity.
}
\label{traceability:fig:cosine_sim_vs_scale_inverted}
\end{figure*}
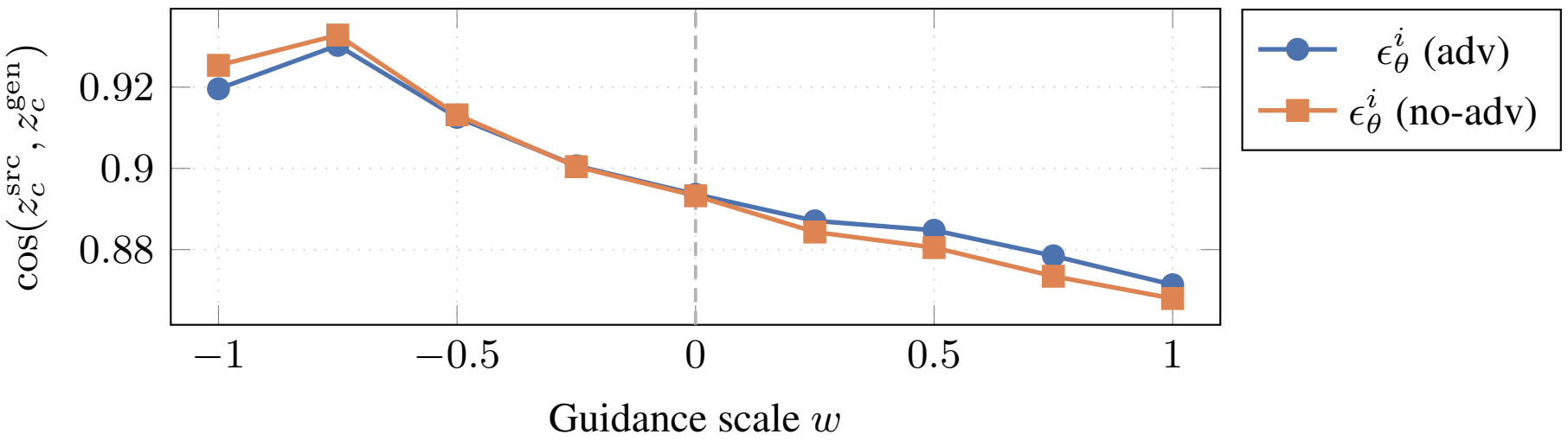

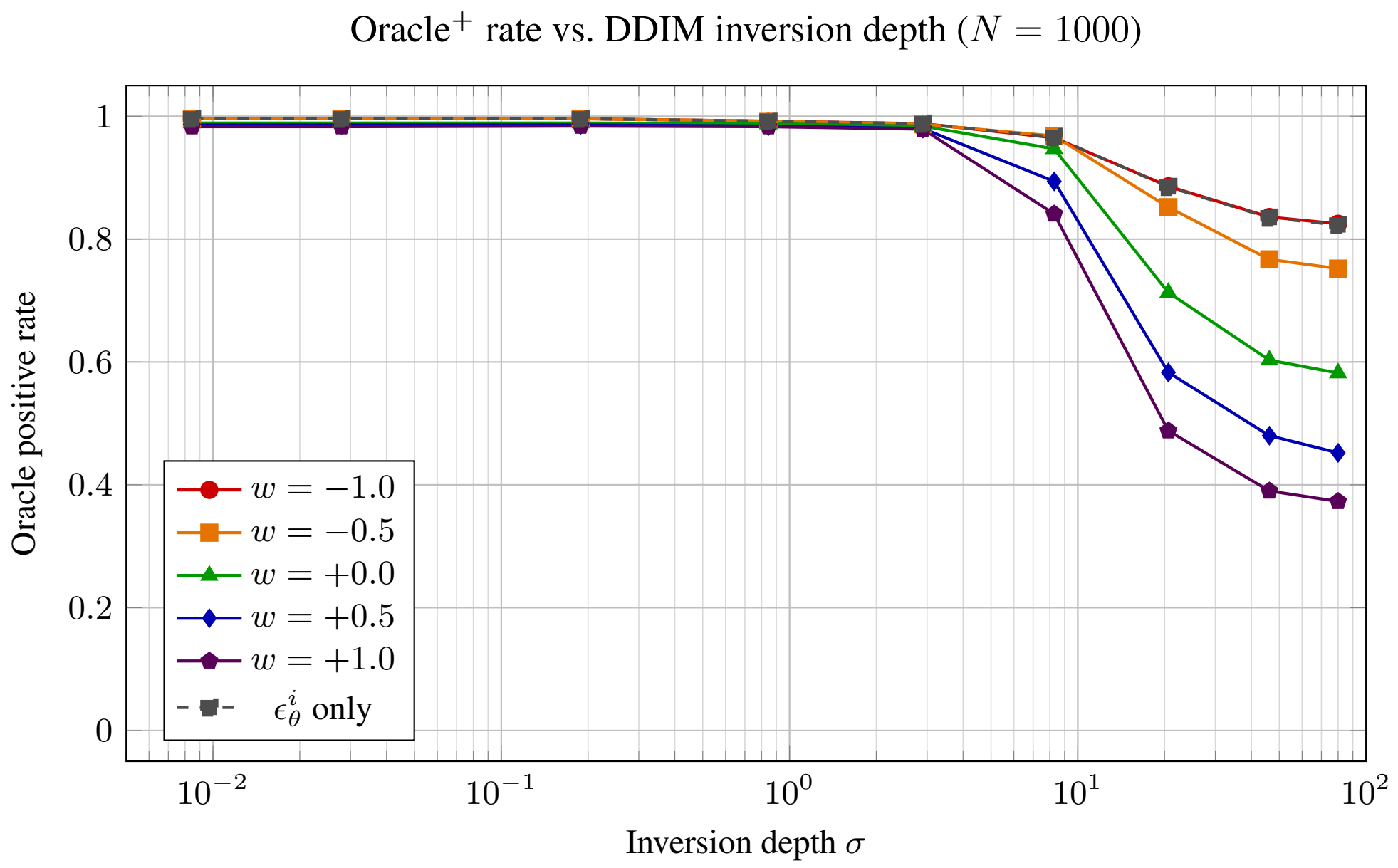
\begin{figure*}[t]
\centering
\begin{tikzpicture}
\begin{axis}[
  width=0.75\textwidth,
  height=0.45\textwidth,
  xmode=log,
  xmin=0.005, xmax=100,
  ymin=-0.05, ymax=1.05,
  xlabel={Inversion depth $\sigma$},
  ylabel={Oracle positive rate},
  title={Oracle$^+$ rate vs.\ DDIM inversion depth ($N=1000$)},
  legend pos=south west,
  legend style={font=\small},
  grid=both,
  grid style={line width=0.3pt, draw=gray!30},
  major grid style={line width=0.4pt, draw=gray!50},
  tick label style={font=\small},
  label style={font=\small},
  mark size=2pt,
]
\addplot[color=red!80!black, mark=*, thick] coordinates {(0.00845305,0.9960) (0.0279158,0.9960) (0.188599,0.9960) (0.84462,0.9920) (2.90153,0.9880) (8.28158,0.9650) (20.6141,0.8860) (46.1864,0.8360) (80,0.8250)};
\addlegendentry{$w=-1.0$}
\addplot[color=orange!90!black, mark=square*, thick] coordinates {(0.00845305,0.9960) (0.0279158,0.9960) (0.188599,0.9960) (0.84462,0.9920) (2.90153,0.9870) (8.28158,0.9680) (20.6141,0.8520) (46.1864,0.7670) (80,0.7520)};
\addlegendentry{$w=-0.5$}
\addplot[color=green!60!black, mark=triangle*, thick] coordinates {(0.00845305,0.9890) (0.0279158,0.9890) (0.188599,0.9890) (0.84462,0.9880) (2.90153,0.9840) (8.28158,0.9470) (20.6141,0.7130) (46.1864,0.6030) (80,0.5820)};
\addlegendentry{$w=+0.0$}
\addplot[color=blue!70!black, mark=diamond*, thick] coordinates {(0.00845305,0.9860) (0.0279158,0.9860) (0.188599,0.9860) (0.84462,0.9840) (2.90153,0.9800) (8.28158,0.8940) (20.6141,0.5830) (46.1864,0.4800) (80,0.4520)};
\addlegendentry{$w=+0.5$}
\addplot[color=violet!70!black, mark=pentagon*, thick] coordinates {(0.00845305,0.9830) (0.0279158,0.9830) (0.188599,0.9840) (0.84462,0.9830) (2.90153,0.9790) (8.28158,0.8410) (20.6141,0.4880) (46.1864,0.3900) (80,0.3730)};
\addlegendentry{$w=+1.0$}
\addplot[color=black!70, mark=square*, dashed, thick] coordinates {(0.00845305,0.9960) (0.0279158,0.9960) (0.188599,0.9960) (0.84462,0.9920) (2.90153,0.9880) (8.28158,0.9660) (20.6141,0.8850) (46.1864,0.8350) (80,0.8230)};
\addlegendentry{$\epsilon_\theta^i$ only}
\end{axis}
\end{tikzpicture}
\caption[Oracle$^+$ rate versus DDIM inversion depth]{%
    Oracle$^+$ rate as a function of DDIM inversion depth $\sigma$ for guided generation
    with scale $w \in \{-1.0, -0.5, 0.0, 0.5, 1.0\}$ (solid) and $\epsilon_\theta^i$-only
    denoising (dashed), evaluated on $N{=}1000$ CelebA training images.
    At shallow inversions ($\sigma \lesssim 1$) all curves plateau near $1.0$: the latent
    has barely moved from $x_0$, so the round-trip output is trivially recognised as
    the source.
    The transition region $\sigma \approx 3$--$20$ reveals the guidance effect: positive
    scales ($w > 0$) steer generation away from the source identity and drive the
    memorisation rate down substantially, while negative scales amplify the identity
    signal and keep rates high.
    At full inversion ($\sigma{=}80$), $w{=}{+}1.0$ achieves a memorisation rate of only
    ${\approx}0.06$, compared to ${\approx}0.64$ for $\epsilon_\theta^i$-only.
    As expected, $w{=}{-}1.0$ and $\epsilon_\theta^i$-only overlap throughout, confirming that
    negative-unity guidance is mathematically equivalent to pure identity-conditioned
    denoising (sanity check).
}
\label{traceability:fig:lcmem_vs_sigma}
\end{figure*}

Fig.~\ref{traceability:fig:cosine_sim_vs_scale} shows class-conditioning retention as a function of guidance scale $w$, comparing adversarial training of $\epsilon_\theta^i$ against a no-adversarial variant (Eq.~\ref{traceability:eq:guidance}).
A guidance scale of $w{=}{-}1$ reduces to using only $\epsilon_\theta^i$ for generation.
As expected, without adversarial learning the model relies more on attribute features, which leads to higher cosine similarity between the generated and training class embeddings $z_c$.
At $w{=}0$ (full class-conditional generation) both variants are equivalent, as both use the same $\epsilon_\theta^c$.
Increasing $w > 0$ to steer away from the source identity reveals the advantage of adversarial learning: because $\epsilon_\theta^i$ retains fewer attribute features, subtracting it from $\epsilon_\theta^c$ removes less of the attribute signal and preserves more of the target attribute profile compared to the no-adversarial counterpart.
Interestingly, the no-adversarial model peaks in attribute retention at $w{=}0.25$; this regime is still within the region where the Oracle predicts the same identity as the training image (see Fig.~\ref{traceability:fig:lcmem_vs_sigma}).
Fig.~\ref{traceability:fig:cosine_sim_vs_scale_inverted} repeats this analysis when sampling is initialised from the DDIM-inverted latent of the source image (traceable mode); the same qualitative ordering holds, with adversarial training preserving more $z_c$ at $w{>}0$.

\subsection{Full Qualitative Generation Examples}
\label{traceability:sec:app_qualitative}

\begin{figure}[h]
    \centering
    \includegraphics[width=\linewidth]{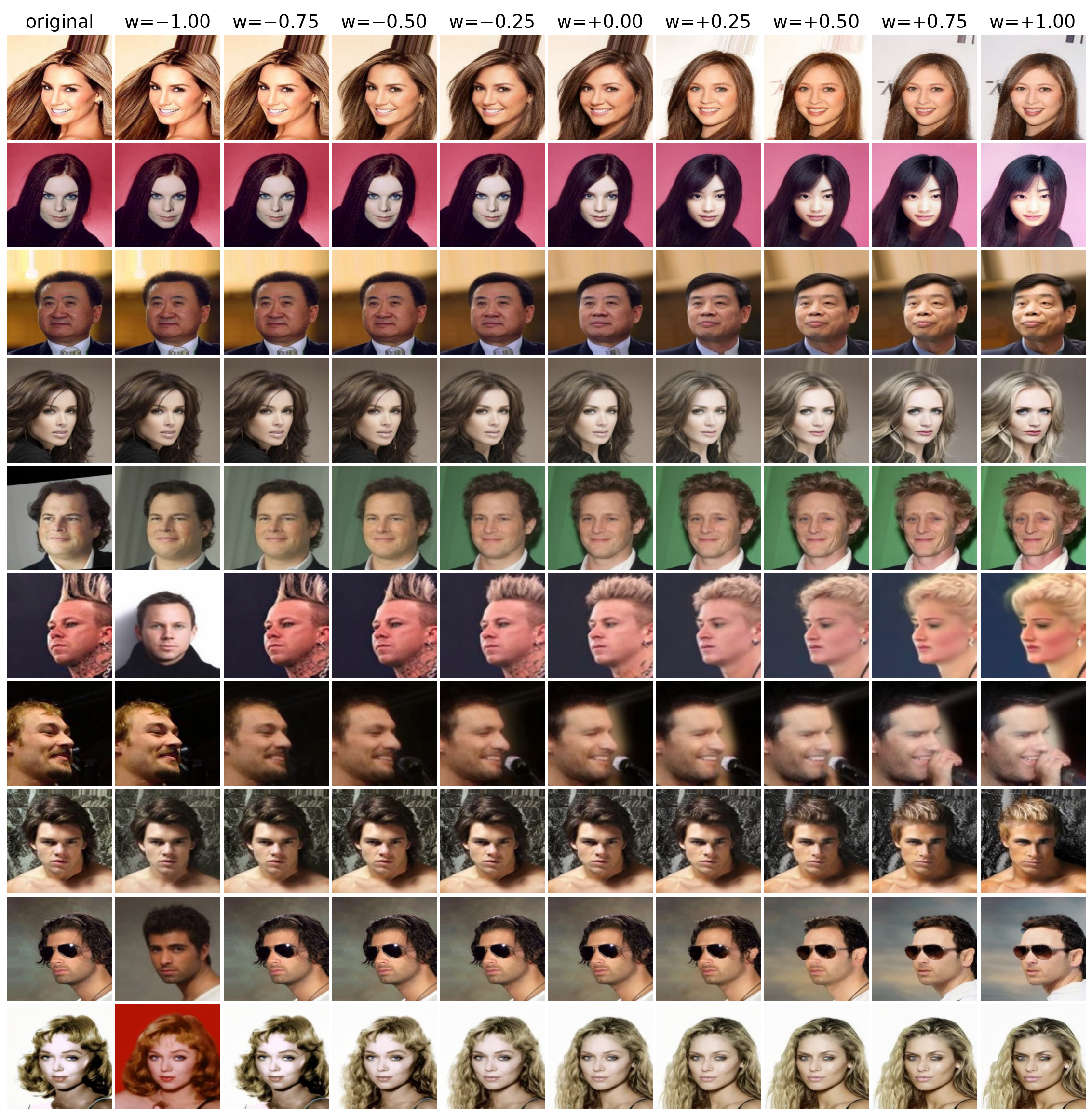}
    \caption[CelebA: source identities at varying guidance scales]{%
        CelebA: 10 source identities at guidance scales
        $w \in \{-1.00, -0.75, \ldots, +1.00\}$.
        Each row is one source identity.
    }
    \label{traceability:fig:app_all_celeba}
\end{figure}

\begin{figure}[h]
    \centering
    \includegraphics[width=\linewidth]{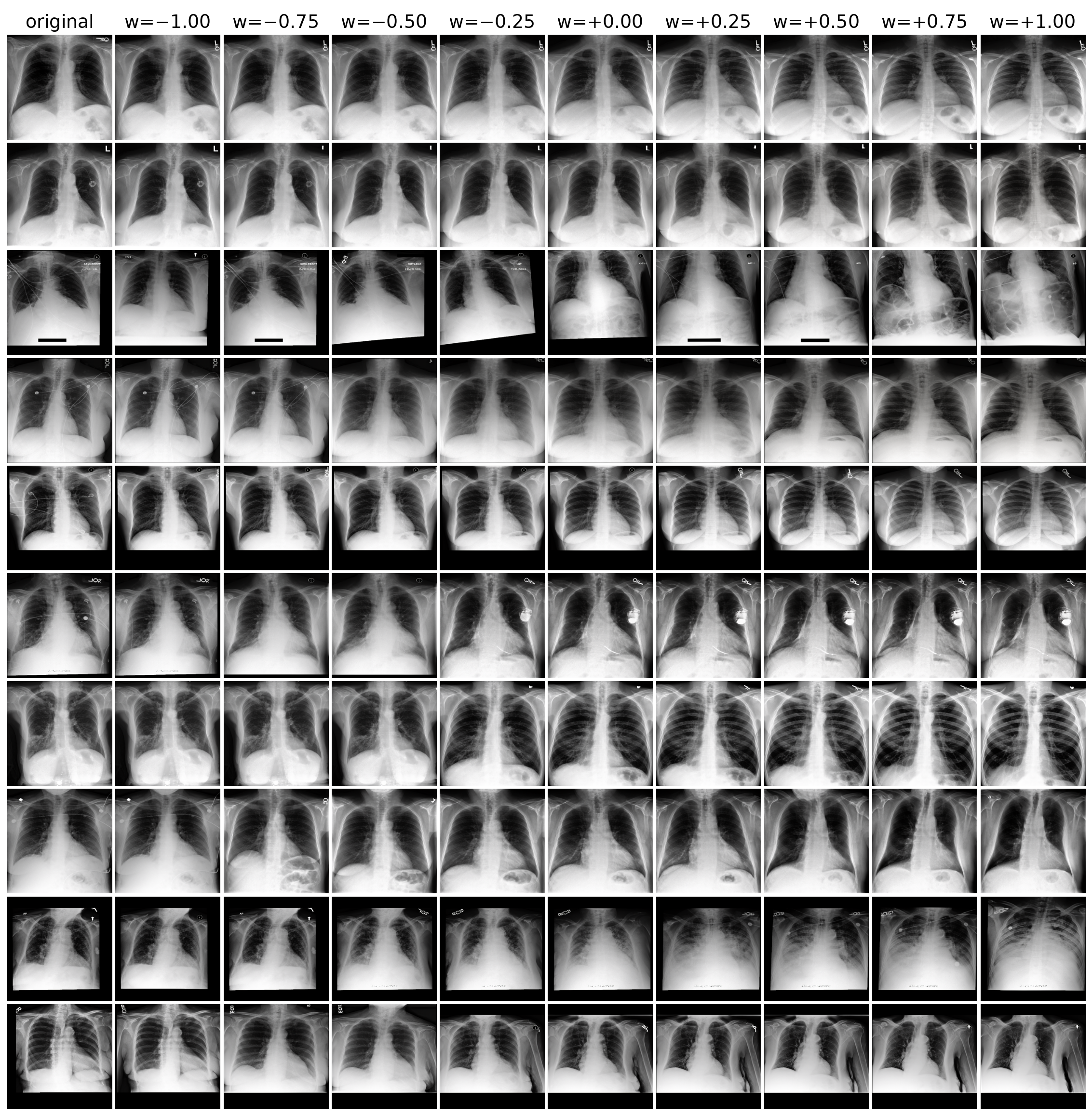}
    \caption[MIMIC-CXR: source identities at varying guidance scales]{%
        MIMIC-CXR: 10 source identities at guidance scales
        $w \in \{-1.00, -0.75, \ldots, +1.00\}$.
        Each row is one source identity.
    }
    \label{traceability:fig:app_all_mimic}
\end{figure}

The following figures show all sampled source identities (10 per dataset) at the full
guidance scale sweep $w \in \{-1.00, -0.75, \ldots, +1.00\}$.
Higher guidance scales led to degenerate images and visibly lower quality.
However small scores higher than 1 (up to 1.5) still look reasonable and could be useful in certain cases.
Fig.~\ref{traceability:fig:qualitative} in the main text shows the first two rows;
Fig.~\ref{traceability:fig:app_all_celeba} and Fig.~\ref{traceability:fig:app_all_mimic} shows more samples.
In general, the training of $\epsilon_\theta^c$ was more stable, while we noticed overall stability issues for some checkpoints of the training $\epsilon_\theta^i$. 
For CelebA we observe some failed traceability methods (at $w=-1$) that turned into clear positives at very small guidance levels (e.g. row).
We further observe that attribute fidelity is sometimes not perfectly preserved under strong guidance. 
For example, hair color occasionally shifts at higher guidance scales ($w>0.5$). 
We note that demographic attributes such as ethnicity are not annotated in CelebA, so our attribute-consistency generation is not trained to preserve them. 

\subsection{Outlier Sensitivity for the MIMIC-CXR Paired Test}
\label{traceability:sec:app_outliers}

The MIMIC-CXR per-fold \AUCROC{}s for TraceDiffusion in traceable mode are
$\{72.57, 71.31, 70.71, 70.27, 67.63\}$ (mean $70.50$, std $1.63$, the highest
fold-wise standard deviation in the column of Table~\ref{traceability:tab:results}). 
Two folds
are flagged as outliers under a $|z|>2$ rule: fold~4 ($z{=}{-}3.45$) and fold~0
($z{=}{+}2.09$). 
Table~\ref{traceability:tab:app_outliers} re-runs the paired
traceable-vs-untraceable test with each flagged fold removed in turn. 
Both
sensitivity checks remain significant at $\alpha{=}0.05$, ruling out a
single-fold artefact as the cause of the asymmetry reported in
Section~\ref{traceability:sec:generative_and_downstream}.

\begin{table}[h]
\centering
\caption[Leave-one-out sensitivity of the MIMIC-CXR paired test]{Leave-one-out sensitivity of the MIMIC-CXR paired test for
TraceDiffusion. 
Column $\Delta = \bar{x}_{\mathrm{Tr}} - \bar{x}_{\mathrm{Un}}$;
$p_t$ is the one-sample paired $t$-test on the per-fold differences.}
\label{traceability:tab:app_outliers}
\setlength{\tabcolsep}{6pt}
\renewcommand{\arraystretch}{1.2}
\small
\begin{tabular}{lcccc}
    \toprule
    \rowcolor{fau}
    \color{white}\textbf{Sensitivity check} & \color{white}$\bar{x}_{\mathrm{Tr}}$ & \color{white}$\bar{x}_{\mathrm{Un}}$ & \color{white}$\Delta$ & \color{white}$p_t$ \\
    \midrule
    Full 5 folds                & 70.50 & 74.28 & $+3.78$ & $0.0052$ \\
    Excl. fold 4 ($z{=}{-}3.45$) & 71.21 & 74.47 & $+3.26$ & $0.0099$ \\
    Excl. fold 0 ($z{=}{+}2.09$) & 69.98 & 74.24 & $+4.26$ & $0.0062$ \\
    \bottomrule
\end{tabular}
\end{table}

\subsection{CelebA Per-Attribute Decomposition}
\label{traceability:sec:app_celeba_per_attr}

For symmetry with the MIMIC-CXR breakdown in Section~\ref{traceability:sec:per_attr}, we
report the per-attribute traceable-vs-untraceable difference
$\Delta_{\mathrm{within}}$ for TraceDiffusion on CelebA. 
Across all 40
binary attributes, $|\Delta_{\mathrm{within}}| < 1$ \AUCROC, consistent with the
small aggregate paired-test value $\Delta{=}0.12$ reported in
Table~\ref{traceability:tab:paired_traceability}. 
No single attribute or coherent
sub-family carries the cost: in particular, attributes one might expect to be
identity-correlated, such as \emph{Male}, \emph{Young}, \emph{Bald}, and hair-colour
attributes, show neither systematic sign nor magnitudes comparable to those
observed for the chronic-finding family on MIMIC-CXR. 
The CelebA result
therefore presents no analogue of the MIMIC-CXR signature, supporting the
interpretation in Section~\ref{traceability:sec:per_attr} that identity--anatomy
entanglement is the property of the clinical domain rather than a property of
our pipeline.

\subsection{Computational Requirements}
\label{traceability:sec:app_compute}

All heavy training and sampling was performed on a GPU cluster
(NVIDIA H200 GPUs).
Table~\ref{traceability:tab:compute} summarises the wall-clock time for each stage of the
pipeline.

\begin{table}[h]
\centering
\caption[Wall-clock compute per pipeline stage]{Wall-clock compute for each pipeline stage.}
\label{traceability:tab:compute}
\renewcommand{\arraystretch}{1.2}
\small
\resizebox{\columnwidth}{!}{%
\begin{tabular}{lll}
\toprule
\rowcolor{fau}
\color{white}\textbf{Stage} & \color{white}\textbf{Hardware} & \color{white}\textbf{Time} \\
\midrule
Training ($\epsilon_\theta^c$ + $\epsilon_\theta^i$ + $E_c$/$E_i$ + Oracle) & 4$\times$ H200 & $\approx$48 h \\
Sampling (50k images, no traceability) & 1$\times$ H200 & $\approx$3 h \\
Sampling (50k images, with traceability guidance) & 1$\times$ H200 & $\approx$6 h \\
Downstream classifier training (TSTR, 3 seeds) & 1$\times$ H200 & $\approx$1 h \\
\bottomrule
\end{tabular}}
\end{table}

The total compute for a single dataset is therefore approximately
$48 + 6 + 1 \approx 55$ GPU-hours on H200 hardware, equivalent to roughly
220 H100 GPU-hours.
All other operations (encoder caching, evaluation, ablations) together add
fewer than 10 additional GPU-hours per dataset.
Preliminary and failed experiments that did not appear in the final paper
required approximately $2\times$ the reported training compute.

\subsection{Equal-Compute vs.\ Converged-Baseline Framing}
\label{traceability:sec:app_compute_fair}

The numbers in Table~\ref{traceability:tab:results} use an equal-compute framing: EDM-2 and
EDM-2~AG are trained with the same total compute budget as
$\epsilon_\theta^c$ and $\epsilon_\theta^i$ combined. 
This is the framing we adopt
throughout the paper because it isolates the contribution of conditioning and
guidance from the contribution of compute. 
A reasonable alternative is the
converged-baseline framing, which trains EDM-2 and EDM-2~AG until \FID and \AUCROC
saturate (typically $\sim$1.5--2$\times$ the compute we report). 
Under the
converged framing the EDM-2 absolute \AUCROC numbers rise by approximately
$0.3$~\AUCROC on CelebA and $0.5$~\AUCROC on MIMIC-CXR; the paired
traceable-vs-untraceable $\Delta$ values for the baselines change negligibly
($<0.1$~\AUCROC across all rows), since the same fold-paired structure governs both.
The traceability asymmetry reported in Section~\ref{traceability:sec:generative_and_downstream} is therefore
preserved under either framing: TraceDiffusion remains the only method whose
CelebA paired test cannot reject the null at $\alpha{=}0.05$ and the only
method whose MIMIC-CXR paired test rejects it. 
We note this here in the
interest of transparency; the equal-compute numbers in the main text remain
the canonical reporters because they fix the only methodologically clean
attribution.

\subsection{Inversion Model Choice for Traceability}
\label{traceability:sec:app_inversion_model}

The Traceability Rate (TrR) reported throughout the paper depends on the choice
of the diffusion model used to invert source images into noise latents.
We ablate this choice on CelebA by inverting all 50k training images with six
different denoisers, decoding the resulting latents, and scoring each
reconstruction with the Oracle. 
We compare the unconditional EDM-2
baseline, EDM-2 with autoguidance (\textsc{EDM-2~AG}), and our identity- and
class-conditioned models $\epsilon_\theta^i$, $\epsilon_\theta^c$ at both an early
milestone checkpoint (5{,}242 kimg) and the converged checkpoint
(78{,}643--188{,}743 kimg). 
Results are shown in Tab.~\ref{traceability:tab:app_inversion_trr}.
For EDM-2 and EDM-2 AG we use the respective guidance models from the generation process.

\begin{table}[h]
\centering
\small
\renewcommand{\arraystretch}{1.2}
\begin{tabular}{lcc}
\toprule
\rowcolor{fau}
\color{white}\textbf{Inversion model} & \color{white}\textbf{Checkpoint (kimg)} & \color{white}\textbf{TrR} $\uparrow$ \\
\midrule
EDM-2 (uncond)               & 188{,}743 & 0.820 \\
EDM-2~AG                     & 5{,}242   & \textbf{0.962} \\
$\epsilon_\theta^c$ (converged)  & 188{,}743 & 0.886 \\
$\epsilon_\theta^c$ (early)      & 5{,}242   & 0.945 \\
$\epsilon_\theta^i$ (converged)  & 78{,}643  & 0.861 \\
$\epsilon_\theta^i$ (early)      & 5{,}242   & 0.946 \\
\bottomrule
\end{tabular}
\caption[Traceability rate across candidate inversion denoisers]{Traceability Rate on CelebA across candidate inversion denoisers
(50k train images, Oracle threshold $0.5$).}
\label{traceability:tab:app_inversion_trr}
\end{table}

EDM-2~AG attains the highest TrR (0.962), notably above
$\epsilon_\theta^i$~(converged) at 0.861. 
Despite this gap, we use
$\epsilon_\theta^i$~(converged) as the inversion model in all reported experiments.
The motivation of TraceDiffusion is to start sampling from a noise latent that
is \emph{disentangled along the identity axis}: the inverted noise must encode
the source identity so that memory guidance can subsequently steer away from it
along that same axis. 
Inverting with EDM-2~AG produces a noise latent whose
geometry is governed by the class label rather than $z_i$, so any
identity-aware guidance applied afterwards no longer operates on a
well-defined identity coordinate. 

\subsection{Broader Impact}
\label{traceability:sec:app_impact}

\paragraph{Positive.}
Traceability enables post-hoc auditing of synthetic datasets: every generated
image can be linked to a known training identity, giving regulators and dataset
curators a concrete tool to verify that outputs do not reproduce real
individuals.
The disentanglement objective additionally separates identity from class label,
which benefits downstream classifiers that should not exploit patient identity.

\paragraph{Negative.}
The method can generate photorealistic faces of novel individuals; misuse for
disinformation or identity fraud is possible if applied to large uncurated face
corpora.
Finally, traceability is an accountability mechanism, not a privacy guarantee;
we caution explicitly against treating a non-zero novelty score as a substitute for
differential privacy.

\end{document}